\documentclass{article}
\usepackage{amsmath, amsthm, amssymb}
\usepackage{thmtools, thm-restate}

\usepackage{expl3}  %
\ExplSyntaxOn
  _to_str:N
\ExplSyntaxOff
\usepackage{ifthen}
\usepackage{pgffor}

\usepackage[acronym,nowarn,section,nonumberlist]{glossaries}
\setacronymstyle{long-sc-short}
\glsdisablehyper  %
\makeglossaries

\setacronymstyle{long-short}

\newacronym{auc}{AUC}{Area Under the Curve}
\newacronym{asr}{ASR}{Automatic Speech Recognition}

\newacronym{bert}{BERT}{Bidirectional Encoder Representations from Transformers}
\newacronym{bow}{BOW}{Bag of Words}
\newacronym{boe}{BOE}{Bag of Embeddings}
\newacronym{borep}{BOREP}{Bag of Random Embedding Projections}
\newacronym{bn}{BN}{Bayesian Network}
\newacronym{byol}{BYOL}{Bootstrap Your Own Latent}

\newacronym{cdf}{CDF}{Cumulative Distribution Function}
\newacronym{cnn}{CNN}{Convolutional Neural Network}
\newacronym{cka}{CKA}{Centered Kernel Alignment}
\newacronym{ctc}{CTC}{Connectionist Temporal Classification}
\newacronym{cvae}{CVAE}{Conditional Variational Auto-Encoder}
\newacronym{cv}{CV}{Computer Vision}

\newacronym{dag}{DAG}{Directed Acyclic Graph}
\newacronym{ddpg}{DDPG}{Deep Deterministic Policy Gradient}
\newacronym{dino}{DINO}{DIstillation with NO labels}
\newacronym{dgm}{DGM}{Directed Graphical Model}
\newacronym{dnn}{DNN}{Deep Neural Network}

\newacronym{ecfp4}{ECFP4}{Extended Connectivity Fingerprint}
\newacronym{ehr}{EHR}{Electronic Health Record}
\newacronym{ema}{EMA}{Exponential Moving Average}
\newacronym{esn}{ESN}{Echo State Network}
\newacronym{esr}{ESR}{Exponential Scaling Rule}

\newacronym{fft}{FFT}{Fast Fourier Transform}
\newacronym{fs}{FS}{FastSent}
\newacronym{ft}{FT}{Fourier Transform}

\newacronym{gat}{GAT}{Graph Attention Network}
\newacronym{gcn}{GCN}{Graph Convolutional Network}
\newacronym{gd}{GD}{Gradient Descent}
\newacronym{gdl}{GDL}{Geometric Deep Learning}
\newacronym{ggnn}{GGNN}{Gated Graph Neural Network}
\newacronym{glove}{GloVe}{Global Vectors for Word Representation}
\newacronym{gnn}{GNN}{Graph Neural Network}
\newacronym{gru}{GRU}{Gated Recurrent Unit}

\newacronym{hmm}{HMM}{Hidden Markov Model}

\newacronym{idf}{IDF}{Inverse Document Frequency}
\newacronym{ig}{IG}{Information Gain}
\newacronym{iid}{i.i.d.}{Independent and Identically Distributed}
\newacronym{isan}{ISAN}{Input Switched Affine Network}

\newacronym{jsd}{JSD}{Jensen-Shannon Divergence}

\newacronym{kld}{KLD}{Kullback-Leibler Divergence}
\newacronym{knn}{KNN}{$K$-Nearest Neighbour}

\newacronym{lhs}{L.H.S.}{left hand side}
\newacronym{lstm}{LSTM}{Long Short Term Memory Network}
\newacronym{lsr}{LSR}{Linear Scaling Rule}

\newacronym{map}{MAP}{Maximum a Posteriori}
\newacronym{mc}{MC}{Monte Carlo}
\newacronym{mdl}{MDL}{Minimum Description Length}
\newacronym{ml}{ML}{Machine Learning}
\newacronym{mlp}{MLP}{Multi-Layer Perceptron}
\newacronym{mle}{MLE}{Maximum Likelihood Estimation}
\newacronym{mnist}{MNIST}{Modified National Institute of Standards and Technology}

\newacronym{nlp}{NLP}{Natural Language Processing}
\newacronym{nn}{NN}{Neural Network}
\newacronym{nll}{NLL}{Negative Log Likelihood}

\newacronym{ode}{ODE}{Ordinary Differential Equation}
\newacronym{oh}{OH}{One-Hot}

\newacronym{pdf}{PDF}{Probability Density Function}
\newacronym{pgm}{PGM}{Probabilistic Graphical Model}
\newacronym{pl}{PL}{Pseudo-Label}

\newacronym{relu}{ReLU}{Rectified Linear Unit}
\newacronym{rdf}{RDF}{Resource Description Framework}
\newacronym{rgb}{RGB}{Red Green Blue}
\newacronym{rgc}{RGC}{Relational Graph Convolution}
\newacronym{rgat}{RGAT}{Relational Graph Attention Network}
\newacronym{rgcn}{RGCN}{Relational Graph Convolutional Network}
\newacronym{rhs}{R.H.S.}{Right Hand Side}
\newacronym{rl}{RL}{Reinforcement Learning}
\newacronym{rnn}{RNN}{Recurrent Neural Network}
\newacronym{roc}{ROC}{Receiver Operating Characteristic}

\newacronym{sde}{SDE}{Stochastic Differential Equation}
\newacronym{sgd}{SGD}{Stochastic Gradient Descent}
\newacronym{sg}{SG}{SkipGram}
\newacronym{simclr}{SimCLR}{Simple framework for Contrastive Learning of visual Representations}
\newacronym{ssl}{SSL}{Self-Supervised Learning}
\newacronym{st}{ST}{SkipThought}
\newacronym{sts}{STS}{Semantic Textual Similarity}
\newacronym{swa}{SWA}{Stochastic Weight Averaging}
\newacronym{swag}{SWAG}{SWA-Gaussian}

\newacronym{termfreq}{TF}{Term Frequency}
\newacronym{tfidf}{TF-IDF}{Term Frequency-Inverse Document Frequency}
\newacronym[plural=TPPs, descriptionplural={Temporal Point Processes}]{tpp}{TPP}{Temporal Point Process}

\newacronym{vae}{VAE}{Variational Auto-Encoder}
\newacronym{vit}{ViT}{Vision Transformer}

\newacronym{wer}{WER}{Word Error Rate}
\newacronym{wirgat}{WIRGAT}{Within-Relation Graph Attention}
\newacronym{wl}{WL}{Weisfeiler-Lehman}

\def\tco{{\it train-clean-100}}
\def\tct{{\it train-clean-360}}
\def\tof{{\it train-other-500}}
\def\devclean{{\it dev-clean}}
\def\devother{{\it dev-other}}
\def\testclean{{\it test-clean}}
\def\testother{{\it test-other}}
\newcommand{\librispeech}{{LibriSpeech}}

\usepackage{bm}
\usepackage[T1]{fontenc}  %
\usepackage[type1]{libertine}
\usepackage[libertine]{newtxmath}
\usepackage{mathtools}

\usepackage{hyperref}
\usepackage{xcolor}
\definecolor{hrefblue}{RGB}{84,151,193}
\definecolor{hrefred}{RGB}{227,94,105}
\hypersetup{
  colorlinks   = true, %
  urlcolor     = hrefblue, %
  linkcolor    = hrefblue, %
  citecolor   = hrefred %
}
\usepackage{cleveref}
\creflabelformat{equation}{#2\textup{#1}#3}
\usepackage{caption}
\usepackage{subcaption}
\usepackage{booktabs}       %

\usepackage[font=small]{caption}

\usepackage{enumitem}

\usepackage{placeins}
\usepackage{titletoc}
\usepackage[page,header]{appendix}

\usepackage{algorithm}
\usepackage{algpseudocode}

\usepackage{ifthen}
\usepackage{setspace}

\DeclareRobustCommand{\mathup}[1]{\begingroup\changegreek\mathrm{#1}\endgroup}
\DeclareRobustCommand{\mathbfup}[1]{\begingroup\changegreekbf\mathbf{#1}\endgroup}
\DeclareRobustCommand{\mathbit}[1]{\bm{\mathit{#1}}}

\DeclareMathAlphabet{\mathsfit}{\encodingdefault}{\sfdefault}{m}{sl}
\SetMathAlphabet{\mathsfit}{bold}{\encodingdefault}{\sfdefault}{bx}{n}
\newcommand{\tens}[1]{\bm{\mathsfit{#1}}}

\newcommand{\constantvector}{\bm}               %

\newcommand{\constantmatrix}{\bm}               %
\newcommand{\constantmatrixgreek}{\mathbit}

\newcommand{\randomscalar}{\textnormal}         %
\newcommand{\randomscalargreek}{\mathup}

\newcommand{\randomvector}{\mathbf}             %
\newcommand{\randomvectorgreek}{\mathbfup}

\newcommand{\randommatrix}{\mathbf}             %
\newcommand{\randommatrixgreek}{\mathbfup}

\newcommand{\graphstyle}{\mathcal}              %

\newcommand{\tensorstyle}{\tens}                %

\newcommand{\setstyle}{\mathbb}                %

\def\alphabet{a,b,c,d,e,f,g,h,i,j,k,l,m,n,o,p,q,r,s,t,u,v,w,x,y,z}
\def\Alphabet{A,B,C,D,E,F,G,H,I,J,K,L,M,M,O,P,Q,R,S,T,U,V,W,X,Y,Z}
\def\greekalphabet{alpha,beta,gamma,delta,epsilon,varepsilon,zeta,eta,theta,vartheta,iota,kappa,varkappa,lambda,mu,nu,xi,pi,varpi,rho,varrho,sigma,varsigma,tau,upsilon,phi,varphi,chi,psi,omega}
\def\GreekAlphabet{Gamma,Delta,Theta,Lambda,Xi,Pi,Sigma,Upsilon,Phi,Psi,Omega}

\makeatletter
\def\changegreek{\@for\next:=\greekalphabet
	\do{\expandafter\let\csname\next\expandafter\endcsname\csname\next up\endcsname}}
\def\changegreekbf{\@for\next:=\greekalphabet
	\do{\expandafter\def\csname\next\expandafter\endcsname\expandafter{%
			\expandafter\bm\expandafter{\csname\next up\endcsname}}}}
\makeatother

\foreach \x in \alphabet {
	\expandafter\xdef\csname v\x\endcsname{\noexpand\ensuremath{\noexpand\constantvector{\x}}}

	\expandafter\xdef\csname ev\x\endcsname{\noexpand\ensuremath{\noexpand\x}}

	\ifthenelse{\equal{\x}{m}}{}{
		\expandafter\xdef\csname r\x\endcsname{\noexpand\ensuremath{\noexpand\randomscalar{\x}}}
	}

	\expandafter\xdef\csname rv\x\endcsname{\noexpand\ensuremath{\noexpand\randomvector{\x}}}
}

\foreach \x in \greekalphabet {
	\expandafter\xdef\csname v\x\endcsname{\noexpand\ensuremath{\noexpand\constantvector{\csname \x\endcsname}}}

	\expandafter\xdef\csname ev\x\endcsname{\noexpand\ensuremath{\noexpand{\csname \x \endcsname}}}

	\expandafter\xdef\csname r\x\endcsname{\noexpand\ensuremath{\noexpand\randomscalargreek{\csname \x\endcsname}}}

	\expandafter\xdef\csname rv\x\endcsname{\noexpand\ensuremath{\noexpand\randomvectorgreek{\csname \x\endcsname}}}
}

\foreach \x in \Alphabet {
	\expandafter\xdef\csname m\x\endcsname{\noexpand\ensuremath{\noexpand\constantmatrix{\x}}}

	\expandafter\xdef\csname em\x\endcsname{\noexpand\ensuremath{\noexpand\x}}

	\expandafter\xdef\csname rm\x\endcsname{\noexpand\ensuremath{\noexpand\randommatrix{\x}}}

	\expandafter\xdef\csname t\x\endcsname{\noexpand\ensuremath{\noexpand\tensorstyle{\x}}}

	\expandafter\xdef\csname g\x\endcsname{\noexpand\ensuremath{\noexpand\graphstyle{\x}}}

	\ifthenelse{\equal{\x}{E}}{}{
		\expandafter\xdef\csname s\x\endcsname{\noexpand\ensuremath{\noexpand\setstyle{\x}}}
	}

}

\foreach \x in \GreekAlphabet {
	\expandafter\xdef\csname m\x\endcsname{\noexpand\ensuremath{\noexpand\constantmatrixgreek{\csname \x\endcsname}}}

	\expandafter\xdef\csname rm\x\endcsname{\noexpand\ensuremath{\noexpand\randommatrixgreek{\csname \x\endcsname}}}
}

\DeclareMathOperator*{\argmin}{arg\,min}

\DeclareMathOperator{\Var}{Var}

\newcommand{\E}{\mathbb{E}}
\newcommand{\Ls}{\mathcal{L}}
\newcommand{\R}{\mathbb{R}}

\def\eps{{\epsilon}}

\newtheorem{definition}{Definition}[section]

\newtheorem{theorem}{Theorem}[section]
\newtheorem{corollary}{Corollary}[theorem]

\newcommand{\inParentheses}[1]{\left(#1\right)}

\newcommand{\anonymous}{0}

\usepackage[final]{neurips_2023}

\title{How to Scale Your EMA}

\newcommand{\compute}{\paragraph{Compute} \emph{[This section has been redacted to preserve anonymity during the peer-review process. If this work is accepted, the full details compute used for these experiments, including: the experiments presented, hyperparameter optimization, and the development process, will be provided.]}}

\newcommand*\samethanks[1][\value{footnote}]{\footnotemark[#1]}

\author{
  Dan Busbridge\thanks{Primary contributor. For a detailed breakdown of author contributions see \Cref{sec:attribution}.}
  \And
  Jason Ramapuram\samethanks
  \And 
  Pierre Ablin\samethanks \And  
  Tatiana Likhomanenko\samethanks \AND
  Eeshan Gunesh Dhekane \And
  Xavier Suau \And
  Russ Webb \And
  \vspace{-0.5cm} \\ \phantom{Mystery} \\
  Apple \\ \\
  \footnotesize\texttt{\{dbusbridge, jramapuram, p\_ablin, antares,}\\
  \footnotesize\texttt{eeshan, xsuaucuadros, rwebb\}@apple.com} \\
\\
}

\begin{document}

\maketitle
\begin{abstract}
Preserving training dynamics across batch sizes is an important tool for practical machine learning as it enables the trade-off between batch size and wall-clock time.
This trade-off is typically enabled by a scaling rule, for example, in stochastic gradient descent, one should scale the learning rate linearly with the batch size. 
Another important machine learning tool is the model EMA,
a functional copy of a target model, whose 
parameters move towards those of its target model according to an Exponential Moving Average (EMA) at a rate parameterized by a momentum hyperparameter.  
This model EMA can improve the robustness and generalization of supervised learning, stabilize pseudo-labeling, and provide a learning signal for Self-Supervised Learning (SSL). 
Prior works have not considered the optimization of the model EMA when performing scaling, leading to different training dynamics across batch sizes and lower model performance.
In this work, we provide a scaling rule for optimization in the presence of a model EMA and demonstrate the rule's validity across a range of architectures, optimizers, and data modalities. 
We also show the rule's validity where the model EMA contributes to the optimization of the target model, enabling us to train EMA-based pseudo-labeling and SSL methods at small and large batch sizes. For SSL, we enable training of BYOL up to batch size 24,576
without sacrificing performance, a 6$\times$ wall-clock time reduction under idealized hardware settings.
\end{abstract}

\section{Introduction}
\label{sec:introduction}

With data and models becoming progressively larger
\citep{DBLP:conf/nips/ChenKSNH20,DBLP:journals/corr/abs-2001-08361,DBLP:journals/corr/abs-2108-07258,DBLP:journals/corr/abs-2206-04615}, 
the ability to reduce training wall--clock time is a requirement for practical \gls{ml} at scale.
Optimizer scaling rules allow us to find faster learning procedures that produce similar results.
For example, the \emph{linear scaling rule} for \gls{sgd}
\citep{DBLP:journals/corr/Krizhevsky14,DBLP:journals/corr/GoyalDGNWKTJH17},
states that the learning rate should be scaled linearly with the batch size.
This optimizer scaling works \emph{both ways}.
Access to larger computational resources means one can train equivalent models in reduced wall-clock time.
Alternatively, with access to limited computational resources, larger distributed computations can be replicated at increased wall-clock time.

Many \gls{ml} algorithms rely on a \emph{model EMA},
a functional copy of a \emph{target model}\footnote{The target model usually undergoes gradient-based optimization, but this does not have to be the case.}, whose 
parameters move towards those of its target model according to an \gls{ema} (\Cref{def:emaUpdateDefinition}) at a rate parameterized by a momentum hyperparameter $\rho$.  
\begin{restatable}[EMA Update]{definition}{firstema}
\label{def:emaUpdateDefinition}
     The \gls{ema} update for the model \gls{ema} parameters $\rvzeta_t$ following target model parameters  $\rvtheta_t$ at iteration $t$ with momentum $\rho\equiv1-\beta_\rho$ is
    \label{def:ema}
    \begin{equation}
    \rvzeta_{t+1}
    =
    \rho \,\rvzeta_t + (1-\rho)\,\rvtheta_t
    \equiv
    (1-\beta_\rho) \,\rvzeta_t +\beta_\rho\,\rvtheta_t.
    \end{equation}
\end{restatable}
The \emph{model \gls{ema}} has a number of desirable properties:
i) 
the model \gls{ema} inhabits wider minima than the target model, reducing overfitting and improving generalization
\citep{Ruppert1988EfficientEF,Polyak92,DBLP:conf/iclr/HuangLP0HW17,DBLP:conf/uai/IzmailovPGVW18,DBLP:conf/cvpr/HeCXLDG22};
ii)
compared to the target model, the model \gls{ema} moves slowly, 
making it useful as a stabilizer for networks governing Bellman updates in 
reinforcement learning,
\citep{DBLP:journals/corr/LillicrapHPHETS15};
and iii)
the model \gls{ema} is relatively cheap to compute, whilst providing a valid model but \emph{different} to the target model.
This third property has made the model \gls{ema} a common choice for the \emph{teacher} in many distillation setups, 
from semi-supervised learning \citep{DBLP:conf/nips/TarvainenV17,sohn2020fixmatch,manohar2021kaizen,higuchi2022momentum},
to \gls{ssl} methods like 
\gls{byol} \citep{DBLP:conf/nips/GrillSATRBDPGAP20},
DINO \citep{DBLP:journals/corr/abs-2104-14294},
and data2vec \citep{baevski2022data2vec,DBLP:journals/corr/abs-2212-07525}.

Despite its significant role in optimization, a recipe for adapting the \gls{ema} Update (\Cref{def:emaUpdateDefinition}) when changing batch size has,
to the best of our knowledge, been absent.
To address this, we derive an \gls{ema} Scaling Rule (\Cref{def:ema-sr})
which states how the \gls{ema} momentum $\rho$ hyperparameter \emph{should} be modified\footnote{We stress that the study of momentum in gradient-based optimizers is not the focus of this work.
We refer to \citet{DBLP:conf/iclr/SmithL18,li2019stochastic} for a discussion on scaling rules for these methods.
}.
\begin{restatable}[\gls{ema} Scaling Rule]{definition}{firstemascaling}
\label{def:emaScalingRuleExponentialVersion}
    When computing the \gls{ema} update (\Cref{def:ema}) of a model undergoing stochastic optimization with batch size $\hat B=\kappa B$,
    use a momentum $\hat\rho=\rho^\kappa$ and scale other optimizers according to their own scaling rules.
    \label{def:ema-sr}
\end{restatable}
In \Cref{def:ema-sr}, the momentum $\rho$, which is defined at batch size $B$, typically corresponds to a ``good hyperparameter choice'', although this does not need to be the case in general.
In this paper, we make the following contributions.
\begin{enumerate}[leftmargin=0.75cm]
    \item With the assumptions of \citet{DBLP:journals/corr/GoyalDGNWKTJH17}, we derive an \gls{ema} Scaling Rule: the \gls{ema} update \emph{momentum} should be scaled \emph{exponentially} with the batch size (\Cref{def:ema-sr}).
    \item To validate this EMA Scaling Rule theoretically, we propose \gls{sde} approximations of optimization in the presence of a model \gls{ema} (\Cref{subsec:ema-sdes}). 
    This model \gls{ema} contributes to the loss, covering semi-supervised learning and \gls{ssl}.
    We prove that these approximations are first order weak approximations, and that our \gls{ema} Scaling Rule is correct in the \gls{sde} limit under realistic gradient assumptions (\Cref{cor:validity-scaling-rule}).
    \item We empirically validate the \gls{ema} Scaling Rule in synthetic settings (\Cref{subsec:toy-experiment})
    and real-world settings where the model \gls{ema} plays an increasingly significant role in optimization: 
    i) 
    where the model \gls{ema} is used during inference instead of the target model (\Cref{subsec:supervised-polyakking}); 
    ii) 
    pseudo-labeling, 
    where the model \gls{ema} (\emph{teacher}) follows the target model (\emph{student}), and the \emph{student} is optimized 
    on a mixture of a) labeled data and b) data without labels, whose pseudo-labels are produced by the \emph{teacher} (\Cref{subsec:semi-supervised}); 
    and iii)
    self-supervised learning, which is the same as the semi-supervised case, except there is no labeled data (\Cref{subsec:self-supervised}).
    \item We observe that pseudo-labeling and \gls{ssl} training dynamics during optimizer warm-up are not always able to be replicated at large batch sizes using \emph{only} the \gls{ema} Scaling Rule.
    We propose and verify practical methods to overcome this limitation, enabling us to scale to a batch size of 24,576 with BYOL \glspl{vit}, reducing wall-clock training by 6$\times$ under idealized hardware scenarios while maintaining performance of the batch size 4096 baseline.
\end{enumerate}
Finally, to aid practitioners looking to scale, in \Cref{app:scaling-toolbox} we provide a \emph{Scaling Toolbox}, which gives practical advice on how to scale systematically, collecting known scaling rules, and explaining how to think about the \gls{sde} perspective of optimization.

\section{The EMA Scaling Rule}
\label{sec:a-momentum-scaling-rule}

We begin with an informal discussion of scaling rules and motivate the existence of an exponential scaling rule for the momentum parameter controlling the update of the model \gls{ema}.

\subsection{Background and an informal discussion of scaling rules}
\label{sec:a-momentum-scaling-rule-background}

Consider a model with parameters $\rvtheta_t$ at iteration $t$
updated with \gls{sgd} (\Cref{def:sgd}).
\begin{definition}[SGD Update]
     The \gls{sgd} update for a model with parameters $\rvtheta_t$ at iteration $t$ given a minibatch $\sB=\{x^{(b)}\sim P_{\rvx}:b=1,2,\ldots,B\}$ of $B=|\sB|$ samples with learning rate $\eta$ is
    \begin{equation}
    \rvtheta_{t+1}
    =
    \rvtheta_t - \eta \times \frac1B
    \sum_{x\in\sB} \nabla_{\rvtheta} \Ls(x;\rvtheta_{t}),
    \end{equation}
    where $\Ls$ is the loss function, 
    $\nabla_\theta \Ls(x;\theta_t)$ is the parameter gradient for the sample $x$ at iteration $t$, and the $x\in \sB$ are \gls{iid} from $P_{\rvx}$.
    \label{def:sgd}
\end{definition}
Iterating over a sequence of independent minibatches 
$\sB_0, \sB_1, \ldots, \sB_{\kappa-1}$
produces model parameters
\begin{align}
    \rvtheta_{t+\kappa}
    =
    \rvtheta_t - \eta \times \frac1B
    \sum_{j=0}^{\kappa-1}
    \sum_{x\in\sB_j} \nabla_{\rvtheta} \Ls(x;\rvtheta_{t+j}).
    \label{eq:sgd}
\end{align}
If gradients vary slowly 
$\nabla_{\rvtheta}\Ls(x;\theta_{t+j})\approx \nabla_{\rvtheta}\Ls(x;\theta_{t})$, $j=0,\ldots,\kappa-1$,
\emph{one} \gls{sgd} step with
$\hat\eta=\kappa\eta$ on a batch $\widehat\sB=\cup_i\sB_i$ of size $\hat B=\kappa B$
results in $\hat\rvtheta_{t+1}\approx\rvtheta_{t+k}$,
yielding the \gls{sgd} Scaling Rule (\Cref{def:lsr}).
\begin{definition}[\gls{sgd} Scaling Rule]
    When running \gls{sgd} (\Cref{def:sgd}) with batch size $\hat B=\kappa B$,
    use a learning rate $\hat\eta=\kappa\eta$ \citep{DBLP:journals/corr/Krizhevsky14,DBLP:journals/corr/GoyalDGNWKTJH17}.
    \label{def:lsr}
\end{definition}
For clarity in this work, we adopt the naming convention \emph{[Algorithm Name] Scaling Rule},
which means all parameters of those algorithms are appropriately scaled from batch size $B$ to $\kappa B$.

As discussed in \citet{DBLP:journals/corr/GoyalDGNWKTJH17},
although the assumption of slowly changing gradients is strong, if it is true, then
$\rvtheta_{t+k}\approx \hat\rvtheta_{t+1}$
\emph{only} if $\hat\eta=\kappa\eta$.
The validity of the \gls{sgd} Scaling Rule has been formally studied.
In particular, there was ambiguity regarding whether the scaling should be a square-root or linear \citep{DBLP:journals/corr/Krizhevsky14}.
\gls{sde} approaches have resolved this ambiguity, and have been used to estimate the scaling $\kappa$ when the \gls{sgd} Scaling Rule is no longer guaranteed to hold \citep{DBLP:conf/nips/LiMA21}.

To address model parameter \glspl{ema}, we first restate the \gls{ema} Update (\Cref{def:ema}).
\firstema*
The model \gls{ema} parameters $\rvzeta$ do not typically receive gradient information, we take the convention that $\rho$ is close to one, and the $\beta_\rho$ subscript will be omitted where it is clear from the context.

Assuming again that gradients change slowly $\nabla_{\rvtheta}\Ls(x;\rvtheta_{t+j},\rvzeta_{t+j})\approx \nabla_{\rvtheta}\Ls(x;\rvtheta_{t},\rvzeta_{t})\approx \rvg$, for gradient $\rvg$,
iterating over $\kappa$ independent minibatches produces model states (see \Cref{app:matrix-calculations} for derivation)
\begin{align}
\label{eq:scalingRuleSummaryEquation}
\begin{bmatrix}
\rvtheta_{t+\kappa}
\\
\rvzeta_{t+\kappa}
\\
\rvg
\end{bmatrix}
=
\begin{bmatrix}
1 & 0 & -\eta \\
(1-\rho) & \rho & 0\\
0 & 0 & 1
\end{bmatrix}^\kappa
\cdot 
\begin{bmatrix}
\rvtheta_{t}
\\
\rvzeta_{t}
\\
\rvg
\end{bmatrix}
=
\begin{bmatrix}
\rvtheta_{t}-\eta\,\kappa \,\rvg
\\
\rho^\kappa \, \rvzeta_{t}
+(1-\rho^\kappa) \, \rvtheta_t
+\mathcal O\left(\eta\times \beta_\rho\right)
\\
\rvg
\end{bmatrix}.
\end{align}
The first row is the \gls{sgd} Scaling Rule (\Cref{def:lsr}). The third row 
implements the \emph{slowly changing gradients} assumption for the first row.
The second row is equivalent to a single \gls{ema} update (\Cref{def:ema}) with momentum $\hat\rho=\rho^\kappa$; we can take a \emph{single} \gls{sgd} update with batch size $\hat B=\kappa B$ and learning rate $\hat\eta=\kappa\eta$, and a 
\emph{single} \gls{ema} update with momentum $\hat\rho=\rho^\kappa$, and we get $(\hat\rvtheta_{t+1},\hat\rvzeta_{t+1})\approx(\rvtheta_{t+\kappa},\rvzeta_{t+\kappa})$ up to terms $\mathcal O(\eta \times \mathcal \beta_\rho)$.
This yields the \gls{ema} Scaling Rule (\Cref{def:ema-sr}).
\firstemascaling*

The \gls{ema} Scaling Rule was derived for \gls{sgd}, and is extended to other optimizers in the following way. 
An optimizer scaling rule ensures $\hat\rvtheta_{t+1}=\rvtheta_{t+\kappa}$,
satisfying identification for the first row.
Next, the zeroth order term in $\eta\times\beta_{\rho}$ in the second row in \Cref{eq:scalingRuleSummaryEquation} is optimizer-independent, and therefore unchanged.
Finally, the first order terms in $\eta\times\beta_{\rho}$ in the second row, corresponding to the scaling rule error, are an \gls{ema} accumulation of target model $\rvtheta$ updates under optimization, which is still 
$\mathcal{O}(\eta \times \mathcal  \beta_\rho)$, although its functional form may be different for different optimizers.

The above discussion is intended to give an intuition for why the \gls{ema} momentum should be scaled exponentially.
As we have used the same slow-moving gradient assumption as the original \gls{sgd} Scaling Rule,
this may cast doubt on whether our rule is correct.
To remove this ambiguity, we will follow 
\citet{DBLP:conf/iclr/SmithL18,DBLP:conf/nips/LiMA21,DBLP:conf/nips/MalladiLPA22}, and show that the \gls{ema} Scaling Rule (\Cref{def:ema-sr}) is correct in the \gls{sde} limit under more realistic gradient assumptions. 

\subsection{The EMA Scaling Rule through the lens of stochastic differential equations}
\label{subsec:ema-sdes}

\glspl{sde} are a tool typically used to obtain scaling rules from first principles~\citep{DBLP:conf/nips/LiMA21,DBLP:conf/nips/MalladiLPA22}.
In the following, we use \glspl{sde} to obtain strong theoretical guarantees for the 
\gls{ema} Scaling Rule found in \Cref{sec:a-momentum-scaling-rule-background}.
We consider the following discrete dynamics for \gls{ema}:
\begin{align}
    \label{eq:iterations}
    \begin{split}
    \rvtheta_{k+1} &= \rvtheta_{k} - \eta\, \rvg_k,
    \enspace \text{with }
    \rvg_k=\nabla f(\rvtheta_k, \rvzeta_k) + \sigma \, \rvepsilon_k, 
    \text{ and }
    \rvepsilon_k \sim \mathcal{E}_\sigma(\rvtheta_k, \rvzeta_k),\\
    \rvzeta_{k+1} &= \rho \, \rvzeta_k + (1-\rho) \,\rvtheta_k,
    \end{split}
\end{align}
where $\sigma>0$ is the noise scale, 
$\mathcal{E}_\sigma(\rvtheta_k,  \rvzeta_k)$
is the gradient noise distribution, assumed to be zero-mean and variance 
$\mSigma(\rvtheta_k,  \rvzeta_k)$ 
independent of $\sigma$, and 
$\nabla f(\rvtheta_k, \rvzeta_k)\equiv\nabla_{\rvtheta} f(\rvtheta_k, \rvzeta_k)$.
We posit a dependency of the loss $f$ on the EMA $\rvzeta$ in order to cover semi-supervised (\Cref{subsec:semi-supervised}) and \gls{ssl} (\Cref{subsec:self-supervised}).
The case of Polyak-Ruppert averaging (\Cref{subsec:supervised-polyakking}), is covered by letting $f$ be independent of $\rvzeta$.

We aim to obtain an \gls{sde} approximation of \Cref{eq:iterations} as $\eta$ goes to zero.
The scaling rule for iterations of $\rvtheta$ is well known~\citep{DBLP:conf/nips/LiMA21}: we let $\sigma_0 = \sigma \sqrt{\eta}$.
The analysis of \Cref{sec:a-momentum-scaling-rule-background}
gives the scaling rule $\hat{\eta} = \eta \kappa$ and $\hat{\rho} = \rho^{\kappa}$.
Linearizing this rule near 
$\eta = 0$ 
gives 
$\hat{\rho} = 1 - \kappa\times(1 - \rho)$, which is a linear relationship between $1 -\rho$ and $\eta$. 
We therefore let $\beta_0=(1 - \rho) / \eta$ and consider the SDE 
\begin{align}
    \label{eq:sde-sgd}
    \begin{split}
        d\Theta_t &= - \nabla f(\Theta_t, Z_t)\,dt 
        +
        \sigma_0\,\mSigma(\Theta_t, Z_t)^{\frac12}\,dW_t,
        \enspace \text{with }
        W_t \text{ a Wiener process},\\
        dZ_t &= \beta_0(\Theta_t - Z_t)dt,
    \end{split}
\end{align}
where $\Theta_t$ and $Z_t$ are \gls{sde} variables relating to model and \gls{ema} parameters respectively.
The SDE in \Cref{eq:sde-sgd} approximates the discrete iterations of \Cref{eq:iterations} when the learning rate $\eta$ goes to zero.
One way to see this is that an Euler-Maruyama discretization of the SDE with learning rate $\eta$ exactly recovers the discrete iterations.
More formally, we have \Cref{thm:sde-for-sgd-ema}, which is in the same spirit as those found in~\cite{DBLP:conf/nips/LiMA21,DBLP:conf/nips/MalladiLPA22}. In the theorem, $G^\alpha$ is the set of functions with derivatives up to order $\alpha$ that have at most polynomial growth (see~\Cref{def:polynomial-growth}).
\begin{theorem}[SDE for SGD + EMA; informal see~\Cref{thm:app:sde}]
     Assume that $f$ is continuously differentiable, with $f\in G^3$.
     Let 
     $\Theta_t,Z_t$ 
     be solutions of  \Cref{eq:sde-sgd},
     and $\rvtheta_k,\rvzeta_k$ iterations of \Cref{eq:iterations}
     with
     $\mSigma^{\frac12}\in G^2$. 
     Then, for any time horizon $T >0$ and function $g\in G^2$, there exists a constant $c>0$ independent of $\eta$ such that 
    \begin{equation}
        \max_{k=0,\,\dots\,,\,\lfloor T /\eta \rfloor} |\mathbb{E}[g(\Theta_{\eta k}, Z_{\eta k})] - \mathbb{E}[g(\rvtheta_k, \rvzeta_k)]| \leq c\times  \eta .
    \end{equation}
    \label{thm:sde-for-sgd-ema}
    \vspace{-0.5cm}
\end{theorem}
\Cref{thm:sde-for-sgd-ema} formalizes the intuition that the SDE is an accurate approximation of the discrete iterations. In turn, it allows validating the scaling rule in the same spirit as in~\citet{DBLP:conf/nips/MalladiLPA22}.
\begin{corollary}[Validity of the EMA Scaling Rule]
     Assume that $f$ is continuously differentiable, with $f\in G^3$ and $\mSigma^{\frac12}\in G^2$. 
     Let $\rvtheta_k^{(B)}, \rvzeta_k^{(B)}$ be iterations of \Cref{eq:iterations} with batch size $B$ and hyperparameters $\eta, \rho$. 
     Let $\rvtheta_k^{(\kappa B)}, \rvzeta_k^{(\kappa B)}$ be iterates with batch size $\kappa B$, and $\hat{\eta}$ determined by the \gls{sgd} Scaling Rule (\Cref{def:lsr}) and $\hat{\rho}$ determined by the \gls{ema} Scaling Rule (\Cref{def:ema-sr}). 
     Then, for any time horizon $T >0$ and function $g\in G^2$, there exists a constant $c>0$ independent of $\eta$ such that 
    \begin{equation}
        \max_{k=0,\,\dots\,,\, \lfloor T /\eta \rfloor} |\mathbb{E}[g(\rvtheta_{\lfloor k / \kappa \rfloor}^{(\kappa B)}, \rvzeta_{\lfloor k / \kappa \rfloor}^{(\kappa B)})] - \mathbb{E}[g(\rvtheta_k^{(B)}, \rvzeta_k^{(B)})]| \leq c\times  \eta .
    \end{equation}
    \label{cor:validity-scaling-rule}
    \vspace{-0.5cm}
\end{corollary}
\Cref{cor:validity-scaling-rule} shows that two trajectories with different batch sizes are close in the limit of small learning rate, demonstrating the validity of \Cref{def:ema-sr}.
A natural follow-up question is 
\emph{what happens when an adaptive optimizer is used instead of SGD?}
\citet{DBLP:conf/nips/MalladiLPA22} study this without an \gls{ema} and characterize how hyperparameters change with the noise scale.
In particular, they show that under a high gradient noise hypothesis, there exists a limiting SDE. 
In \Cref{app:ema-approximation-theorem}, we derive the limiting SDEs for RMSProp and Adam with an EMA.
Although a formal proof of closeness between the iterations and these SDEs is beyond the scope of this work, these \glspl{sde} indicate that the EMA Scaling Rule holds for adaptive algorithms. 
We demonstrate this empirically in \Cref{sec:experiments}.

\section{Experiments}
\label{sec:experiments}

\begin{table}[t!]
  \caption{The role of the model \gls{ema} $\rvzeta$ in the optimization of $(\rvtheta, \rvzeta)$ given a target model $\rvtheta$ for different techniques, ordered by increasing influence of the \gls{ema} model.
  All statements assume a momentum $0\leq\rho<1$ and that the target model $\rvtheta$ is subject to stochastic optimization at a batch size $B$. }
  \label{tab:different-ema}
  \centering
  \small
  \begin{tabular}{p{0.22\textwidth}p{0.71\textwidth}}
    \toprule
    \textsc{Technique} & \textsc{Role of Model \gls{ema}} \\
    \midrule
    \textsc{Polyak-Ruppert averaging, Sec. \ref{subsec:supervised-polyakking}} & 
    $\rvtheta$ undergoes optimization and is tracked by
    $\rvzeta$, which does not affect $\rvtheta$. 
    $\rvzeta$ is an estimate of $\rvtheta$ with a time horizon and variance determined by $B$ and $\rho$.
     \\ \midrule
    \textsc{Continuous pseudo-labeling, Sec.~\ref{subsec:semi-supervised}} &
    \emph{Pre-Training} is as above in Polyak-Ruppert Averaging.
    \emph{After Pre-Training}, $\rvzeta$ (\emph{teacher}) produces targets for $\rvtheta$ (\emph{student}) from unlabeled data, which is combined with labeled data.
    The optimization endpoint is dependent on $B$ and $\rho$.\\ \midrule
    \textsc{Self-supervised learning, Sec.~\ref{subsec:self-supervised}} & 
    As above in \emph{After Pre-Training}, except there is no labeled data.
    The optimization endpoint is dependent on $B$ and $\rho$.
    \\
    \bottomrule
  \end{tabular}
  \vspace{-0.3cm}
\end{table}

Now that we have derived and shown the validity of the \gls{ema} Scaling Rule, 
we verify it empirically.
The experiments validate the \gls{ema} Scaling Rule for a variety of uses of \gls{ema} and are ordered by increasing influence of the role of \gls{ema} on the optimization procedure (see \Cref{tab:different-ema}).
The baseline in all of our experiments is \emph{without the EMA Scaling Rule}, which applies all known relevant scaling rules \emph{except} the EMA Scaling Rule, and represents previous best practice.

\subsection{Polyak-Ruppert averaging in a simple setting}
\label{subsec:toy-experiment}

At inference, it is typical to use a model \gls{ema}, known as Polyak-Ruppert Averaging (\Cref{def:polyak-ruppert-average}).

\begin{definition}[Polyak-Ruppert Average]
    When optimizing model parameters $\rvtheta$, compute their \gls{ema} $\rvzeta$ (\Cref{def:ema}).
    Use $\rvzeta$ instead of $\rvtheta$ at inference \citep{Polyak92,Ruppert1988EfficientEF}.
    \label{def:polyak-ruppert-average}
\end{definition}

We begin by showing the \gls{ema} Scaling Rule is \emph{required} to match parameter trajectories in a simple setting.
Consider the optimization of $\rtheta$ in a \emph{noisy parabola}  whose loss $\Ls(\rtheta)$ is parameterized by coefficients for curvature $a>0$,
scaled additive noise $b\geq0$,
and additive noise $c\geq0$:
\begin{align}
    \Ls(\rtheta)
    &=\frac a2\,\rtheta^2,
    &\rtheta_{k+1} &= \rtheta_{k} - \eta \,\rg_k, & \rg_k&=a\,\rtheta_k + \repsilon_k,
    & \repsilon_k\sim \mathcal{N}\left(0, \tfrac{b \,\rg_k^2 + c}\kappa \right).
\end{align}
The scaling factor $\kappa$ in the covariance denominator implements gradient noise reduction as scaling (i.e. batch size) increases \citep{DBLP:journals/corr/abs-1711-04623}.
Let $\rtheta\in\R$ be optimized with \gls{sgd} (\Cref{def:sgd}) and $\rzeta\in\R$ be a Polyak-Ruppert average (\Cref{def:polyak-ruppert-average}) for $\rtheta$ with momentum $\rho=1-\beta$ .
At scaling $\kappa=1$, we use $\beta_B=\eta_B=10^{-4}$ 
and $I_B=10^4$ iterations, to yield a total time $T=I_B\times \eta_B=1$.
To keep gradients $\mathcal O(1)$ and gradient noise non-negligible, we take
$a=1$, $b=0.5$, and $c=0$.
\begin{figure}[ht]
     \centering
     \begin{subfigure}[b]{0.45\textwidth}
         \centering
         \includegraphics[width=\textwidth]{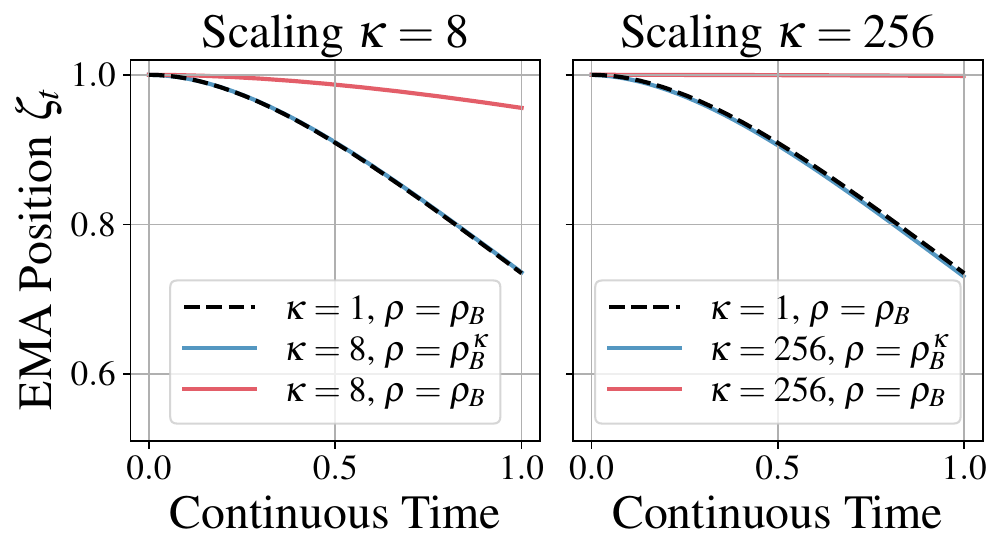}
         \caption{Trajectory of the model \gls{ema} $\rvzeta$ under different scalings $\kappa$, with $1-\rho_B=\eta_B=10^{-4}$.}
         \label{fig:parabola-single-runs}
     \end{subfigure}
     \hfill
     \begin{subfigure}[b]{0.52\textwidth}
         \centering
         \includegraphics[width=\textwidth]{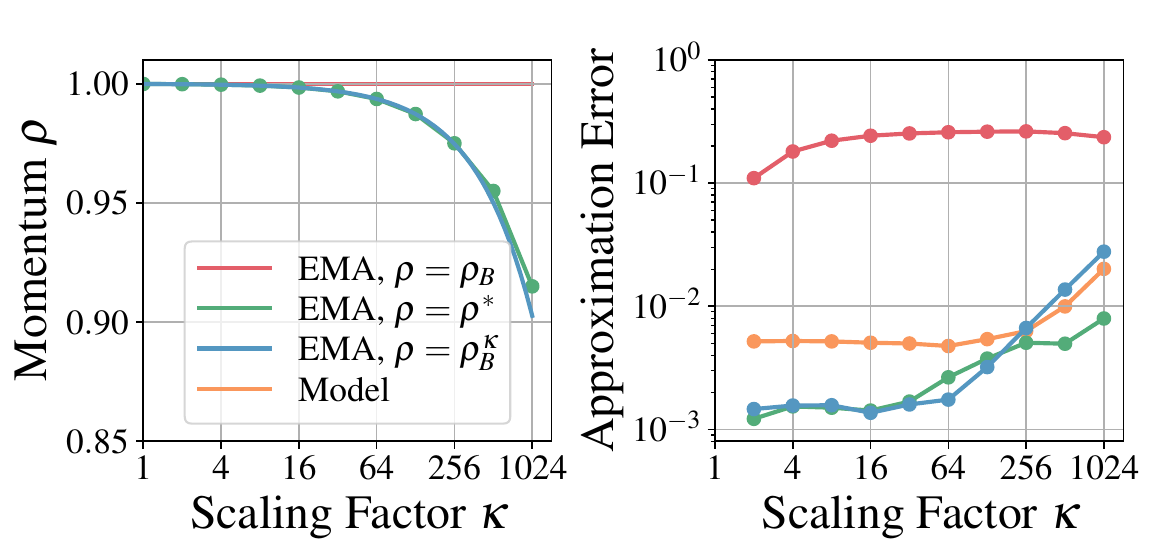}
         \caption{Choices for momentum (left) with corresponding approximation errors (\Cref{eq:optimal-momentum}) (right).}
         \label{fig:curve-approximation-error}
     \end{subfigure}
     \caption{(a) We show the effect of scaling by comparing model \gls{ema} trajectories of the baseline 
     ($\kappa=1$, black dashed) to $\kappa=8$ (left) and $\kappa=256$ (right), 
     with ($\rho=\rho_B^\kappa$, blue) and without 
     ($\rho=\rho_B$, red) the EMA Scaling Rule.
     (b, left) The momentum according for different scaling rules and the empirically optimal $\rho^*$ (\Cref{eq:optimal-momentum}).
     (b, right) The approximation error (\Cref{eq:optimal-momentum}) of trajectories in (b, left) and the target model (orange). 
     }
     \vspace{-0.2cm}
     \label{fig:toy-experiment}
\end{figure}

First, we observe the effect of scaling on a single run (\Cref{fig:parabola-single-runs}) by tracking the position of the model \gls{ema}.
We see that at scaling $\kappa=8$ or $\kappa=256$, the runs using the \gls{ema} Scaling Rule match the baseline trajectory,
whereas the runs using the baseline momentum do not, with a greater deviation induced by greater scaling $\kappa$.
Even at $\kappa=8$, there is a significant difference between scaled and unscaled trajectories, despite the seemingly small numerical difference of their momenta\footnote{Momentum enters optimization exponentially; small changes can lead to very different updates.}.

Second, we consider whether the \gls{ema} Scaling Rule is optimal.
To do this, inspired by the \gls{sde} analysis (\Cref{subsec:ema-sdes}), 
we define the approximation error, $\text{Err}(\rho,\kappa,g)$, of a test function $g$ for a given scaling $\kappa$ using momentum $\rho$, and the value of the momentum $\rho^*(\kappa,g)$ that minimizes this error:
\begin{align}
    \rho^*(\kappa,g)=
    &
    \argmin_\rho
    \text{Err}(\rho,\kappa,g),
    &
    \text{Err}(\rho,\kappa,g)
    &\equiv
    \max_{k=0,\ldots,T/\eta}
    \left|
    \E \,g(\rvzeta_k)
    -
    \E \,g(\rvzeta^{(\kappa,\rho)}_{k/\kappa})
    \right|.
    \label{eq:optimal-momentum}
\end{align}
For scalings $\kappa\in\{1,2,4,\ldots,1024\}$, we determine the optimal momentum $\rho^*$ and compare it to the \gls{ema} Scaling Rule (\Cref{fig:curve-approximation-error}, left).
The scaling rule tracks the $\rho^*$ until $\kappa=256$, when
the $\rho^*$ become systematically higher.
We see target model error increase at $\kappa=256$ (\Cref{fig:curve-approximation-error}, right). 
As the target model error is \gls{ema}-independent, this indicates that the \gls{sgd} Scaling Rule is breaking.
At the lower scaling $\kappa=64$, there is an inflection point in the \gls{ema} Scaling Rule approximation error, before the model error grows.
This difference indicates the $\mathcal{O}(\eta\times \beta_\rho)$ terms of \Cref{eq:scalingRuleSummaryEquation} are beginning to influence the \gls{ema} update.
Finally, 
these observations are true in $D=100$ dimensions, (\Cref{app:noisy-parabola}), and
we stress that \emph{not} changing the momentum at every scaling $\kappa$ induces large approximation error, indicating there is merit to using the \gls{ema} Scaling Rule.

\subsection{Supervised learning on real data with Polyak-Ruppert averaging}
\label{subsec:supervised-polyakking}

We now turn to real-world classification where
the target model $\rvtheta$ optimizes a parametric log-likelihood
$\max_{\rvtheta} \log p(\rvy | \rvx; \rvtheta)$ 
with inputs and labels $(\vx,\vy)$ drawn from a joint distribution $p(\rvy, \rvx)$.

\begin{figure}[t]
    \centering
    \includegraphics[width=0.99\textwidth]{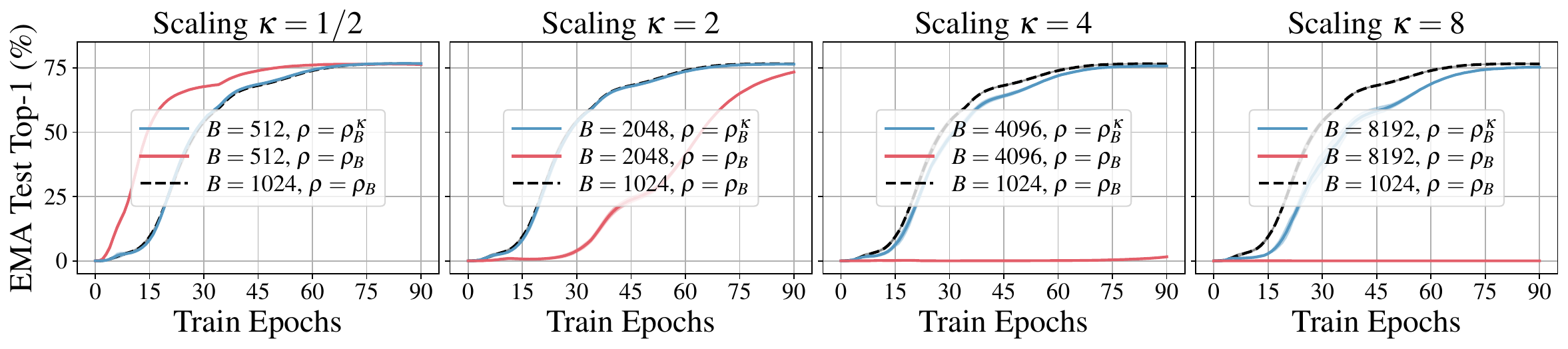}
    \caption{
    \emph{ResNetv2-50 Polyak-Ruppert averaging on ImageNet1k} for different scalings $\kappa$.
    The baseline model ($\kappa=1$, black dashed) uses batch size 1024 and momentum $\rho_B=0.9999$,
    is scaled down to a batch size of 512 (left), and up to a batch size of 4096 (right) with (blue, $\rho=\rho_B^\kappa$) and without (red, $\rho=\rho_B$) the EMA Scaling Rule (\Cref{def:ema-sr}). Bands indicate the mean and standard deviation across three runs.}
    \vspace{-0.2cm}
    \label{fig:r50-polyak}
\end{figure}

{\bf Image Classification}~~~
We consider a variant of the original \gls{sgd} Scaling Rule result \citep{DBLP:journals/corr/GoyalDGNWKTJH17} and train a ResNetv2 \citep{DBLP:conf/eccv/HeZRS16}
on ImageNet1k \citep{DBLP:journals/corr/RussakovskyDSKSMHKKBBF14} (\Cref{fig:r50-polyak}) using a three step learning rate schedule. 
The base momentum $\rho_B=0.9999$ at batch size 1024 was found by hyperparameter optimizing for \gls{ema} test performance, and we seek to achieve this optimized performance at different batch sizes.
We \emph{do not} apply the \gls{ema} Scaling Rule on the Batch Normalization \citep{DBLP:conf/icml/IoffeS15} statistics\footnote{Since Batch Normalization statistics use an \gls{ema} update, it is reasonable to ask whether the \gls{ema} Scaling Rule should be applied.
We investigate this in \Cref{subsec:polyak-bn}.
We find one \emph{should} apply the scaling rule, however, the effect is less significant than the application of the \gls{ema} Scaling Rule to model parameters.}. 
We observe that \emph{without} the EMA Scaling Rule, there is a significant drop in model \gls{ema} test performance, whereas \emph{with} the \gls{ema} Scaling Rule, we can approximate the baseline model \gls{ema} test top-1 performance across all batch sizes.
We match baseline \gls{ema} statistics across the full trajectory batch size 2048, where the test \gls{ema} performance diverges.
This is due to non-\gls{ema} test performance dropping for high $\kappa$ (see \Cref{app:subsec:polyak-image-classification}).
We observe that model \gls{ema} top-1 is approximately 0.2\% to 0.3\% higher than the target model.

{\bf \gls{asr}}~~~ 
We train a transformer \citep{DBLP:conf/nips/VaswaniSPUJGKP17}
using the \gls{ctc} loss~\citep{graves2006connectionist} and Adam optimizer on the \tco{} subset (100h) of LibriSpeech~\citep{panayotov2015librispeech} (for details see \Cref{app:speech}).
We apply the Adam Scaling Rule (\citet{DBLP:conf/nips/MalladiLPA22}, \Cref{def:adam-sr}) and use dynamic batching (minibatch size $\times$ sequence length $=\text{const}=290s$,
and $s$ indicates audio duration in seconds).

\begin{figure}[th]
    \centering
    \includegraphics[width=\textwidth]{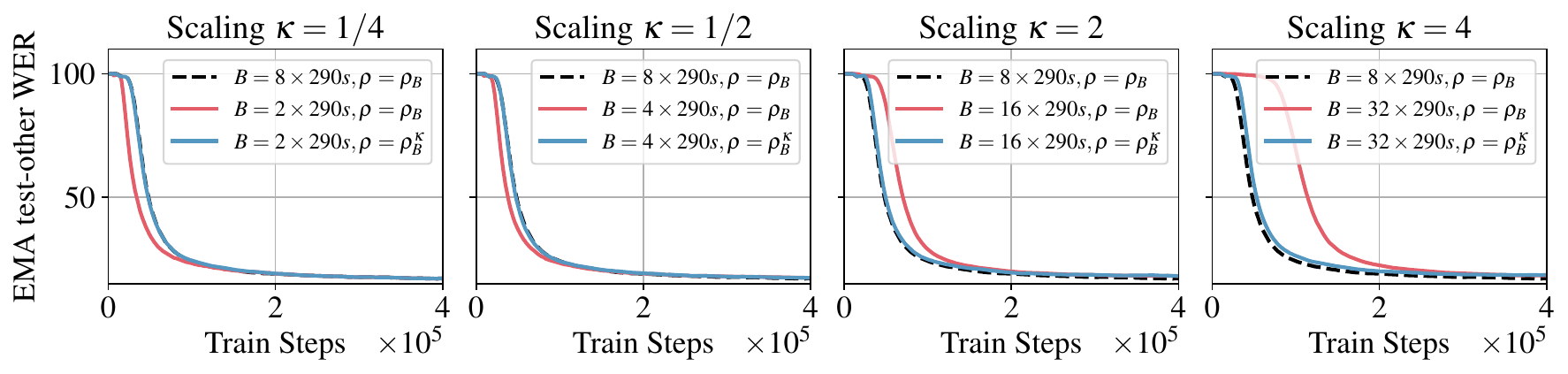}
    \caption{
    \emph{Transformer Polyak-Ruppert averaging on LibriSpeech (trained on \tco{})} with different scalings $\kappa$.
    The baseline ($\kappa=1$, black dashed)
    is trained with Adam and momentum $\rho_B=0.99995$ at a \emph{dynamic batch size} $B=8\times 290s$, which corresponds to a single train step on the $x$-axis. 
    We investigate dynamic batch sizes down to 
    $B=2\times 290s$ (left) and up to 
    $B=32\times 290s$ (right), 
    with (blue, $\rho=\rho_B^\kappa$), and without (red, $\rho=\rho_B$) the \gls{ema} Scaling Rule.
    The Adam Scaling Rule (\citet{DBLP:conf/nips/MalladiLPA22}, \Cref{def:adam-sr}) is used throughout. 
    }
    \label{fig:speech-polyak}
    \vspace{-0.5cm}
\end{figure}

\emph{Without} the EMA Scaling Rule, there is a significant difference in model \gls{ema} test \gls{wer} trajectories compared to the baseline, whereas \emph{with} the \gls{ema} Scaling Rule, trajectories match, as is shown in \Cref{fig:speech-polyak}. 
We note that compared to image classification, in \gls{asr}, the model \gls{ema} converges to similar final performance irrespective of use of the scaling rule. 
This convergence is due to the longer training time compared to the \gls{ema} horizon as discussed in \Cref{tab:different-ema} (see \Cref{app:asymptoticAnalysis} for a proof sketch).
Although in this specific case one can achieve similar \emph{final performance} without the \gls{ema} Scaling Rule, it is \emph{necessary} to use the \gls{ema} Scaling Rule in order to replicate the full training trajectory, which gives \emph{guarantees} on properties like final performance (see \Cref{cor:validity-scaling-rule}).
We also observe
a growing gap between the baseline and \gls{ema}-scaled trajectories as we increase~$\kappa$. 
Inspecting the train loss and non-EMA test WER, which \emph{do not} depend on the \gls{ema} update (see \Cref{fig:app-speech-polyak}, \Cref{subsec:speech-detailed}), indicates this is due to a breakdown of the Adam Scaling Rule.
\emph{In summary, evaluation on ASR shows that the EMA Scaling Rule holds in practice for sequential data with dynamic batch sizes, as well as when using adaptive optimization.}

\subsection{Semi-supervised speech recognition via pseudo-labeling}
\label{subsec:semi-supervised}

We continue using the same \gls{asr} model and training pipeline of~\Cref{subsec:supervised-polyakking}.
However, we consider semi-supervised learning via continuous pseudo-labeling where labeled (\tco{}, 100h) and unlabeled (the rest of LibriSpeech, 860h) data are given during training,
and the model \gls{ema} is involved in the overall optimization~\citep{likhomanenko2020slimipl,likhomanenko2022continuous, manohar2021kaizen,higuchi2022momentum}. 
We first pre-train a target model (\emph{student}) on a limited labeled set for a short period (e.g. 20k steps of $B=8\times 290s$\footnote{Note that number of steps is batch size dependent and should be scaled by $1/\kappa$ (see \Cref{app:scaling-toolbox}).}).
Concurrently, the student updates a model \gls{ema} (\emph{teacher}).
After pre-training,
we continue training the student with both labeled and unlabeled data,
with the teacher first transcribing unlabeled data from the batch producing
\glspl{pl}.
These \glspl{pl} are treated by the student as ground-truth transcriptions, and standard supervised optimization is performed.

\begin{figure}[t!]
    \centering
    \includegraphics[width=\textwidth]{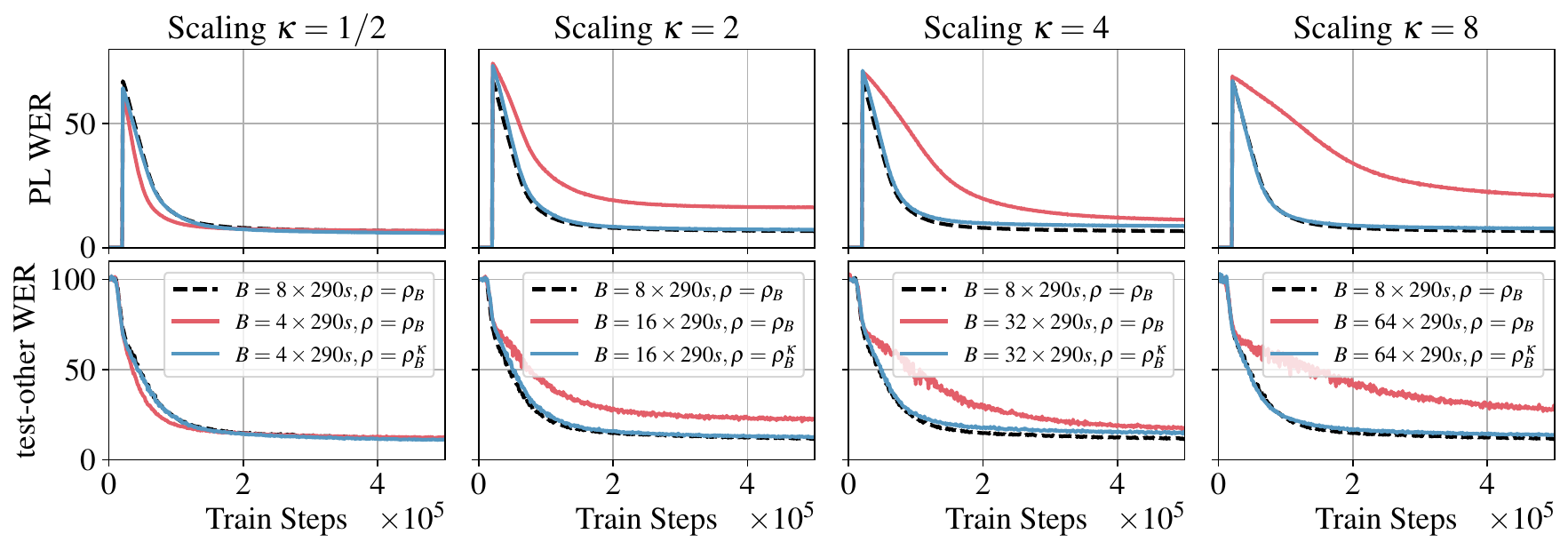}
    \caption{
    \emph{Transformer pseudo-labeling on LibriSpeech} with different scalings $\kappa$. 
    The baseline ($\kappa=1$, black dashed)
    is trained with Adam at a \emph{dynamic batch size} of $8\times 290$ seconds, which corresponds to a single train step on the $x$-axis.
    The model \gls{ema} (\emph{teacher}) is updated with momentum $\rho_B=0.9999$.
    We investigate dynamic batch sizes down to $B=4\times 290s$ (left) and up to $B=64\times 290s$ (right),
    with (blue, $\rho=\rho_B^\kappa$) and without (red, $\rho=\rho_B$) the \gls{ema} Scaling Rule.
    The Adam Scaling Rule (\citet{DBLP:conf/nips/MalladiLPA22}, \Cref{def:adam-sr}) is used throughout.
    For $\kappa\leq2$, we start pseudo-labeling after $20\text{k}/\kappa$ training steps; while for $\kappa>2$, we start when pre-training WER matches the baseline WER.
    }
    \label{fig:speech-pl-9999}
    \vspace{-0.6cm}
\end{figure}

Compared to Polyak-Ruppert Averaging (\Cref{subsec:supervised-polyakking}), where the model \gls{ema} plays no role in the joint optimization, 
we observe that in \gls{pl} it is \emph{essential} to employ the \gls{ema} Scaling Rule in order to match the model trajectories at scaled batch sizes.
When the \gls{ema} Scaling Rule is not used, \Cref{fig:speech-pl-9999} reveals a significant difference in  \gls{pl} quality trajectory, leading to a higher test \gls{wer}.

For $\kappa > 2$, we found the Adam Scaling Rule does not perfectly match the reference trajectory in the pre-training phase.
This results in a significantly different \gls{pl} quality
at the start of pseudo-labeling (20k steps of $B=8\times 290s$), which affects the training dynamics~\citep{berrebbi2023continuous}.
To alleviate the Adam Scaling Rule mismatch effect for $\kappa > 2$, we postpone the pseudo-labeling until pre-training on labeled data gives similar validation \gls{wer}, see Appendix~\ref{app:speech}. 
With this heuristic, we can match the baseline trajectory with the \gls{ema} Scaling Rule up to $\kappa=8$ (\Cref{fig:speech-pl-9999}).

\emph{In summary, (a) model EMA affects the optimization process of pseudo-labeling in ASR resulting in
the necessity of EMA Scaling Rule to be applied while scaling the batch size; (b) an optimizer scaling rule breakdown results in the EMA Scaling Rule breakdown but this effect can be alleviated by longer pre-training on labeled data having similar PLs quality at the start across different scalings.}

\subsection{Self-supervised image representation learning}
\label{subsec:self-supervised}

Finally, we turn our attention to  distillation based
\acrfull{ssl}.
where the model \gls{ema} is the \emph{teacher}
\citep{DBLP:conf/nips/GrillSATRBDPGAP20,DBLP:journals/taslp/NiizumiTOHK23,DBLP:journals/corr/abs-2104-14294,DBLP:journals/corr/abs-2304-07193}.

\begin{figure}[h!]
    \centering
    \includegraphics[width=0.99\textwidth]{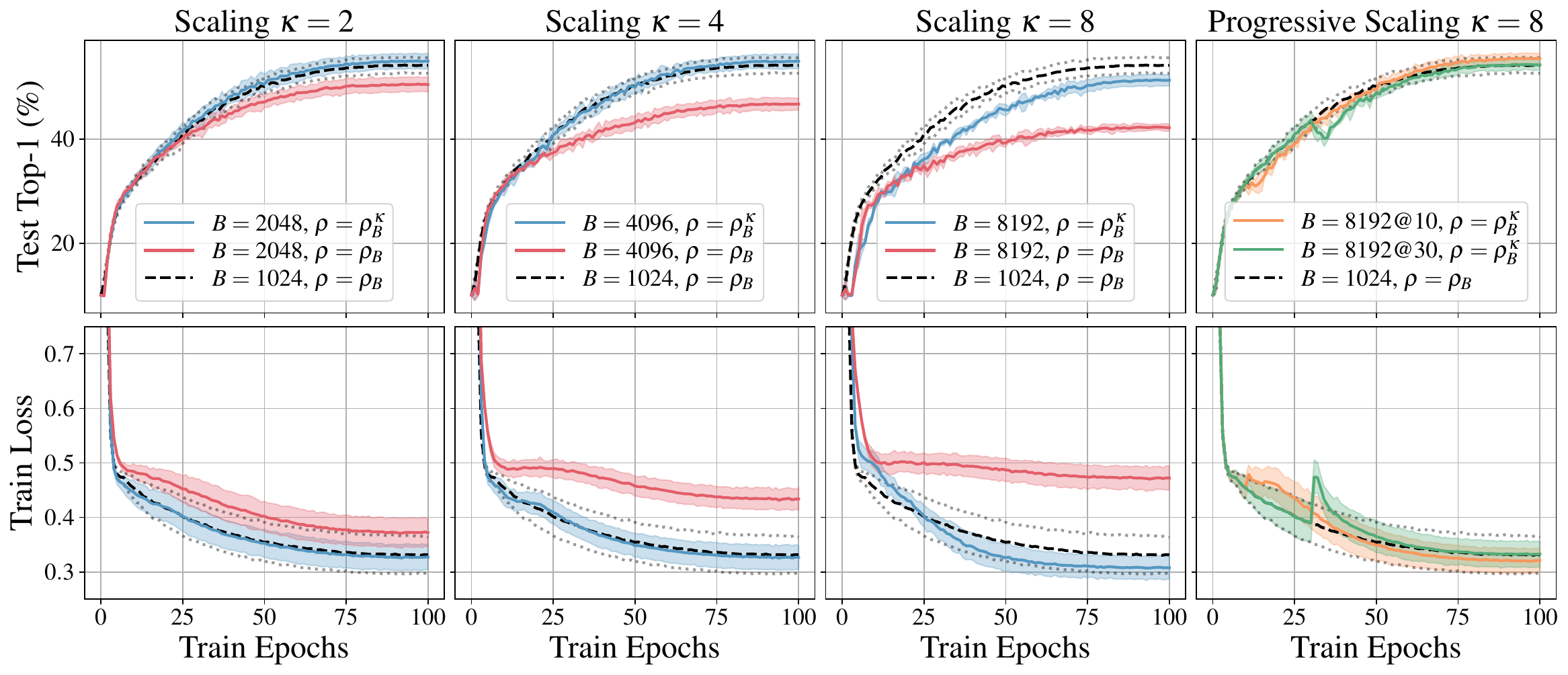}
    \caption{
    \emph{ResNet-18 BYOL on CIFAR10}
    for different $\kappa$.
    The baseline ($\kappa=1$, black dashed)
    uses batch size 1024 and momentum $\rho_B=0.992$,
    and is scaled from batch size 2048 (left) to 8192 (third) with (blue, $\rho=\rho_B^\kappa$) and without (red, $\rho=\rho_B$) the \gls{ema} Scaling Rule.
    At $\kappa=8$ we also run \emph{progressive scaling} (right), with transitions at 10 (green) and 30 (orange) epochs.
    Bands indicate mean and standard deviation across three runs. }
    \label{fig:r18-byol}
    \vspace{-0.2cm}
\end{figure}

We will use \gls{byol}
(\cite{DBLP:conf/nips/GrillSATRBDPGAP20}, \Cref{def:emaUpdateDefinition})\footnote{
The \gls{byol} \gls{ema} update (\Cref{eq:byol-ema-update}) uses $\rvtheta_{t+1}$ instead of our analyzed $\rvtheta_{t}$ (\Cref{eq:scalingRuleSummaryEquation}).
The effect upon the overall \gls{ema} update is $\mathcal{O}(\eta\times\beta_\rho)$ and so is captured by the \gls{ema} Scaling Rule (\Cref{def:ema-sr}).
}
for our investigation into scaling as it is well-studied \citep{DBLP:conf/icml/TianCG21,DBLP:journals/corr/abs-2302-04817}, relatively simple to implement due to minimal hyper-parameters, and obtains competitive results \citep{DBLP:conf/nips/GrillSATRBDPGAP20,DBLP:journals/corr/abs-2209-15589}.
Since \gls{byol} learns through self-referential distillation, momentum plays a significant role in optimization. 
We analyze: i) a ResNet-18 \citep{DBLP:conf/cvpr/HeZRS16} on CIFAR10 \citep{CIFAR10} (\Cref{fig:r18-byol}) using SGD (\Cref{def:sgd}); and ii) a \gls{vit}-B/16 \citep{DBLP:conf/iclr/DosovitskiyB0WZ21} on ImageNet1k using AdamW \citep{DBLP:conf/iclr/LoshchilovH19}.
A recipe for \gls{byol} using \glspl{vit} is provided in \Cref{app:byol-vit}.

\begin{figure}[h!]
    \centering
    \includegraphics[width=0.99\textwidth]{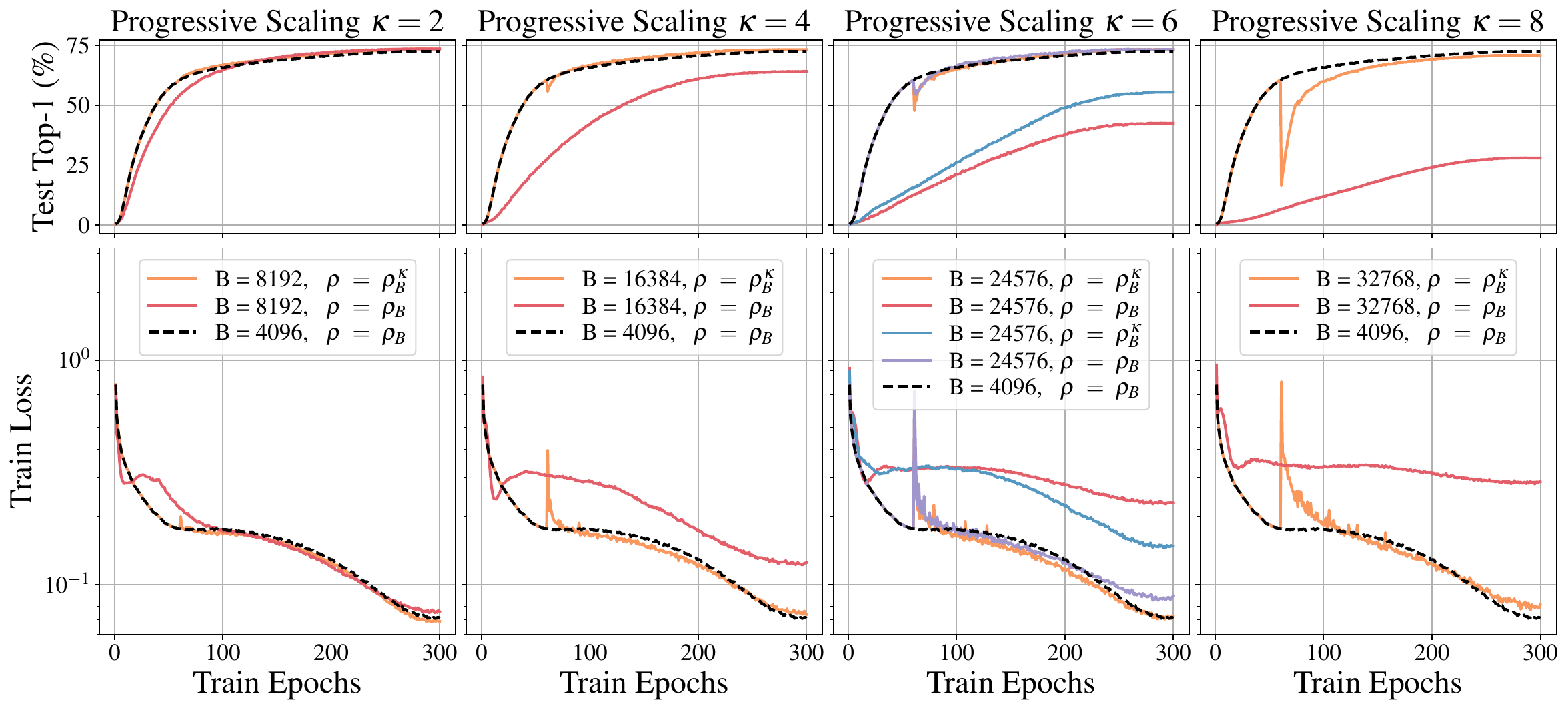}
    \caption{
    \emph{BYOL ViT-B/16 on ImageNet1k} for different scalings $\kappa$.
    The baseline model ($\kappa=1$, black dashed) uses batch size 4096 and teacher momentum
    $\rho_B=0.99$, and is scaled from batch size 8192 (left) to 32768 (right) with progressive scaling and the \gls{ema} Scaling Rule (\Cref{def:progressive-scaling}) (orange, $\rho=\rho_B^\kappa$), with the \gls{ema} Scaling Rule but without progressive scaling (blue, $\rho=\rho_B^\kappa$), without the \gls{ema} Scaling Rule but with progressive scaling (purple, $\rho=\rho_B$), and without either (red, $\rho=\rho_B$). Progressive scaling transitions from the reference model at epoch 60. See \Cref{app:byol-progressive-scaling-regimes} for a discussion on BYOL progressive scaling.
    }
    \label{fig:vitb-byol}
    \vspace{-0.2cm}
\end{figure}

{\bf ResNet-18 on CIFAR-10}~
We begin with a ResNet-18 model and short training duration to enable quick iteration,
and an \gls{sgd} optimizer as it has as \emph{known} scaling rule. 
This allows us to probe the \gls{ema} Scaling Rule without potential confounders like poor gradient-based optimizer scaling\footnote{For competitive performance with the reference \gls{byol}
\citep{DBLP:conf/nips/GrillSATRBDPGAP20}
using a ResNet-50, adaptive optimization, and longer training duration, see 
\Cref{subsec:byol-additional}
and
\Cref{fig:r50-byol}.}. 

We observe that \emph{without} the \gls{ema} Scaling Rule, there is a  drop in test top-1 linear probe (\Cref{def:linear-probe}) performance compared to the baseline, whereas \emph{with} the \gls{ema} Scaling Rule, we closely match the baseline model until batch size 4096.
We show that this result is consistent for a range of base learning rates $\eta_B$ and momenta $\rho_B$ in \Cref{subsec:byol-sensitivity-analysis}.
At batch size 8192, we see a performance gap between the scaled model using the \gls{ema} Scaling Rule and the baseline.
We speculate that this is due to dynamics early in the \gls{byol} training process that are challenging to replicate at larger batch sizes.
To test, and potentially circumvent this, we introduce \emph{Progressive Scaling} (\Cref{def:progressive-scaling}).
\begin{definition}[Progressive Scaling, informal; see \Cref{subsec:dynamic-batch-scaling}] 
    Given batch size $B$ and hyperparameters at $B$, 
    slowly increase the batch size to the desired largest batch size during training.
    At any intermediate batch size $\hat B=\kappa B$, all hyperparameters are scaled according to their scaling rules.
    \label{def:progressive-scaling}
\end{definition}
We see that transitioning to the higher batch size \emph{during} the warmup period results in a model optimization trajectory that diverges from the baseline, whereas transitioning \emph{after} warmup results in matching final trajectories of the scaled and baseline models.
In summary, \emph{progressive scaling} allows us to match \gls{byol} dynamics at large batch sizes, provided we transition after the warmup period.
This observation is consistent with our hypothesis regarding \gls{byol} dynamics during warmup.

{\bf Vision Transformers on ImageNet1k}~
\label{subsec:vit_byol}
Progressive Scaling coupled with the \gls{ema} Scaling Rule is required when scaling \gls{byol} \glspl{vit} (\Cref{fig:vitb-byol}),
enabling baseline loss tracking to a batch size of 24,576. 
Perfect scaling fails at batch size 32,768,
 consistent with observations in supervised learning 
\citep{DBLP:journals/corr/GoyalDGNWKTJH17,DBLP:conf/aaai/HuoGH21}.
Despite the breakdown, there is only a small drop in 1.6\% probe performance when using the \gls{ema} Scaling Rule, compared to as 44.56\% drop \emph{without} it.
We also observe that it is sometimes possible to match test model performance using \emph{only} Progressive Scaling and \emph{not} the \gls{ema} Scaling Rule, although this still induces a training loss mismatch.
We stress that such an approach is \emph{not} guaranteed to work and discuss when this approach succeeds and fails in \Cref{app:byol-progressive-scaling-regimes} and \Cref{fig:robustness-cartoon}.

At the transition point between batch sizes, 
an impulse perturbation\footnote{Instead of a single large batch transition as in \Cref{fig:vitb-byol} we perform a sequential transition in \Cref{app:byol-waterfall}. We find that a slow increase in batch size minimizes the magnitude of the perturbation and leads to a final model with higher effective linear probe top-1 than the reference by approximately $1.17\%$.} is measured at the student, visible from the training loss. 
This 
is recovered from by the learning process, 
and the new model matches the reference batch size. 
This perturbation happens in both the AdamW and \gls{sgd} settings, 
leading us to 
suspect
this is due to the \gls{byol} learning process, rather than an artifact of optimizer or momentum scaling. However, since this is not directly related to the EMA Scaling Rule proposed in this work, we defer this analysis to future investigation.

\section{Related work}
\label{sec:related-work}

{\bf Optimizer scaling rules from \glspl{sde}}~
The \gls{sde} perspective has uncovered optimizer scaling rules and allowed an understanding of their limitations.
\citet{DBLP:conf/iclr/SmithL18} used \glspl{sde} to uncover the \gls{sgd} Scaling Rule, while 
\citep{DBLP:conf/nips/LiMA21} used \glspl{sde} to explain that rule's breakdown in terms of discretization error.
The \gls{sde} analysis was extended to adaptive optimization by
\citep{DBLP:conf/nips/MalladiLPA22}, producing an Adam Scaling Rule (\Cref{def:adam-sr}),
indicating that along with the learning rate, the $\beta_{1,2}$ and $\epsilon$ parameters transform.
The $\beta_{1,2}$ transformation is consistent with the \gls{ema} Scaling Rule in the \gls{sde} limit.
Our work differs as it considers a model EMA that alters the objective.

{\bf Varying the batch size during training}~
\citet{DBLP:conf/iclr/SmithKYL18}
investigated the benefits of scheduling the batch size at a fixed learning rate as an alternative to scheduling the learning rate at a fixed batch size.
These two are equivalent through the \gls{sgd} Scaling Rule.
The authors \emph{do not} scale the optimizer hyperparameters during this procedure, as they are intentionally replicating the training dynamics of a learning rate schedule.
This is in contrast with \emph{Progressive Scaling} (\Cref{def:progressive-scaling}) which scales the hyperparameters to \emph{maintain} the optimization process at different levels of discretization.

{\bf Large batch training of SSL distillation methods} 
\gls{ssl} methods learn representations without labels, meaning they can take advantage of web-scale data.
Large batch optimization is required to make use of this data in a reasonable amount of time.
\citet{DBLP:conf/nips/GrillSATRBDPGAP20} demonstrated algorithmic robustness when \emph{reducing} the batch size through gradient accumulation and EMA update skipping, which implements an approximation of our \gls{ema} Scaling Rule for $\kappa<1$.
Our work provides a recipe to scale down \emph{and up} in $\kappa$. 
MoCo-v3 \citep{DBLP:conf/iccv/ChenXH21} enables contrastively distilled \gls{vit}s up to a batch size of 6144, where the model drops in performance. 
More recently, methods like DINO \citep{DBLP:conf/nips/CaronMMGBJ20} present a worse scenario, and are unable to scale beyond batch size 1024 \citep{DBLP:journals/corr/abs-2209-15589}.
In contrast, our work presents practical tools to scale to large batch sizes in the presence of an \gls{ema}, enabling practical training of these \gls{ssl} methods on large scale data.

\section{Conclusion}
\label{sec:conclusion}

We provide an \gls{ema} Scaling Rule: when changing the batch size by a factor of $\kappa$,  exponentiate the momentum of the \gls{ema} update to the power of $\kappa$.
This scaling rule should be applied in addition to optimizer scaling rules
(for example, linearly scaling the SGD learning rate),
and
enables the scaling of methods which rely on \gls{ema} and are sensitive to the choice of \gls{ema} momentum.

We prove the validity of the \gls{ema} Scaling Rule by deriving 
first-order \gls{sde} approximations of discrete model optimization when a model \gls{ema} is present and can contribute to the model objective.
We demonstrate empirical support 
for a variety of uses of \gls{ema}, ordered by increasing influence of the role of \gls{ema} on the optimization procedure: supervised model tracking (i.e. Polyak-Ruppert averaging) in speech and vision domains, pseudo-labeling in speech, and self-supervised image representation learning.
In almost all scenarios, using the \gls{ema} Scaling Rule
enables matching of training dynamics under batch size modification, whereas not using it results in significant differences in optimization trajectories. 
For example, we can scale the \gls{byol} self-supervised method to a batch size of 24,576 without any performance loss \emph{only} when using the \gls{ema} Scaling Rule.

While learning rate scaling rules are relatively commonplace in \gls{ml}, 
the role of \gls{ema} has been overlooked.
With this work, 
we highlight the importance of scaling the \gls{ema} momentum,
and hope that future works will use the \gls{ema} Scaling Rule to scale the \gls{ema} momentum correctly, in the same way that learning rates and other optimizer hyperparameters are scaled.

\ifthenelse{\equal{\anonymous}{0}}{\section{Acknowledgements}
\label{sec:acknowledgements}

We thank
Miguel Sarabia del Castillo,
Adam Golinski,
Pau Rodriguez Lopez,
Skyler Seto,
Amitis Shidani,
Barry Theobald,
Vimal Thilak,
Floris Weers,
Luca Zappella, and
Shaungfei Zhai
for their helpful feedback and critical discussions throughout the process of writing this paper;
Okan Akalin,
Hassan Babaie, 
Denise Hui,
Mubarak Seyed Ibrahim, 
Li Li, 
Cindy Liu, 
Rajat Phull,
Evan Samanas, 
Guillaume Seguin, 
and the wider Apple infrastructure team for assistance with developing and running scalable, fault tolerant code; 
and 
Kaifeng Lyu and
Abhishek Panigrahi
for discussion and details regarding scaling rules for adaptive optimizers.
Names are in alphabetical order by last name within group.
}{}

\bibliography{main.bbl}

\begin{thebibliography}{68}
\providecommand{\natexlab}[1]{#1}
\providecommand{\url}[1]{\texttt{#1}}
\expandafter\ifx\csname urlstyle\endcsname\relax
  \providecommand{\doi}[1]{doi: #1}\else
  \providecommand{\doi}{doi: \begingroup \urlstyle{rm}\Url}\fi

\bibitem[Ba et~al.(2016)Ba, Kiros, and Hinton]{DBLP:journals/corr/BaKH16}
Lei~Jimmy Ba, Jamie~Ryan Kiros, and Geoffrey~E. Hinton.
\newblock Layer normalization.
\newblock \emph{CoRR}, abs/1607.06450, 2016.
\newblock URL \url{http://arxiv.org/abs/1607.06450}.

\bibitem[Baevski et~al.(2022{\natexlab{a}})Baevski, Babu, Hsu, and
  Auli]{DBLP:journals/corr/abs-2212-07525}
Alexei Baevski, Arun Babu, Wei{-}Ning Hsu, and Michael Auli.
\newblock Efficient self-supervised learning with contextualized target
  representations for vision, speech and language.
\newblock \emph{CoRR}, abs/2212.07525, 2022{\natexlab{a}}.
\newblock \doi{10.48550/arXiv.2212.07525}.
\newblock URL \url{https://doi.org/10.48550/arXiv.2212.07525}.

\bibitem[Baevski et~al.(2022{\natexlab{b}})Baevski, Hsu, Xu, Babu, Gu, and
  Auli]{baevski2022data2vec}
Alexei Baevski, Wei-Ning Hsu, Qiantong Xu, Arun Babu, Jiatao Gu, and Michael
  Auli.
\newblock Data2vec: A general framework for self-supervised learning in speech,
  vision and language.
\newblock In \emph{International Conference on Machine Learning}, pp.\
  1298--1312. PMLR, 2022{\natexlab{b}}.

\bibitem[Berrebbi et~al.(2023)Berrebbi, Collobert, Bengio, Jaitly, and
  Likhomanenko]{berrebbi2023continuous}
Dan Berrebbi, Ronan Collobert, Samy Bengio, Navdeep Jaitly, and Tatiana
  Likhomanenko.
\newblock Continuous pseudo-labeling from the start.
\newblock In \emph{The Eleventh International Conference on Learning
  Representations}, 2023.
\newblock URL \url{https://openreview.net/forum?id=m3twGT2bAug}.

\bibitem[Bommasani et~al.(2021)Bommasani, Hudson, Adeli, Altman, Arora, von
  Arx, Bernstein, Bohg, Bosselut, Brunskill, Brynjolfsson, Buch, Card,
  Castellon, Chatterji, Chen, Creel, Davis, Demszky, Donahue, Doumbouya,
  Durmus, Ermon, Etchemendy, Ethayarajh, Fei{-}Fei, Finn, Gale, Gillespie,
  Goel, Goodman, Grossman, Guha, Hashimoto, Henderson, Hewitt, Ho, Hong, Hsu,
  Huang, Icard, Jain, Jurafsky, Kalluri, Karamcheti, Keeling, Khani, Khattab,
  Koh, Krass, Krishna, Kuditipudi, and
  et~al.]{DBLP:journals/corr/abs-2108-07258}
Rishi Bommasani, Drew~A. Hudson, Ehsan Adeli, Russ~B. Altman, Simran Arora,
  Sydney von Arx, Michael~S. Bernstein, Jeannette Bohg, Antoine Bosselut, Emma
  Brunskill, Erik Brynjolfsson, Shyamal Buch, Dallas Card, Rodrigo Castellon,
  Niladri~S. Chatterji, Annie~S. Chen, Kathleen Creel, Jared~Quincy Davis,
  Dorottya Demszky, Chris Donahue, Moussa Doumbouya, Esin Durmus, Stefano
  Ermon, John Etchemendy, Kawin Ethayarajh, Li~Fei{-}Fei, Chelsea Finn, Trevor
  Gale, Lauren Gillespie, Karan Goel, Noah~D. Goodman, Shelby Grossman, Neel
  Guha, Tatsunori Hashimoto, Peter Henderson, John Hewitt, Daniel~E. Ho, Jenny
  Hong, Kyle Hsu, Jing Huang, Thomas Icard, Saahil Jain, Dan Jurafsky,
  Pratyusha Kalluri, Siddharth Karamcheti, Geoff Keeling, Fereshte Khani, Omar
  Khattab, Pang~Wei Koh, Mark~S. Krass, Ranjay Krishna, Rohith Kuditipudi, and
  et~al.
\newblock On the opportunities and risks of foundation models.
\newblock \emph{CoRR}, abs/2108.07258, 2021.
\newblock URL \url{https://arxiv.org/abs/2108.07258}.

\bibitem[Brock et~al.(2021)Brock, De, Smith, and
  Simonyan]{DBLP:conf/icml/BrockDSS21}
Andy Brock, Soham De, Samuel~L. Smith, and Karen Simonyan.
\newblock High-performance large-scale image recognition without normalization.
\newblock In Marina Meila and Tong Zhang (eds.), \emph{Proceedings of the 38th
  International Conference on Machine Learning, {ICML} 2021, 18-24 July 2021,
  Virtual Event}, volume 139 of \emph{Proceedings of Machine Learning
  Research}, pp.\  1059--1071. {PMLR}, 2021.
\newblock URL \url{http://proceedings.mlr.press/v139/brock21a.html}.

\bibitem[Caron et~al.(2020)Caron, Misra, Mairal, Goyal, Bojanowski, and
  Joulin]{DBLP:conf/nips/CaronMMGBJ20}
Mathilde Caron, Ishan Misra, Julien Mairal, Priya Goyal, Piotr Bojanowski, and
  Armand Joulin.
\newblock Unsupervised learning of visual features by contrasting cluster
  assignments.
\newblock In Hugo Larochelle, Marc'Aurelio Ranzato, Raia Hadsell,
  Maria{-}Florina Balcan, and Hsuan{-}Tien Lin (eds.), \emph{Advances in Neural
  Information Processing Systems 33: Annual Conference on Neural Information
  Processing Systems 2020, NeurIPS 2020, December 6-12, 2020, virtual}, 2020.
\newblock URL
  \url{https://proceedings.neurips.cc/paper/2020/hash/70feb62b69f16e0238f741fab228fec2-Abstract.html}.

\bibitem[Caron et~al.(2021)Caron, Touvron, Misra, J{\'{e}}gou, Mairal,
  Bojanowski, and Joulin]{DBLP:journals/corr/abs-2104-14294}
Mathilde Caron, Hugo Touvron, Ishan Misra, Herv{\'{e}} J{\'{e}}gou, Julien
  Mairal, Piotr Bojanowski, and Armand Joulin.
\newblock Emerging properties in self-supervised vision transformers.
\newblock \emph{CoRR}, abs/2104.14294, 2021.
\newblock URL \url{https://arxiv.org/abs/2104.14294}.

\bibitem[Chen et~al.(2020)Chen, Kornblith, Swersky, Norouzi, and
  Hinton]{DBLP:conf/nips/ChenKSNH20}
Ting Chen, Simon Kornblith, Kevin Swersky, Mohammad Norouzi, and Geoffrey~E.
  Hinton.
\newblock Big self-supervised models are strong semi-supervised learners.
\newblock In Hugo Larochelle, Marc'Aurelio Ranzato, Raia Hadsell,
  Maria{-}Florina Balcan, and Hsuan{-}Tien Lin (eds.), \emph{Advances in Neural
  Information Processing Systems 33: Annual Conference on Neural Information
  Processing Systems 2020, NeurIPS 2020, December 6-12, 2020, virtual}, 2020.
\newblock URL
  \url{https://proceedings.neurips.cc/paper/2020/hash/fcbc95ccdd551da181207c0c1400c655-Abstract.html}.

\bibitem[Chen et~al.(2021)Chen, Xie, and He]{DBLP:conf/iccv/ChenXH21}
Xinlei Chen, Saining Xie, and Kaiming He.
\newblock An empirical study of training self-supervised vision transformers.
\newblock In \emph{2021 {IEEE/CVF} International Conference on Computer Vision,
  {ICCV} 2021, Montreal, QC, Canada, October 10-17, 2021}, pp.\  9620--9629.
  {IEEE}, 2021.
\newblock \doi{10.1109/ICCV48922.2021.00950}.
\newblock URL \url{https://doi.org/10.1109/ICCV48922.2021.00950}.

\bibitem[Dosovitskiy et~al.(2021)Dosovitskiy, Beyer, Kolesnikov, Weissenborn,
  Zhai, Unterthiner, Dehghani, Minderer, Heigold, Gelly, Uszkoreit, and
  Houlsby]{DBLP:conf/iclr/DosovitskiyB0WZ21}
Alexey Dosovitskiy, Lucas Beyer, Alexander Kolesnikov, Dirk Weissenborn,
  Xiaohua Zhai, Thomas Unterthiner, Mostafa Dehghani, Matthias Minderer, Georg
  Heigold, Sylvain Gelly, Jakob Uszkoreit, and Neil Houlsby.
\newblock An image is worth 16x16 words: Transformers for image recognition at
  scale.
\newblock In \emph{9th International Conference on Learning Representations,
  {ICLR} 2021, Virtual Event, Austria, May 3-7, 2021}. OpenReview.net, 2021.
\newblock URL \url{https://openreview.net/forum?id=YicbFdNTTy}.

\bibitem[Duchi et~al.(2010)Duchi, Hazan, and Singer]{DBLP:conf/colt/DuchiHS10}
John~C. Duchi, Elad Hazan, and Yoram Singer.
\newblock Adaptive subgradient methods for online learning and stochastic
  optimization.
\newblock In Adam~Tauman Kalai and Mehryar Mohri (eds.), \emph{{COLT} 2010 -
  The 23rd Conference on Learning Theory, Haifa, Israel, June 27-29, 2010},
  pp.\  257--269. Omnipress, 2010.
\newblock URL
  \url{http://colt2010.haifa.il.ibm.com/papers/COLT2010proceedings.pdf\#page=265}.

\bibitem[Fetterman \& Albrecht(2020)Fetterman and
  Albrecht]{Fetterman_Albrecht_2020}
Abe Fetterman and Josh Albrecht.
\newblock Understanding self-supervised and contrastive learning with
  "bootstrap your own latent" (byol), Aug 2020.
\newblock URL
  \url{https://generallyintelligent.ai/understanding-self-supervised-contrastive-learning.html}.

\bibitem[Goyal et~al.(2017)Goyal, Doll{\'{a}}r, Girshick, Noordhuis,
  Wesolowski, Kyrola, Tulloch, Jia, and He]{DBLP:journals/corr/GoyalDGNWKTJH17}
Priya Goyal, Piotr Doll{\'{a}}r, Ross~B. Girshick, Pieter Noordhuis, Lukasz
  Wesolowski, Aapo Kyrola, Andrew Tulloch, Yangqing Jia, and Kaiming He.
\newblock Accurate, large minibatch {SGD:} training imagenet in 1 hour.
\newblock \emph{CoRR}, abs/1706.02677, 2017.
\newblock URL \url{http://arxiv.org/abs/1706.02677}.

\bibitem[Graves et~al.(2006)Graves, Fern{\'a}ndez, Gomez, and
  Schmidhuber]{graves2006connectionist}
Alex Graves, Santiago Fern{\'a}ndez, Faustino Gomez, and J{\"u}rgen
  Schmidhuber.
\newblock Connectionist temporal classification: labelling unsegmented sequence
  data with recurrent neural networks.
\newblock In \emph{Proceedings of the 23rd international conference on Machine
  learning}, pp.\  369--376, 2006.

\bibitem[Grill et~al.(2020)Grill, Strub, Altch{\'{e}}, Tallec, Richemond,
  Buchatskaya, Doersch, Pires, Guo, Azar, Piot, Kavukcuoglu, Munos, and
  Valko]{DBLP:conf/nips/GrillSATRBDPGAP20}
Jean{-}Bastien Grill, Florian Strub, Florent Altch{\'{e}}, Corentin Tallec,
  Pierre~H. Richemond, Elena Buchatskaya, Carl Doersch, Bernardo~{\'{A}}vila
  Pires, Zhaohan Guo, Mohammad~Gheshlaghi Azar, Bilal Piot, Koray Kavukcuoglu,
  R{\'{e}}mi Munos, and Michal Valko.
\newblock Bootstrap your own latent - {A} new approach to self-supervised
  learning.
\newblock In Hugo Larochelle, Marc'Aurelio Ranzato, Raia Hadsell,
  Maria{-}Florina Balcan, and Hsuan{-}Tien Lin (eds.), \emph{Advances in Neural
  Information Processing Systems 33: Annual Conference on Neural Information
  Processing Systems 2020, NeurIPS 2020, December 6-12, 2020, virtual}, 2020.
\newblock URL
  \url{https://proceedings.neurips.cc/paper/2020/hash/f3ada80d5c4ee70142b17b8192b2958e-Abstract.html}.

\bibitem[He et~al.(2020)He, Gu, Shen, and Ranzato]{He2020Revisiting}
Junxian He, Jiatao Gu, Jiajun Shen, and Marc'Aurelio Ranzato.
\newblock Revisiting self-training for neural sequence generation.
\newblock In \emph{International Conference on Learning Representations}, 2020.
\newblock URL \url{https://openreview.net/forum?id=SJgdnAVKDH}.

\bibitem[He et~al.(2015)He, Zhang, Ren, and Sun]{DBLP:conf/iccv/HeZRS15}
Kaiming He, Xiangyu Zhang, Shaoqing Ren, and Jian Sun.
\newblock Delving deep into rectifiers: Surpassing human-level performance on
  imagenet classification.
\newblock In \emph{2015 {IEEE} International Conference on Computer Vision,
  {ICCV} 2015, Santiago, Chile, December 7-13, 2015}, pp.\  1026--1034. {IEEE}
  Computer Society, 2015.
\newblock \doi{10.1109/ICCV.2015.123}.
\newblock URL \url{https://doi.org/10.1109/ICCV.2015.123}.

\bibitem[He et~al.(2016{\natexlab{a}})He, Zhang, Ren, and
  Sun]{DBLP:conf/cvpr/HeZRS16}
Kaiming He, Xiangyu Zhang, Shaoqing Ren, and Jian Sun.
\newblock Deep residual learning for image recognition.
\newblock In \emph{2016 {IEEE} Conference on Computer Vision and Pattern
  Recognition, {CVPR} 2016, Las Vegas, NV, USA, June 27-30, 2016}, pp.\
  770--778. {IEEE} Computer Society, 2016{\natexlab{a}}.
\newblock \doi{10.1109/CVPR.2016.90}.
\newblock URL \url{https://doi.org/10.1109/CVPR.2016.90}.

\bibitem[He et~al.(2016{\natexlab{b}})He, Zhang, Ren, and
  Sun]{DBLP:conf/eccv/HeZRS16}
Kaiming He, Xiangyu Zhang, Shaoqing Ren, and Jian Sun.
\newblock Identity mappings in deep residual networks.
\newblock In Bastian Leibe, Jiri Matas, Nicu Sebe, and Max Welling (eds.),
  \emph{Computer Vision - {ECCV} 2016 - 14th European Conference, Amsterdam,
  The Netherlands, October 11-14, 2016, Proceedings, Part {IV}}, volume 9908 of
  \emph{Lecture Notes in Computer Science}, pp.\  630--645. Springer,
  2016{\natexlab{b}}.
\newblock \doi{10.1007/978-3-319-46493-0\_38}.
\newblock URL \url{https://doi.org/10.1007/978-3-319-46493-0\_38}.

\bibitem[He et~al.(2022)He, Chen, Xie, Li, Doll{\'{a}}r, and
  Girshick]{DBLP:conf/cvpr/HeCXLDG22}
Kaiming He, Xinlei Chen, Saining Xie, Yanghao Li, Piotr Doll{\'{a}}r, and
  Ross~B. Girshick.
\newblock Masked autoencoders are scalable vision learners.
\newblock In \emph{{IEEE/CVF} Conference on Computer Vision and Pattern
  Recognition, {CVPR} 2022, New Orleans, LA, USA, June 18-24, 2022}, pp.\
  15979--15988. {IEEE}, 2022.
\newblock \doi{10.1109/CVPR52688.2022.01553}.
\newblock URL \url{https://doi.org/10.1109/CVPR52688.2022.01553}.

\bibitem[Higuchi et~al.(2022)Higuchi, Moritz, Le~Roux, and
  Hori]{higuchi2022momentum}
Yosuke Higuchi, Niko Moritz, Jonathan Le~Roux, and Takaaki Hori.
\newblock Momentum pseudo-labeling: Semi-supervised asr with continuously
  improving pseudo-labels.
\newblock \emph{IEEE Journal of Selected Topics in Signal Processing},
  16\penalty0 (6):\penalty0 1424--1438, 2022.

\bibitem[Huang et~al.(2017)Huang, Li, Pleiss, Liu, Hopcroft, and
  Weinberger]{DBLP:conf/iclr/HuangLP0HW17}
Gao Huang, Yixuan Li, Geoff Pleiss, Zhuang Liu, John~E. Hopcroft, and Kilian~Q.
  Weinberger.
\newblock Snapshot ensembles: Train 1, get {M} for free.
\newblock In \emph{5th International Conference on Learning Representations,
  {ICLR} 2017, Toulon, France, April 24-26, 2017, Conference Track
  Proceedings}. OpenReview.net, 2017.
\newblock URL \url{https://openreview.net/forum?id=BJYwwY9ll}.

\bibitem[Huo et~al.(2021)Huo, Gu, and Huang]{DBLP:conf/aaai/HuoGH21}
Zhouyuan Huo, Bin Gu, and Heng Huang.
\newblock Large batch optimization for deep learning using new complete
  layer-wise adaptive rate scaling.
\newblock In \emph{Thirty-Fifth {AAAI} Conference on Artificial Intelligence,
  {AAAI} 2021, Thirty-Third Conference on Innovative Applications of Artificial
  Intelligence, {IAAI} 2021, The Eleventh Symposium on Educational Advances in
  Artificial Intelligence, {EAAI} 2021, Virtual Event, February 2-9, 2021},
  pp.\  7883--7890. {AAAI} Press, 2021.
\newblock URL \url{https://ojs.aaai.org/index.php/AAAI/article/view/16962}.

\bibitem[IEEE(2019)]{iee}
IEEE.
\newblock Ieee standard for floating-point arithmetic.
\newblock \emph{IEEE Std 754-2019 (Revision of IEEE 754-2008)}, pp.\  1--84,
  2019.
\newblock \doi{10.1109/IEEESTD.2019.8766229}.

\bibitem[Ioffe \& Szegedy(2015)Ioffe and Szegedy]{DBLP:conf/icml/IoffeS15}
Sergey Ioffe and Christian Szegedy.
\newblock Batch normalization: Accelerating deep network training by reducing
  internal covariate shift.
\newblock In Francis~R. Bach and David~M. Blei (eds.), \emph{Proceedings of the
  32nd International Conference on Machine Learning, {ICML} 2015, Lille,
  France, 6-11 July 2015}, volume~37 of \emph{{JMLR} Workshop and Conference
  Proceedings}, pp.\  448--456. JMLR.org, 2015.
\newblock URL \url{http://proceedings.mlr.press/v37/ioffe15.html}.

\bibitem[Izmailov et~al.(2018)Izmailov, Podoprikhin, Garipov, Vetrov, and
  Wilson]{DBLP:conf/uai/IzmailovPGVW18}
Pavel Izmailov, Dmitrii Podoprikhin, Timur Garipov, Dmitry~P. Vetrov, and
  Andrew~Gordon Wilson.
\newblock Averaging weights leads to wider optima and better generalization.
\newblock In Amir Globerson and Ricardo Silva (eds.), \emph{Proceedings of the
  Thirty-Fourth Conference on Uncertainty in Artificial Intelligence, {UAI}
  2018, Monterey, California, USA, August 6-10, 2018}, pp.\  876--885. {AUAI}
  Press, 2018.
\newblock URL \url{http://auai.org/uai2018/proceedings/papers/313.pdf}.

\bibitem[Jastrzebski et~al.(2017)Jastrzebski, Kenton, Arpit, Ballas, Fischer,
  Bengio, and Storkey]{DBLP:journals/corr/abs-1711-04623}
Stanislaw Jastrzebski, Zachary Kenton, Devansh Arpit, Nicolas Ballas, Asja
  Fischer, Yoshua Bengio, and Amos~J. Storkey.
\newblock Three factors influencing minima in {SGD}.
\newblock \emph{CoRR}, abs/1711.04623, 2017.
\newblock URL \url{http://arxiv.org/abs/1711.04623}.

\bibitem[Kaplan et~al.(2020)Kaplan, McCandlish, Henighan, Brown, Chess, Child,
  Gray, Radford, Wu, and Amodei]{DBLP:journals/corr/abs-2001-08361}
Jared Kaplan, Sam McCandlish, Tom Henighan, Tom~B. Brown, Benjamin Chess, Rewon
  Child, Scott Gray, Alec Radford, Jeffrey Wu, and Dario Amodei.
\newblock Scaling laws for neural language models.
\newblock \emph{CoRR}, abs/2001.08361, 2020.
\newblock URL \url{https://arxiv.org/abs/2001.08361}.

\bibitem[Kingma \& Ba(2015)Kingma and Ba]{DBLP:journals/corr/KingmaB14}
Diederik~P. Kingma and Jimmy Ba.
\newblock Adam: {A} method for stochastic optimization.
\newblock In Yoshua Bengio and Yann LeCun (eds.), \emph{3rd International
  Conference on Learning Representations, {ICLR} 2015, San Diego, CA, USA, May
  7-9, 2015, Conference Track Proceedings}, 2015.
\newblock URL \url{http://arxiv.org/abs/1412.6980}.

\bibitem[Kish(1965)]{KishDesignEffect}
Leslie Kish.
\newblock \emph{Survey Sampling}, volume~59.
\newblock Cambridge University Press, 1965.
\newblock \doi{10.1017/S0003055400132113}.

\bibitem[Koppula et~al.(2022)Koppula, Li, Shelhamer, Jaegle, Parthasarathy,
  Arandjelovic, Carreira, and H{\'{e}}naff]{DBLP:journals/corr/abs-2209-15589}
Skanda Koppula, Yazhe Li, Evan Shelhamer, Andrew Jaegle, Nikhil Parthasarathy,
  Relja Arandjelovic, Jo{\~{a}}o Carreira, and Olivier~J. H{\'{e}}naff.
\newblock Where should {I} spend my flops? efficiency evaluations of visual
  pre-training methods.
\newblock \emph{CoRR}, abs/2209.15589, 2022.
\newblock \doi{10.48550/arXiv.2209.15589}.
\newblock URL \url{https://doi.org/10.48550/arXiv.2209.15589}.

\bibitem[Krizhevsky(2014)]{DBLP:journals/corr/Krizhevsky14}
Alex Krizhevsky.
\newblock One weird trick for parallelizing convolutional neural networks.
\newblock \emph{CoRR}, abs/1404.5997, 2014.
\newblock URL \url{http://arxiv.org/abs/1404.5997}.

\bibitem[Krizhevsky et~al.(2014)Krizhevsky, Nair, and Hinton]{CIFAR10}
Alex Krizhevsky, Vinod Nair, and Geoffrey Hinton.
\newblock Cifar-10 (canadian institute for advanced research).
\newblock 2014.
\newblock URL \url{http://www.cs.toronto.edu/~kriz/cifar.html}.

\bibitem[Li et~al.(2018)Li, Xu, Taylor, Studer, and
  Goldstein]{DBLP:conf/nips/Li0TSG18}
Hao Li, Zheng Xu, Gavin Taylor, Christoph Studer, and Tom Goldstein.
\newblock Visualizing the loss landscape of neural nets.
\newblock In Samy Bengio, Hanna~M. Wallach, Hugo Larochelle, Kristen Grauman,
  Nicol{\`{o}} Cesa{-}Bianchi, and Roman Garnett (eds.), \emph{Advances in
  Neural Information Processing Systems 31: Annual Conference on Neural
  Information Processing Systems 2018, NeurIPS 2018, December 3-8, 2018,
  Montr{\'{e}}al, Canada}, pp.\  6391--6401, 2018.
\newblock URL
  \url{https://proceedings.neurips.cc/paper/2018/hash/a41b3bb3e6b050b6c9067c67f663b915-Abstract.html}.

\bibitem[Li et~al.(2019)Li, Tai, and Weinan]{li2019stochastic}
Qianxiao Li, Cheng Tai, and E~Weinan.
\newblock Stochastic modified equations and dynamics of stochastic gradient
  algorithms i: Mathematical foundations.
\newblock \emph{The Journal of Machine Learning Research}, 20\penalty0
  (1):\penalty0 1474--1520, 2019.

\bibitem[Li et~al.(2021)Li, Malladi, and Arora]{DBLP:conf/nips/LiMA21}
Zhiyuan Li, Sadhika Malladi, and Sanjeev Arora.
\newblock On the validity of modeling {SGD} with stochastic differential
  equations (sdes).
\newblock In Marc'Aurelio Ranzato, Alina Beygelzimer, Yann~N. Dauphin, Percy
  Liang, and Jennifer~Wortman Vaughan (eds.), \emph{Advances in Neural
  Information Processing Systems 34: Annual Conference on Neural Information
  Processing Systems 2021, NeurIPS 2021, December 6-14, 2021, virtual}, pp.\
  12712--12725, 2021.
\newblock URL
  \url{https://proceedings.neurips.cc/paper/2021/hash/69f62956429865909921fa916d61c1f8-Abstract.html}.

\bibitem[Likhomanenko et~al.(2021{\natexlab{a}})Likhomanenko, Xu, Kahn,
  Synnaeve, and Collobert]{likhomanenko2020slimipl}
Tatiana Likhomanenko, Qiantong Xu, Jacob Kahn, Gabriel Synnaeve, and Ronan
  Collobert.
\newblock slimipl: Language-model-free iterative pseudo-labeling.
\newblock \emph{Proc. Interspeech}, 2021{\natexlab{a}}.

\bibitem[Likhomanenko et~al.(2021{\natexlab{b}})Likhomanenko, Xu, Synnaeve,
  Collobert, and Rogozhnikov]{likhomanenko2021cape}
Tatiana Likhomanenko, Qiantong Xu, Gabriel Synnaeve, Ronan Collobert, and Alex
  Rogozhnikov.
\newblock Cape: Encoding relative positions with continuous augmented
  positional embeddings.
\newblock \emph{Advances in Neural Information Processing Systems}, 34,
  2021{\natexlab{b}}.

\bibitem[Likhomanenko et~al.(2022)Likhomanenko, Collobert, Jaitly, and
  Bengio]{likhomanenko2022continuous}
Tatiana Likhomanenko, Ronan Collobert, Navdeep Jaitly, and Samy Bengio.
\newblock Continuous soft pseudo-labeling in {ASR}.
\newblock In \emph{I Can't Believe It's Not Better Workshop: Understanding Deep
  Learning Through Empirical Falsification}, 2022.
\newblock URL \url{https://openreview.net/forum?id=aoiqVW4ui51}.

\bibitem[Lillicrap et~al.(2016)Lillicrap, Hunt, Pritzel, Heess, Erez, Tassa,
  Silver, and Wierstra]{DBLP:journals/corr/LillicrapHPHETS15}
Timothy~P. Lillicrap, Jonathan~J. Hunt, Alexander Pritzel, Nicolas Heess, Tom
  Erez, Yuval Tassa, David Silver, and Daan Wierstra.
\newblock Continuous control with deep reinforcement learning.
\newblock In Yoshua Bengio and Yann LeCun (eds.), \emph{4th International
  Conference on Learning Representations, {ICLR} 2016, San Juan, Puerto Rico,
  May 2-4, 2016, Conference Track Proceedings}, 2016.
\newblock URL \url{http://arxiv.org/abs/1509.02971}.

\bibitem[Loshchilov \& Hutter(2019)Loshchilov and
  Hutter]{DBLP:conf/iclr/LoshchilovH19}
Ilya Loshchilov and Frank Hutter.
\newblock Decoupled weight decay regularization.
\newblock In \emph{7th International Conference on Learning Representations,
  {ICLR} 2019, New Orleans, LA, USA, May 6-9, 2019}. OpenReview.net, 2019.
\newblock URL \url{https://openreview.net/forum?id=Bkg6RiCqY7}.

\bibitem[Malladi et~al.(2022)Malladi, Lyu, Panigrahi, and
  Arora]{DBLP:conf/nips/MalladiLPA22}
Sadhika Malladi, Kaifeng Lyu, Abhishek Panigrahi, and Sanjeev Arora.
\newblock On the sdes and scaling rules for adaptive gradient algorithms.
\newblock In \emph{NeurIPS}, 2022.
\newblock URL
  \url{http://papers.nips.cc/paper\_files/paper/2022/hash/32ac710102f0620d0f28d5d05a44fe08-Abstract-Conference.html}.

\bibitem[Manohar et~al.(2021)Manohar, Likhomanenko, Xu, Hsu, Collobert, Saraf,
  Zweig, and Mohamed]{manohar2021kaizen}
Vimal Manohar, Tatiana Likhomanenko, Qiantong Xu, Wei-Ning Hsu, Ronan
  Collobert, Yatharth Saraf, Geoffrey Zweig, and Abdelrahman Mohamed.
\newblock Kaizen: Continuously improving teacher using exponential moving
  average for semi-supervised speech recognition.
\newblock In \emph{2021 IEEE Automatic Speech Recognition and Understanding
  Workshop (ASRU)}, pp.\  518--525. IEEE, 2021.

\bibitem[Niizumi et~al.(2023)Niizumi, Takeuchi, Ohishi, Harada, and
  Kashino]{DBLP:journals/taslp/NiizumiTOHK23}
Daisuke Niizumi, Daiki Takeuchi, Yasunori Ohishi, Noboru Harada, and Kunio
  Kashino.
\newblock {BYOL} for audio: Exploring pre-trained general-purpose audio
  representations.
\newblock \emph{{IEEE} {ACM} Trans. Audio Speech Lang. Process.}, 31:\penalty0
  137--151, 2023.
\newblock \doi{10.1109/TASLP.2022.3221007}.
\newblock URL \url{https://doi.org/10.1109/TASLP.2022.3221007}.

\bibitem[Oquab et~al.(2023)Oquab, Darcet, Moutakanni, Vo, Szafraniec, Khalidov,
  Fernandez, Haziza, Massa, El{-}Nouby, Assran, Ballas, Galuba, Howes, Huang,
  Li, Misra, Rabbat, Sharma, Synnaeve, Xu, J{\'{e}}gou, Mairal, Labatut,
  Joulin, and Bojanowski]{DBLP:journals/corr/abs-2304-07193}
Maxime Oquab, Timoth{\'{e}}e Darcet, Th{\'{e}}o Moutakanni, Huy Vo, Marc
  Szafraniec, Vasil Khalidov, Pierre Fernandez, Daniel Haziza, Francisco Massa,
  Alaaeldin El{-}Nouby, Mahmoud Assran, Nicolas Ballas, Wojciech Galuba,
  Russell Howes, Po{-}Yao Huang, Shang{-}Wen Li, Ishan Misra, Michael~G.
  Rabbat, Vasu Sharma, Gabriel Synnaeve, Hu~Xu, Herv{\'{e}} J{\'{e}}gou, Julien
  Mairal, Patrick Labatut, Armand Joulin, and Piotr Bojanowski.
\newblock Dinov2: Learning robust visual features without supervision.
\newblock \emph{CoRR}, abs/2304.07193, 2023.
\newblock \doi{10.48550/arXiv.2304.07193}.
\newblock URL \url{https://doi.org/10.48550/arXiv.2304.07193}.

\bibitem[Ott et~al.(2019)Ott, Edunov, Baevski, Fan, Gross, Ng, Grangier, and
  Auli]{ott2019fairseq}
Myle Ott, Sergey Edunov, Alexei Baevski, Angela Fan, Sam Gross, Nathan Ng,
  David Grangier, and Michael Auli.
\newblock fairseq: A fast, extensible toolkit for sequence modeling.
\newblock In \emph{Proceedings of NAACL-HLT 2019: Demonstrations}, 2019.

\bibitem[Panayotov et~al.(2015)Panayotov, Chen, Povey, and
  Khudanpur]{panayotov2015librispeech}
Vassil Panayotov, Guoguo Chen, Daniel Povey, and Sanjeev Khudanpur.
\newblock Librispeech: an asr corpus based on public domain audio books.
\newblock In \emph{2015 IEEE International Conference on Acoustics, Speech and
  Signal Processing (ICASSP)}, pp.\  5206--5210. IEEE, 2015.

\bibitem[Park et~al.(2019)Park, Chan, Zhang, Chiu, Zoph, Cubuk, and
  Le]{park2019specaug}
Daniel~S Park, William Chan, Yu~Zhang, Chung-Cheng Chiu, Barret Zoph, Ekin~D
  Cubuk, and Quoc~V Le.
\newblock Specaugment: A simple data augmentation method for automatic speech
  recognition.
\newblock \emph{Proc. Interspeech 2019}, pp.\  2613--2617, 2019.

\bibitem[Polyak \& Juditsky(1992)Polyak and Juditsky]{Polyak92}
B.~T. Polyak and A.~B. Juditsky.
\newblock Acceleration of stochastic approximation by averaging.
\newblock \emph{SIAM Journal on Control and Optimization}, 30\penalty0
  (4):\penalty0 838--855, 1992.
\newblock \doi{10.1137/0330046}.
\newblock URL \url{https://doi.org/10.1137/0330046}.

\bibitem[Qiao et~al.(2019)Qiao, Wang, Liu, Shen, and
  Yuille]{DBLP:journals/corr/abs-1903-10520}
Siyuan Qiao, Huiyu Wang, Chenxi Liu, Wei Shen, and Alan~L. Yuille.
\newblock Weight standardization.
\newblock \emph{CoRR}, abs/1903.10520, 2019.
\newblock URL \url{http://arxiv.org/abs/1903.10520}.

\bibitem[Richemond et~al.(2020)Richemond, Grill, Altch{\'{e}}, Tallec, Strub,
  Brock, Smith, De, Pascanu, Piot, and
  Valko]{DBLP:journals/corr/abs-2010-10241}
Pierre~H. Richemond, Jean{-}Bastien Grill, Florent Altch{\'{e}}, Corentin
  Tallec, Florian Strub, Andrew Brock, Samuel~L. Smith, Soham De, Razvan
  Pascanu, Bilal Piot, and Michal Valko.
\newblock {BYOL} works even without batch statistics.
\newblock \emph{CoRR}, abs/2010.10241, 2020.
\newblock URL \url{https://arxiv.org/abs/2010.10241}.

\bibitem[Richemond et~al.(2023)Richemond, Tam, Tang, Strub, Piot, and
  Hill]{DBLP:journals/corr/abs-2302-04817}
Pierre~H. Richemond, Allison~C. Tam, Yunhao Tang, Florian Strub, Bilal Piot,
  and Felix Hill.
\newblock The edge of orthogonality: {A} simple view of what makes {BYOL} tick.
\newblock \emph{CoRR}, abs/2302.04817, 2023.
\newblock \doi{10.48550/arXiv.2302.04817}.
\newblock URL \url{https://doi.org/10.48550/arXiv.2302.04817}.

\bibitem[Ruppert(1988)]{Ruppert1988EfficientEF}
David Ruppert.
\newblock Efficient estimations from a slowly convergent robbins-monro process.
\newblock 1988.

\bibitem[Russakovsky et~al.(2014)Russakovsky, Deng, Su, Krause, Satheesh, Ma,
  Huang, Karpathy, Khosla, Bernstein, and
  Fei{-}Fei]{DBLP:journals/corr/RussakovskyDSKSMHKKBBF14}
Olga Russakovsky, Jia Deng, Hao Su, Jonathan Krause, Sanjeev Satheesh, Sean Ma,
  Zhiheng Huang, Andrej Karpathy, Aditya Khosla, Michael~S. Bernstein, and
  Li~Fei{-}Fei.
\newblock Imagenet large scale visual recognition challenge.
\newblock \emph{CoRR}, abs/1409.0575, 2014.
\newblock URL \url{http://arxiv.org/abs/1409.0575}.

\bibitem[Shaw et~al.(2018)Shaw, Uszkoreit, and Vaswani]{shaw2018self}
Peter Shaw, Jakob Uszkoreit, and Ashish Vaswani.
\newblock Self-attention with relative position representations.
\newblock In \emph{Proceedings of the 2018 Conference of the North American
  Chapter of the Association for Computational Linguistics: Human Language
  Technologies, Volume 2 (Short Papers)}, pp.\  464--468, 2018.

\bibitem[Smith \& Le(2018)Smith and Le]{DBLP:conf/iclr/SmithL18}
Samuel~L. Smith and Quoc~V. Le.
\newblock A bayesian perspective on generalization and stochastic gradient
  descent.
\newblock In \emph{6th International Conference on Learning Representations,
  {ICLR} 2018, Vancouver, BC, Canada, April 30 - May 3, 2018, Conference Track
  Proceedings}. OpenReview.net, 2018.
\newblock URL \url{https://openreview.net/forum?id=BJij4yg0Z}.

\bibitem[Smith et~al.(2018)Smith, Kindermans, Ying, and
  Le]{DBLP:conf/iclr/SmithKYL18}
Samuel~L. Smith, Pieter{-}Jan Kindermans, Chris Ying, and Quoc~V. Le.
\newblock Don't decay the learning rate, increase the batch size.
\newblock In \emph{6th International Conference on Learning Representations,
  {ICLR} 2018, Vancouver, BC, Canada, April 30 - May 3, 2018, Conference Track
  Proceedings}. OpenReview.net, 2018.
\newblock URL \url{https://openreview.net/forum?id=B1Yy1BxCZ}.

\bibitem[Sohn et~al.(2020)Sohn, Berthelot, Carlini, Zhang, Zhang, Raffel,
  Cubuk, Kurakin, and Li]{sohn2020fixmatch}
Kihyuk Sohn, David Berthelot, Nicholas Carlini, Zizhao Zhang, Han Zhang,
  Colin~A Raffel, Ekin~Dogus Cubuk, Alexey Kurakin, and Chun-Liang Li.
\newblock Fixmatch: Simplifying semi-supervised learning with consistency and
  confidence.
\newblock \emph{Advances in neural information processing systems},
  33:\penalty0 596--608, 2020.

\bibitem[Srivastava et~al.(2022)Srivastava, Rastogi, Rao, Shoeb, Abid, Fisch,
  Brown, Santoro, Gupta, Garriga{-}Alonso, Kluska, Lewkowycz, Agarwal, Power,
  Ray, Warstadt, Kocurek, Safaya, Tazarv, Xiang, Parrish, Nie, Hussain, Askell,
  Dsouza, Rahane, Iyer, Andreassen, Santilli, Stuhlm{\"{u}}ller, Dai, La,
  Lampinen, Zou, Jiang, Chen, Vuong, Gupta, Gottardi, Norelli, Venkatesh,
  Gholamidavoodi, Tabassum, Menezes, Kirubarajan, Mullokandov, Sabharwal,
  Herrick, Efrat, Erdem, Karakas, and
  et~al.]{DBLP:journals/corr/abs-2206-04615}
Aarohi Srivastava, Abhinav Rastogi, Abhishek Rao, Abu Awal~Md Shoeb, Abubakar
  Abid, Adam Fisch, Adam~R. Brown, Adam Santoro, Aditya Gupta, Adri{\`{a}}
  Garriga{-}Alonso, Agnieszka Kluska, Aitor Lewkowycz, Akshat Agarwal, Alethea
  Power, Alex Ray, Alex Warstadt, Alexander~W. Kocurek, Ali Safaya, Ali Tazarv,
  Alice Xiang, Alicia Parrish, Allen Nie, Aman Hussain, Amanda Askell, Amanda
  Dsouza, Ameet Rahane, Anantharaman~S. Iyer, Anders Andreassen, Andrea
  Santilli, Andreas Stuhlm{\"{u}}ller, Andrew~M. Dai, Andrew La, Andrew~K.
  Lampinen, Andy Zou, Angela Jiang, Angelica Chen, Anh Vuong, Animesh Gupta,
  Anna Gottardi, Antonio Norelli, Anu Venkatesh, Arash Gholamidavoodi, Arfa
  Tabassum, Arul Menezes, Arun Kirubarajan, Asher Mullokandov, Ashish
  Sabharwal, Austin Herrick, Avia Efrat, Aykut Erdem, Ayla Karakas, and et~al.
\newblock Beyond the imitation game: Quantifying and extrapolating the
  capabilities of language models.
\newblock \emph{CoRR}, abs/2206.04615, 2022.
\newblock \doi{10.48550/arXiv.2206.04615}.
\newblock URL \url{https://doi.org/10.48550/arXiv.2206.04615}.

\bibitem[Tarvainen \& Valpola(2017)Tarvainen and
  Valpola]{DBLP:conf/nips/TarvainenV17}
Antti Tarvainen and Harri Valpola.
\newblock Mean teachers are better role models: Weight-averaged consistency
  targets improve semi-supervised deep learning results.
\newblock In Isabelle Guyon, Ulrike von Luxburg, Samy Bengio, Hanna~M. Wallach,
  Rob Fergus, S.~V.~N. Vishwanathan, and Roman Garnett (eds.), \emph{Advances
  in Neural Information Processing Systems 30: Annual Conference on Neural
  Information Processing Systems 2017, December 4-9, 2017, Long Beach, CA,
  {USA}}, pp.\  1195--1204, 2017.
\newblock URL
  \url{https://proceedings.neurips.cc/paper/2017/hash/68053af2923e00204c3ca7c6a3150cf7-Abstract.html}.

\bibitem[Tian et~al.(2021)Tian, Chen, and Ganguli]{DBLP:conf/icml/TianCG21}
Yuandong Tian, Xinlei Chen, and Surya Ganguli.
\newblock Understanding self-supervised learning dynamics without contrastive
  pairs.
\newblock In Marina Meila and Tong Zhang (eds.), \emph{Proceedings of the 38th
  International Conference on Machine Learning, {ICML} 2021, 18-24 July 2021,
  Virtual Event}, volume 139 of \emph{Proceedings of Machine Learning
  Research}, pp.\  10268--10278. {PMLR}, 2021.
\newblock URL \url{http://proceedings.mlr.press/v139/tian21a.html}.

\bibitem[Tieleman et~al.(2012)Tieleman, Hinton, et~al.]{rmsprop}
Tijmen Tieleman, Geoffrey Hinton, et~al.
\newblock Lecture 6.5-rmsprop: Divide the gradient by a running average of its
  recent magnitude.
\newblock \emph{COURSERA: Neural networks for machine learning}, 4\penalty0
  (2):\penalty0 26--31, 2012.

\bibitem[Vaswani et~al.(2017)Vaswani, Shazeer, Parmar, Uszkoreit, Jones, Gomez,
  Kaiser, and Polosukhin]{DBLP:conf/nips/VaswaniSPUJGKP17}
Ashish Vaswani, Noam Shazeer, Niki Parmar, Jakob Uszkoreit, Llion Jones,
  Aidan~N. Gomez, Lukasz Kaiser, and Illia Polosukhin.
\newblock Attention is all you need.
\newblock In Isabelle Guyon, Ulrike von Luxburg, Samy Bengio, Hanna~M. Wallach,
  Rob Fergus, S.~V.~N. Vishwanathan, and Roman Garnett (eds.), \emph{Advances
  in Neural Information Processing Systems 30: Annual Conference on Neural
  Information Processing Systems 2017, December 4-9, 2017, Long Beach, CA,
  {USA}}, pp.\  5998--6008, 2017.
\newblock URL
  \url{https://proceedings.neurips.cc/paper/2017/hash/3f5ee243547dee91fbd053c1c4a845aa-Abstract.html}.

\bibitem[Wu \& He(2018)Wu and He]{DBLP:conf/eccv/WuH18}
Yuxin Wu and Kaiming He.
\newblock Group normalization.
\newblock In Vittorio Ferrari, Martial Hebert, Cristian Sminchisescu, and Yair
  Weiss (eds.), \emph{Computer Vision - {ECCV} 2018 - 15th European Conference,
  Munich, Germany, September 8-14, 2018, Proceedings, Part {XIII}}, volume
  11217 of \emph{Lecture Notes in Computer Science}, pp.\  3--19. Springer,
  2018.
\newblock \doi{10.1007/978-3-030-01261-8\_1}.
\newblock URL \url{https://doi.org/10.1007/978-3-030-01261-8\_1}.

\bibitem[Xu et~al.(2020)Xu, Likhomanenko, Kahn, Hannun, Synnaeve, and
  Collobert]{xu2020iterative}
Qiantong Xu, Tatiana Likhomanenko, Jacob Kahn, Awni Hannun, Gabriel Synnaeve,
  and Ronan Collobert.
\newblock Iterative pseudo-labeling for speech recognition.
\newblock \emph{Proc. Interspeech 2020}, pp.\  1006--1010, 2020.

\bibitem[You et~al.(2017)You, Gitman, and
  Ginsburg]{DBLP:journals/corr/abs-1708-03888}
Yang You, Igor Gitman, and Boris Ginsburg.
\newblock Scaling {SGD} batch size to 32k for imagenet training.
\newblock \emph{CoRR}, abs/1708.03888, 2017.
\newblock URL \url{http://arxiv.org/abs/1708.03888}.

\bibitem[You et~al.(2020)You, Li, Reddi, Hseu, Kumar, Bhojanapalli, Song,
  Demmel, Keutzer, and Hsieh]{DBLP:conf/iclr/YouLRHKBSDKH20}
Yang You, Jing Li, Sashank~J. Reddi, Jonathan Hseu, Sanjiv Kumar, Srinadh
  Bhojanapalli, Xiaodan Song, James Demmel, Kurt Keutzer, and Cho{-}Jui Hsieh.
\newblock Large batch optimization for deep learning: Training {BERT} in 76
  minutes.
\newblock In \emph{8th International Conference on Learning Representations,
  {ICLR} 2020, Addis Ababa, Ethiopia, April 26-30, 2020}. OpenReview.net, 2020.
\newblock URL \url{https://openreview.net/forum?id=Syx4wnEtvH}.

\end{thebibliography}
\bibliographystyle{templates/iclr2021/iclr2021_conference}

\clearpage
\appendix
\hypersetup{linkcolor=black}
\begin{spacing}{0.87}
\appendixpage
\startcontents[sections]
\printcontents[sections]{l}{1}

\clearpage
\end{spacing}
\hypersetup{linkcolor=hrefblue}

\section{Broader impact}
\label{app:broader-impact}

\glsresetall

This work shows how to adapt \gls{ml} optimization in the presence of a model \gls{ema}.
There are a number of benefits to this:
\begin{enumerate}[leftmargin=0.75cm]
    \item Scaling rules democratize the training of \gls{ml} models: they give \gls{ml} researchers the ability to replicate the optimization of large scale systems, even if those researchers \emph{do not} have access to i) significant parallel computational resources \emph{or} ii) the technical tooling to do so.
    \item Our \gls{ema} Scaling Rule lowers compute usage as it removes the necessity for a hyperparameter search over momenta; in the case where our scaling assumptions hold, if we know the value of the optimal momentum $\rho_B$ at some batch size $B$ (for example, the momentum that gives the best transfer performance), then the optimal value at another batch size $\hat B$ is exactly the one given by the \gls{ema} Scaling Rule $\hat\rho=\rho_B^\kappa$, for scaling $\kappa=\hat B/B$.
    \item Our \gls{ema} Scaling Rule enables researchers to more quickly iterate through experimental ideas, and opens up access to large-scale training (for example, larger models and larger datasets) for Pseudo-Labeling and \gls{ssl} techniques.
\end{enumerate}
These points have potential negative consequences:
\begin{enumerate}[leftmargin=0.75cm]
    \item As our \gls{ema} Scaling Rule enables researchers to iterate the same experiments more quickly, and perform large-scale training with \gls{ema}-based methods, this may encourage a greater number of experiments, or the training of larger models. 
    Either of these possibilities leads to greater energy consumption.
    \item As the need to determine momentum hyperparameters has now been removed, researchers who were previously discouraged from attempting to scale these methods due to an \emph{extra} hyperparameter to tune may begin to perform such experiments, leading, once more, to greater energy consumption.
\end{enumerate}
The environmental impact of each of these two points may be significant.

\section{Limitations}
\label{app:limitations}

The \gls{ema} Scaling Rule provides a recipe for producing training dynamics independent of the batch size used in stochastic optimization.
The technology underpinning it will not \emph{always} give the desired behavior, however.

The first issue occurs with the wording present in the \gls{ema} Scaling Rule: \emph{[...] and scale other optimizers according to their own scaling rules} (\Cref{def:ema-sr}):
\begin{enumerate}[leftmargin=0.75cm]
    \item This statement requires that the given \gls{sde} approximation we are using for the model optimizer is itself providing well-behaved scaling, that is, that in the \emph{absence} of a model \gls{ema}, the model optimization trajectories at the batch sizes $B$ and $\kappa B$, with optimizer hyperparameters appropriately scaled, are close.
    In general we know this is not true.
    First, we know that the \gls{sde} approximation for \gls{sgd} breaks at a given $\kappa$ due to discretization error \citep{DBLP:conf/nips/LiMA21}.
    Second, we know that if the gradient noise is not sufficiently large, the \gls{sde} approximation for Adam does not exist \citep{DBLP:conf/nips/MalladiLPA22}, i.e. an \gls{sde} motivated scaling rule has no meaning.
    \item This statement requires knowledge of how to scale the corresponding model optimizer.
    We have principled ways to achieve this for \gls{sgd} \citep{DBLP:conf/nips/LiMA21}, and for the adaptive optimization methods RMSProp and Adam \citep{DBLP:conf/nips/MalladiLPA22}.
    Empirically, a square-root scaling law for LAMB \citep{DBLP:conf/iclr/YouLRHKBSDKH20} has been observed, however, it has not been derived formally.
    Problematically, there is no known hyperparameter scaling law or \gls{sde} approximation known for LARS \citep{DBLP:journals/corr/abs-1708-03888}, which has been used in \gls{byol} \citep{DBLP:conf/nips/GrillSATRBDPGAP20} and many other large-scale training procedures for convolution-based architectures. 
    Despite this, we are able to demonstrate in \Cref{subsec:byol-additional} that a combination of the \gls{ema} Scaling Rule and progressive scaling can match, or surpass baseline \gls{byol} performance at a batch size of 32,768 using LARS, indicating that although the theoretical guarantees may not hold, there is still practical utility in the tools we provide in this work.
    \item It may be the case that the optimal performance attainable by a given model setup exists at a level of discretization/gradient noise where no \gls{sde} exists. 
    In this case, \gls{sde}-derived scaling rules can never be valid, and no scaling of this dynamics can be achieved with known tools.
\end{enumerate}
The second issue is related to the case when the optimizer scaling rule is valid.
In this case, the error for the \gls{ema} Scaling Rule at finite learning rate $\eta$ at large $\kappa$ can be considerable.
In cases where the model \gls{ema} plays a role in the overall optimization, the error introduced by the \gls{ema} Scaling Rule can break the preservation of model dynamics.

Put another way, an optimizer scaling rule and the \gls{ema} Scaling Rule each introduce their own discretization errors. 
In the case where \gls{ema} plays a role in optimization, as soon as the discretization error of \emph{either} the optimizer scaling rule \emph{or} the \gls{ema} Scaling Rule is large, the error for the joint optimization procedure is large.
This is \emph{at least} as bad as cases that \emph{do not} use a model \gls{ema} during the optimization process.

\section{The scaling toolbox: practical methods for enabling systematic scaling}
\label{app:scaling-toolbox}

There are many different components involved in preserving optimization dynamics at different batch sizes. 
In this appendix we collect into a single place the different concepts and values that we found useful in practice, in an attempt to make the practice of scaling as accessible as possible.

\subsection{The continuous time/SDE perspective}
\label{sec:app-sde-perspective}

Here we discuss the mindset difference required when trying to preserve training dynamics.
In \gls{ml} we typically use stochastic optimization, leading us to think of the optimization in terms of \emph{performing updates}, or \emph{stepping the optimizer}.
This notion has become more common in the era of large datasets, where it may be the case that we only see a fraction of the dataset during optimization.

For dynamics preservation under scaling, we suggest that it is simpler to consider the \emph{amount of data} seen by the training process, or alternatively, the amount of \emph{continuous time} in the discretization of \glspl{sde} view.
The reason is the following.
The \gls{sde} scaling rule results 
(\Cref{def:ema-sr}, \cite{li2019stochastic,DBLP:conf/nips/LiMA21,DBLP:conf/nips/MalladiLPA22}) follow from showing that different discretizations of the \gls{sde} are close to that \gls{sde}, providing we appropriately scale hyperparameters (see \Cref{subsec:ema-sdes}).
Each of these discretizations shares the \emph{total continuous time} $T=\hat\eta\times \widehat {N}_{\text{iter}}$\footnote{This is in the case of \gls{sgd}, for RMSProp and Adam one should use $T=\hat{\eta}^2\times \widehat {N}_{\text{iter}}$ \citep{DBLP:conf/nips/MalladiLPA22}.} of the underlying \gls{sde}, but each discretization has a \emph{different} number of iterations $\widehat {N}_{\text{iter}}={N}_{\text{iter}}/\kappa $.

This perspective is already adopted, perhaps by accident in some domains.
For example, in \gls{cv}, 
it is typical to compare model performance after optimization on ImageNet1k after a \emph{number of epochs},
whilst also specifing a learning rate warmup after a \emph{number of epochs}.
This transforms the schedule into the form \emph{wait until the process meets [condition]}, where here \emph{[condition]} is \emph{when the process has seen sufficiently many samples.}

More generally, we can specify any \emph{condition} that is not a property of the discretization procedure itself.
Instead, the discretization procedure should be viewed as a numerical approximation method for the \gls{sde} we are evolving, and the properties of that discretization process (like \emph{number of steps}) are not \emph{of specific interest} in the world view where we do decouple optimization from the batch size.
A specific example of this more general case is present in \Cref{subsec:semi-supervised}, where for scaling $\kappa>2$ we wait until the pre-training \gls{wer} is sufficiently low.

There may be cases where one is working with a setup that is explicitly defined in terms of quantities related to the discretization process.
Indeed, the optimizer hyperparameters are examples of these, and need to be scaled accordingly with $\kappa$.
The other typical example of this is conditions based on the \emph{number of optimizer steps}, rather than the number of epochs.
In this case, these quantities should be scaled to achieve the desired condition in the same amount of time, i.e. as above $\widehat {N}_{\text{iter}}={N}_{\text{iter}}/\kappa$, where ${N}_{\text{iter}}$ is the number of iterations specified at the base batch size $B$. 
Concretely, if training is specified in a number of steps, then doubling the batch size implies you should train for half the number of steps.

\subsection{Scaling rules for optimization}

For ease of reference, we collect all the scaling rules related to batch size modification we are aware of.
We begin with the most well-known, the \gls{sgd} Scaling Rule (\Cref{def:lsr,def:app-sgd-scaling-rule}).
\begin{definition}[\gls{sgd} Scaling Rule]
    When running \gls{sgd} (\Cref{def:sgd}) with batch size $\hat B=\kappa B$,
    use a learning rate $\hat\eta=\kappa\eta$ \citep{DBLP:journals/corr/Krizhevsky14,DBLP:journals/corr/GoyalDGNWKTJH17}. 
    \label{def:app-sgd-scaling-rule}
\end{definition}
The \gls{sgd} Scaling Rule is also known as the Linear Scaling Rule (LSR), although for clarity, this work adopts the naming convention \emph{[Algorithm Name] Scaling Rule},
which means all parameters of those algorithms are appropriately scaled from batch size $B$ to $\kappa B$.

Next, we give the two scaling rules known for the adapative optimizers RMSProp \citep{rmsprop} and Adam \citep{DBLP:journals/corr/KingmaB14} in \Cref{def:rmpsprop-sr} and \Cref{def:adam-sr} respectively.

\begin{definition}[RMSProp Scaling Rule]
    When running RMSProp  \citep{rmsprop} with batch size $\hat B=\kappa B$,
    use a learning rate $\hat\eta=\sqrt\kappa\eta$,
    beta coefficient $\hat\beta=1-\kappa\times(1-\beta)$, and adaptivity parameter
    $\hat\epsilon=\frac\epsilon{\sqrt\kappa}$
    \citep{DBLP:conf/nips/MalladiLPA22}.
    \label{def:rmpsprop-sr}
\end{definition}

\begin{definition}[Adam Scaling Rule]
    When running Adam \citep{DBLP:journals/corr/KingmaB14} with batch size $\hat B=\kappa B$,
    use a learning rate $\hat\eta=\sqrt\kappa\eta$,
    beta coefficients 
    $\hat\beta_1=1-\kappa\times(1-\beta_1)$,
    $\hat\beta_2=1-\kappa\times(1-\beta_2)$, and adaptivity parameter
    $\hat\epsilon=\frac\epsilon{\sqrt\kappa}$
    \citep{DBLP:conf/nips/MalladiLPA22}.
    \label{def:adam-sr}
\end{definition}

Next, we present a contribution of this work, the \gls{ema} Scaling Rule
(\Cref{def:ema-sr,def:app-ema-scaling-rule}), which extends the above scaling rules to allow the presence of a model \gls{ema} which is able to contribute to the overall optimization (see \Cref{app:ema-approximation-theorem,app:matrix-calculations} for derivations).

\begin{definition}[\gls{ema} Scaling Rule]
    When computing the \gls{ema} update (\Cref{def:ema}) of a model undergoing stochastic optimization with batch size $\hat B=\kappa B$,
    use a momentum $\hat\rho=\rho^\kappa$ and scale other optimizers according to their own scaling rules.
\label{def:app-ema-scaling-rule}
\end{definition}

Concretely, if we are using \gls{sgd} in the presence of a model \gls{ema},
\Cref{def:app-sgd-scaling-rule,def:app-ema-scaling-rule} state that we should take 
$\hat\eta=\kappa\eta$ \emph{and} $\hat\rho=\rho^\kappa$ when scaling by $\kappa=\hat B/B$.

The final scaling rule is for weight decay, and follows from the scaling logic discussed in \Cref{sec:app-sde-perspective} and \citet{DBLP:journals/corr/Krizhevsky14}.
If we take the weight decay regularization penalty $\lambda$ defined at batch size $B$, what should the weight decay $\hat\lambda$ be for batch size $\hat B=\kappa B$?
For simplicity, consider $\kappa$ updates of optimization of parameters $\rvtheta_t$ in the presence of weight decay only
\begin{equation}
    \rvtheta_{t+\kappa}
    =\rvtheta_{t+\kappa-1}-\eta\,\lambda\,\rvtheta_{t+\kappa-1}
    =(1 - \eta\,\lambda)\,\rvtheta_{t+\kappa-1}
    =(1 - \eta\,\lambda)^\kappa\,\rvtheta_{t}.
\end{equation}
Therefore, to match the effect of weight decay with a single iteration step, we need to match
\begin{equation}
    1-\hat\eta \,\hat \lambda = (1 - \eta\,\lambda)^\kappa.
\end{equation}
Solving for $\hat\lambda$ and expanding around $\eta\approx 0$ gives
\begin{equation}
    \hat\lambda
    =\frac{1-(1 - \eta\,\lambda)^\kappa}{\hat\eta}
    \approx 
    \frac{\eta}{\hat \eta}\times 
    \kappa \,\lambda
    +\mathcal{O}(\eta).
\end{equation}
This leads to the Weight Decay Scaling Rule (\Cref{def:wd-sr}).

\begin{definition}[Weight Decay Scaling Rule]
    When using weight decay with batch size $\hat B=\kappa B$,
    use a penalty term $\hat\lambda=(\kappa \hat\eta / \eta)\,\lambda$,
    where $\hat\eta$ and $\eta$ represent the scaled and unscaled learning rates of the corresponding optimizer
    \citep{DBLP:journals/corr/Krizhevsky14,DBLP:conf/nips/Li0TSG18,DBLP:conf/iclr/LoshchilovH19}.
    \label{def:wd-sr}
\end{definition}

The Weight Decay Scaling Rule implies that using \emph{linear} scaling for the learning rate $\eta$ then the weight decay penalty is automatically scaled,
and when using \emph{square-root} scaling for the learning rate $\eta$ (e.g. in the case of the Adam Scaling Rule (\Cref{def:adam-sr})) then the weight decay penalty should also be scaled with a \emph{square-root} as is proposed in \citet{DBLP:conf/iclr/LoshchilovH19}.

Finally, we see that if the implementation of weight decay does not have an update scaled by the learning rate, i.e. the update is $\rvtheta_{t+1}=(1-\lambda)\,\rvtheta_t$, then the scaling rule is optimizer-independent, and becomes linear for small weight decay, i.e. $\hat\lambda=\kappa\lambda$,
and for arbitrary $\lambda$ takes the form $\hat\lambda=1-(1-\lambda)^\kappa$.

\subsection{Commonly used values of hyperparameters at different batch sizes}

In the literature it is common to give a base learning rate $\eta$ defined at batch size 256, implicitly using the \gls{sgd} Scaling Rule, even when using the Adam optimizer.
Because the scaling of other optimization hyperparameters was not understood until recently, it is also common to just present these \emph{for the experiment}, e.g. the Adam betas and epsilon, and the \gls{ema} momentum, implicitly defined at the scale of the experiment, for example at batch size 4096.
One way to deal with this in practice is to define a single reference batch size $B$ at which \emph{all} hyperparameters are defined, and then scale from there.
In this case, it is easiest to compute \emph{using linear scaling} the learning rate at the redefined base batch size 
$\eta = \tilde\kappa\,\eta_{\text{orig}}$,
where $\tilde\kappa=B/B_{\text{orig}}$,
and then scale this new reference $\eta$ as $\hat\eta=\kappa\eta$, $\kappa=\hat B/B$, along with e.g. the momentum defined at $B$.

As this process can be slightly frustrating, we have provided tables of typical learning rates in \Cref{tab:common-hparams-lr} and momenta in \Cref{tab:common-hparams-momenta}.

\newcommand{\firstmidrules}{
\cmidrule(r){2-4}
\cmidrule(r){5-7}
}
\newcommand{\secondmidrules}{
\cmidrule(r){2-3}
\cmidrule(r){4-4}
\cmidrule(r){5-5}
\cmidrule(r){6-7}
}

\begin{table}[t]
\centering
\caption{Scaled learning rates $\hat\eta$ at different batch sizes $\hat B=\kappa B$ given reference learning rates $\eta$ defined at batch size $B$. The reference values of each column are boldened. 
Note that this is only valid when there is a notion of \emph{single sample}. In the sequence learning setup (for example, in \Cref{subsec:semi-supervised}), the notion of batch size should be appropriately replaced with the \emph{dynamic batch size}, i.e. total sequence length.}
\small
\label{tab:common-hparams-lr}
\begin{tabular}{cllllll}
\toprule
{} & \multicolumn{3}{c}{$\hat\eta=\kappa\eta$ [SGD]} & \multicolumn{3}{c}{$\hat\eta=\sqrt\kappa\eta$ [RMSProp, Adam]} \\ \firstmidrules
{} & \multicolumn{2}{c}{$B=256$} &         $B=512$ &                                    $B=256$ & \multicolumn{2}{c}{$B=4096$} \\ \secondmidrules
Batch size $\hat B$&                  $\eta=0.1$ &      $\eta=0.3$ &      $\eta=0.1$ &                             $\eta=10^{-3}$ &      $\eta=4.8$ &    $\eta=10^{-3}$ \\
\midrule
$32$                &                    $0.0125$ &        $0.0375$ &       $0.00625$ &                                  $0.00035$ &       $0.42426$ &         $0.00009$ \\
$64$                &                     $0.025$ &         $0.075$ &        $0.0125$ &                                   $0.0005$ &           $0.6$ &         $0.00013$ \\
$128$               &                      $0.05$ &          $0.15$ &         $0.025$ &                                  $0.00071$ &       $0.84853$ &         $0.00018$ \\
$256$               &              $\mathbf{0.1}$ &  $\mathbf{0.3}$ &          $0.05$ &                           $\mathbf{0.001}$ &           $1.2$ &         $0.00025$ \\
$512$               &                       $0.2$ &           $0.6$ &  $\mathbf{0.1}$ &                                  $0.00141$ &       $1.69706$ &         $0.00035$ \\
$1024$              &                       $0.4$ &           $1.2$ &           $0.2$ &                                    $0.002$ &           $2.4$ &          $0.0005$ \\
$2048$              &                       $0.8$ &           $2.4$ &           $0.4$ &                                  $0.00283$ &       $3.39411$ &         $0.00071$ \\
$4096$              &                       $1.6$ &           $4.8$ &           $0.8$ &                                    $0.004$ &  $\mathbf{4.8}$ &  $\mathbf{0.001}$ \\
$8192$              &                       $3.2$ &           $9.6$ &           $1.6$ &                                  $0.00566$ &       $6.78823$ &         $0.00141$ \\
$16384$             &                       $6.4$ &          $19.2$ &           $3.2$ &                                    $0.008$ &           $9.6$ &           $0.002$ \\
$32768$             &                      $12.8$ &          $38.4$ &           $6.4$ &                                  $0.01131$ &      $13.57645$ &         $0.00283$ \\
$65536$             &                      $25.6$ &          $76.8$ &          $12.8$ &                                    $0.016$ &          $19.2$ &           $0.004$ \\
\bottomrule
\end{tabular}
\end{table}

\newcommand{\thirdmidrules}{
\cmidrule(r){2-4}
\cmidrule(r){5-8}
}

\begin{table}[t]
\centering
\caption{Scaled EMA momenta $\hat\rho=\rho^\kappa$ at different batch sizes $\hat B=\kappa B$ given reference momenta $\rho$ defined at batch size $B$. The reference values of each column are boldened. Again in the sequence learning setup, batch size should be appropriately replaced with a notion of sequence length.}
\small
\label{tab:common-hparams-momenta}
\begin{tabular}{clllllll}
\toprule
{} & \multicolumn{3}{c}{$B=256$} & \multicolumn{4}{c}{$B=4096$} \\ \thirdmidrules
Batch size $\hat B$&      $\rho=0.9999$ &      $\rho=0.999$ &      $\rho=0.99$ &      $\rho=0.996$ &      $\rho=0.992$ &      $\rho=0.99$ &      $\rho=0.97$ \\
\midrule
$32$                &          $0.99999$ &         $0.99987$ &        $0.99874$ &         $0.99997$ &         $0.99994$ &        $0.99992$ &        $0.99976$ \\
$64$                &          $0.99997$ &         $0.99975$ &        $0.99749$ &         $0.99994$ &         $0.99987$ &        $0.99984$ &        $0.99952$ \\
$128$               &          $0.99995$ &          $0.9995$ &        $0.99499$ &         $0.99987$ &         $0.99975$ &        $0.99969$ &        $0.99905$ \\
$256$               &  $\mathbf{0.9999}$ &  $\mathbf{0.999}$ &  $\mathbf{0.99}$ &         $0.99975$ &          $0.9995$ &        $0.99937$ &         $0.9981$ \\
$512$               &           $0.9998$ &           $0.998$ &         $0.9801$ &          $0.9995$ &           $0.999$ &        $0.99874$ &         $0.9962$ \\
$1024$              &           $0.9996$ &         $0.99601$ &         $0.9606$ &           $0.999$ &         $0.99799$ &        $0.99749$ &        $0.99241$ \\
$2048$              &           $0.9992$ &         $0.99203$ &        $0.92274$ &           $0.998$ &         $0.99599$ &        $0.99499$ &        $0.98489$ \\
$4096$              &           $0.9984$ &         $0.98412$ &        $0.85146$ &  $\mathbf{0.996}$ &  $\mathbf{0.992}$ &  $\mathbf{0.99}$ &  $\mathbf{0.97}$ \\
$8192$              &           $0.9968$ &         $0.96849$ &        $0.72498$ &         $0.99202$ &         $0.98406$ &         $0.9801$ &         $0.9409$ \\
$16384$             &          $0.99362$ &         $0.93798$ &         $0.5256$ &          $0.9841$ &         $0.96838$ &         $0.9606$ &        $0.88529$ \\
$32768$             &          $0.98728$ &          $0.8798$ &        $0.27625$ &         $0.96844$ &         $0.93776$ &        $0.92274$ &        $0.78374$ \\
$65536$             &          $0.97472$ &         $0.77405$ &        $0.07632$ &         $0.93788$ &          $0.8794$ &        $0.85146$ &        $0.61425$ \\
\bottomrule
\end{tabular}
\end{table}

\subsection{Progressive scaling}
\label{subsec:dynamic-batch-scaling}

In \Cref{subsec:self-supervised} we introduced Progressive Scaling (\Cref{def:progressive-scaling}) to test our hypothesis that early in the \gls{byol} training procedure, there are dynamics that are challenging to replicate at larger batch sizes.
To remove ambiguity, in \Cref{alg:progressive-scaling} we provide pseudo-code for how to use Progressive Scaling.

\begin{algorithm}[t!]
\caption{Stochastic Gradient Descent with Progressive Scaling}\label{alg:progressive-scaling}
\begin{algorithmic}
\Require Base learning rate $\eta$, base momentum $\rho$ for base batch size $B$
\Require Initial target model parameters $\rvtheta$ and model \gls{ema} parameters $\rvzeta$
\Require Epochs $E$ and schedule of batch sizes $\mathcal B=B_1,B_2,\ldots,B_{E}$
\Require Loss function $\Ls$
\For{$e$ in $1,2\ldots,E$}
    \State $\hat B \gets \mathcal B[e]$ \Comment{Get current batch size}
    \State $\kappa \gets \hat B/B$      \Comment{Compute scaling factor}
    \State $\hat\eta \gets \kappa \eta$ \Comment{Get scaled learning rate}
    \State $\hat\rho \gets \rho^\kappa$ \Comment{Get scaled momentum}
    \For{$b$ in $1,2\ldots,\text{floor}(E/\hat B)$}
        \State Sample a minibatch of $\hat B$ samples $\mathcal X=\{\vx^{(1)},\ldots,\vx^{(\hat B)}\}$
        \State $\rvtheta \gets \rvtheta - 
        (\hat \eta / \hat B)
        \sum_{x\in\mathcal X} \nabla_{\rvtheta} \Ls(x;\rvtheta,\rvzeta)$ \Comment{SGD Update}
        \State $\rvzeta \gets \hat\rho \,\rvzeta+(1-\hat\rho)\,\rvtheta$ \Comment{EMA Update}
    \EndFor
\EndFor
\end{algorithmic}
\end{algorithm}

In \Cref{alg:progressive-scaling}, the prefactor of the \gls{sgd} update could also have been written $\eta/B$, although an equivalent use of the base momentum is not possible.

Finally, we outline how to extend \Cref{alg:progressive-scaling} to more complex setups, like those presented in 
\Cref{subsec:self-supervised}:
\begin{enumerate}[leftmargin=0.75cm]
    \item Optimizer scaling rules are used appropriately, for example the Adam scaling rule in case of using the Adam optimizer to update parameters $\rvtheta$.
    \item Schedules for hyperparameters are computed using the base hyperparameters, and are then modified by application of the scaling law in epoch (outer) loop.
    \item Schedules for hyperparameters at the \emph{step} rather than epoch level can be achieved in practice through recomputing the schedule and updating the notion of minibatch index appropriately throughout training.
\end{enumerate}
All of the above techniques are used in \Cref{subsec:self-supervised}.
In addition, scheduling batch sizes within epoch is possible, providing one maintains a notion of computation within some fixed continuous time $T_{\text{fixed}}$. 
We did not investigate this scenario.

\section{EMA approximation theorems with SDEs}
\label{app:ema-approximation-theorem}
\subsection{SGD with model EMA}

We will now derive the EMA scaling rule when tracking model parameters and the model is trained using SGD. We employ a strategy similar to \citet{DBLP:conf/nips/MalladiLPA22}, where we associate to each iterative process a Stochastic Differential Equation (SDE). In order to control the distance between the SDE and the discrete process, we use the tools from \citet{li2019stochastic}.
\begin{definition}[Polynomial growth, Definition 1 in \citep{li2019stochastic}]
\label{def:polynomial-growth}
The set $G$ is the set of continuous functions $\sR^d\to\sR$ with at most polynomial growth, i.e., for $g\in G$ there exists two scalars $\kappa_1, \kappa_2>0$ such that for all $\vx\in\sR^d$, we have $|g(\vx)|\leq \kappa_1(1+\|\vx\|^{\kappa_2})$.

For an integer $\alpha >0$, $G^\alpha$ is the set of functions $\sR^d\to \sR$ that are $\alpha$-times continuously differentiable and such that all their derivatives up to order $\alpha$ are in $G$.
\end{definition}

Similarly to \citet{DBLP:conf/nips/MalladiLPA22}, we use Noisy Gradient Oracle with Scale Parameter (NGOS) to define the update rules on the parameters.
\begin{definition}[Noisy Gradient Oracle with Scale Parameter (NGOS), adaptation of \citep{DBLP:conf/nips/MalladiLPA22}]
\label{def:ngos}
A NGOS is a tuple $\mathcal{G}_\sigma = (f, \mSigma, \mathcal{Z}_\sigma)$. Given a noise scale parameter $\sigma >0$, the NGOS $\mathcal{G}_\sigma$ takes as input the parameters $\vtheta$ and outputs a random vector $\rvg= \nabla f(\vtheta, \vzeta) + \sigma \rvepsilon$ where $\nabla f(\vtheta, \vzeta)$ is the gradient of $f$ with respect to $\vtheta$ at $\inParentheses{\vtheta, \vzeta}$, and $\rvepsilon$ is a random vector drawn from the distribution $\mathcal{Z}_\sigma(\vtheta, \vzeta)$ with zero mean and covariance $\mSigma(\vtheta, \vzeta)$.
\end{definition}

Note that in the above definition, the probability distribution $\mathcal{Z}_\sigma(\vtheta, \vzeta)$ is allowed to change with the scale $\sigma$, but its first two moments --- its mean and its covariance --- are fixed with $\sigma$. We have the following theorem for model EMA under optimization with SGD:
\begin{theorem}[SDE for SGD + EMA]
    \label{thm:app:sde}
Consider the couple $\rvx_k = (\rvtheta_k, \rvzeta_k)$ where $\rvtheta_k$ are the iterates of SGD with a NGOS (\Cref{def:ngos}) and $\rvzeta_k$ is an EMA of $\rvtheta_k$, defined, starting from $\rvx_0 = \vx_0$, by
\begin{align}
    \rvtheta_{k+1} &= \rvtheta_{k} - \eta \rvg_k, \enspace \text{with } \rvg_k=\nabla f(\rvtheta_k, \rvzeta_k) + \sigma \rvepsilon_k , \text{ and }\rvepsilon_k \sim \mathcal{Z}_\sigma(\rvtheta_k, \rvzeta_k),\\
    \rvzeta_{k+1} &= \rho \rvzeta_k + (1-\rho) \rvtheta_k\enspace.
\end{align}
Define $\beta_0 = (1-\rho) / \eta$, $\sigma_0 = \sigma\sqrt{\eta}$, and define the SDE for $X_t = (\Theta_t, Z_t)$, starting from $X_0 = \vx_0$, by
\begin{align}
\label{eq:app:sde-sgd}
    d\Theta_t &= - \nabla f(\Theta_t, Z_t)dt + \sigma_0\mSigma(\Theta_t, Z_t)^{\frac12}dW_t,\enspace\text{with}\enspace W_t \text{ a Wiener process}\\
    \label{eq:app:sde-ema}
    dZ_t &= \beta_0(\Theta_t - Z_t)dt\enspace.
\end{align}
Assume that $f$ is continuously differentiable, with $f\in G^3$ and $\mSigma^{\frac12}\in G^2$ (\Cref{def:polynomial-growth}). Then, for any time horizon $T >0$ and test function $g\in G^2$ , there exists a constant $c>0$ such that 
\begin{equation}
    \max_{k=0,\dots, \lfloor T /\eta \rfloor} |\mathbb{E}[g(X_{\eta k})] - \mathbb{E}[g(\rvx_k)]| \leq c\times  \eta \enspace.
\end{equation}
\end{theorem}
\begin{proof}
The proof uses the same tools as in \citet{li2019stochastic}. Define $\Delta(\vtheta, \vzeta) = \eta (-\nabla f(\vtheta, \vzeta) + \sigma \rvepsilon, \beta_0(\vtheta- \vzeta))$ with $\rvepsilon\sim \mathcal{Z}_{\sigma}(\vtheta, \vzeta)$ the one-step update for the SGD + EMA update, such that $\rvx_{k+1} = \rvx_k +\Delta(\rvx_k)$. We have the first two moments:
\begin{align}
    \mathbb{E}[\Delta(\vtheta, \vzeta)] &= \eta (-\nabla f(\vtheta, \vzeta), \beta_0(\vtheta- \vzeta))\\
    \mathbb{V}[\Delta(\vtheta, \vzeta)] &= \eta \sigma_0^2\begin{bmatrix}\mSigma(\vtheta, \vzeta) & 0 \\ 0 & 0 \end{bmatrix}
\end{align}
and the higher-order moments are $O(\eta^2)$.
Similarly, let $\tilde{\Delta}(\vtheta, \vzeta)$ be the solution at time $\eta$ of the SDE defined by \Cref{eq:sde-sgd} starting from $X_0=(\vtheta, \vzeta)$. From Ito's formula, we also obtain
\begin{align}
    \mathbb{E}[\tilde{\Delta}(\vtheta, \vzeta)] &= \eta (-\nabla f(\vtheta), \beta_0(\vtheta- \vzeta))\\
    \mathbb{V}[\tilde{\Delta}(\vtheta, \vzeta)] &= \eta \sigma_0^2\begin{bmatrix}\mSigma(\vtheta, \vzeta) & 0 \\ 0 & 0 \end{bmatrix}
\end{align}
and the higher-order moments are $O(\eta^2)$.
Hence, the moments of the discrete iteration and of the SDE match up to second order. 
Following the same proof technique as in~\citet{li2019stochastic} then leads to the advertized theorem.
\end{proof}
This theorem is a simple adaptation of the results of~\cite{li2019stochastic}. Intuitively, it is expected that $X_t$ and $\rvx_k$ are close since $\rvx_k$ is the Euler-Maruyama discretization of $X_t$ with learning rate $\eta$. 
We then have the corollary.
\begin{corollary}[Validity of the \gls{ema} Scaling Rule]
    Assume that $f$ is continuously differentiable, with $f\in G^3$ and $\mSigma^{\frac12}\in G^2$. 
    Let $\rvtheta_k^{B}, \rvzeta_k^{B}$ the iterates of the \Cref{eq:iterations} with batch size $B$ and hyperparameters $\eta, \rho$. 
    Let $\rvtheta_k^{\kappa B}, \rvzeta_k^{\kappa B}$ be iterates with batch size $\kappa B$, learning rate $\eta$ determined by the \gls{sgd} Scaling Rule (\Cref{def:lsr}) and momentum determined by the \gls{ema} Scaling Rule, linear version (\Cref{def:ema-sr}). 
    Then, for any time horizon $T >0$ and function $g\in G^2$, there exists a constant $d>0$ such that 
   \begin{equation}
       \max_{k=0,\dots, \lfloor T /\eta \rfloor} |\mathbb{E}[g(\rvtheta_{\lfloor k / \kappa \rfloor}^{\kappa B}, \rvzeta_{\lfloor k / \kappa \rfloor}^{\kappa B})] - \mathbb{E}[g(\rvtheta_k, \rvzeta_k)]| \leq d\times  \eta \enspace.
   \end{equation}
\end{corollary}
\begin{proof}
    The proof is similar to~\citet{DBLP:conf/nips/MalladiLPA22}. Under the scaling rule, both $\rvx_k = (\rvtheta_k, \rvzeta_k)$ and $\hat{\rvx}_{\lfloor k / \kappa \rfloor}= (\rvtheta_{\lfloor k / \kappa \rfloor}^{\kappa B}, \rvzeta_{\lfloor k / \kappa \rfloor}^{\kappa B})$ have the same limiting SDE. Hence we have from the previous theorem that for all test function $g$, we can find $c, c'$ such that  
    \begin{equation}
        \max_{k=0,\dots, \lfloor T /\eta \rfloor} |\mathbb{E}[g(X_{\eta k})] - \mathbb{E}[g(\rvx_k)]| \leq c\times  \eta \text{ and }\max_{k=0,\dots, \lfloor T /\eta \rfloor} |\mathbb{E}[g(X_{\eta k})] - \mathbb{E}[g(\hat{\rvx}_{\lfloor k / \kappa \rfloor})]| \leq c'\times  \eta.
    \end{equation}
    The triangle inequality then gives
    \begin{equation}
        \max_{k=0,\dots, \lfloor T /\eta \rfloor} |\mathbb{E}[g(\hat{\rvx}_{\lfloor k / \kappa \rfloor})]- \mathbb{E}[g(\rvx_k)]|\leq (c + c') \times \eta.
    \end{equation}
    Hence, taking $d = c+ c'$ gives the expected result.
\end{proof}

\subsection{Adaptive gradient methods with model EMA}

We now turn to the case where one uses an adaptive gradient method rather than SGD to train the model.
We follow derivations similar to those of \citet{DBLP:conf/nips/MalladiLPA22}, with an added EMA. 
Like above, we consider that the loss function $f$ also depends on the EMA tracking parameter $\rvzeta_k$.
We begin with RMSProp with EMA, which iterates:
\begin{align}
    \rvv_{k+1} &= \gamma \rvv_k + (1 - \gamma) \rvg_k^2, \enspace \text{with } \rvg_k=\nabla f(\rvtheta_k, \rvzeta_k) + \sigma \rvepsilon_k , \text{ and }\rvepsilon_k \sim \mathcal{Z}_\sigma(\rvtheta_k, \rvzeta_k),\label{app:eq:rmsprop_v}\\
    \rvtheta_{k+1} &= \rvtheta_k - \eta (\sqrt{\rvv_k} + \varepsilon)^{-1} \times \rvg_k\\
    \rvzeta_{k+1} &= \rho \rvzeta_k + (1- \rho) \rvtheta_k.
\end{align}

Like in \citet{DBLP:conf/nips/MalladiLPA22}, we place ourselves in the high noise regime, in which the term $\rvg_k^2$ in \Cref{app:eq:rmsprop_v} is approximated by $\rvg_k^2\simeq \sigma^2\mathrm{diag}(\Sigma(\rvtheta_k, \rvzeta_k))$. We use the same scaling rules, with an additional one for $\rho$:
\begin{equation}
    \label{app:eq:scaling_rmsprop}
    \gamma_0 = (1 - \gamma) / \eta^2,\enspace \sigma_0 = \sigma\eta, \enspace \varepsilon_0 = \varepsilon \eta, \text{ and } \beta_0 = (1-\rho) / \eta^2,
\end{equation}
and we let $\rvu_k = \rvv_k / \sigma^2$. The equations for RMSProp with EMA then become, using only these new variables and $\eta$:
\begin{align}
    \rvu_{k+1} - \rvu_k &= \eta^2 \gamma_0 (\mathrm{diag}(\Sigma(\rvtheta_k, \rvzeta_k)) - \ru_k), \\
    \rvtheta_{k+1} - \rvtheta_k &= - (\sqrt{\rvu_k} + \varepsilon_0)^{-1}\left(\eta^2  \nabla f(\rvtheta_k, \rvzeta_k) + \eta \rvepsilon_k \right)\\
    \rvzeta_{k+1} -\rvzeta_k &= \eta^2\beta_0 (\rvtheta_k - \rvzeta_k).
\end{align}
This formulation makes it clear that these iterations can be seen as the discretization of the SDE
\begin{align}
    dU_t &=  \gamma_0 (\mathrm{diag}(\Sigma(\Theta_t, Z_t)) - U_t)dt, \\
    d\Theta_t &= - (\sigma_0\sqrt{U_t} + \varepsilon_0)^{-1} (\nabla f(\Theta_t, Z_t) dt + \sigma_0 \Sigma(\Theta_t, Z_t)^{1/2}dWt) \\
    dZ_t &= \beta_0 (\Theta_t - Z_t)dt,
\end{align}
with step size $\eta^2$.
Of course, we recover the SDE of \citet{DBLP:conf/nips/MalladiLPA22} in the case where $\beta_0 = 0$.
A formal proof of closeness between the iterates and the SDE trajectory is out of the scope of the present paper since it would imply redoing much of the theoretical work developed in \citet{DBLP:conf/nips/MalladiLPA22}.
Still, the previous informal analysis hints that for RMSProp, the scaling rule in \Cref{app:eq:scaling_rmsprop} should be used.
In other words, given a certain set of hyperparameters $\gamma, \eta$ and $\rho$, if the batch size goes from $B$ to $\hat{B} = \kappa \times B$, the noise level becomes $\hat{\sigma} = \sigma / \sqrt{\kappa}$, and keeping the quantities in \Cref{app:eq:scaling_rmsprop} constant means that we should use as new hyperparameters
$$
\hat{\gamma} = 1 - (1-\gamma) \times \kappa,\enspace \hat{\eta} = \eta \times \sqrt{\kappa}, \text{ and } \hat{\rho} = 1 - (1 - \rho)\times \kappa\enspace.
$$
The linear rule $ \hat{\rho} = 1 - (1 - \rho)\times \kappa$ is at the first order equivalent to the exponential scaling rule $\hat{\rho} = \rho^{\kappa}$.
Hence, even though the limiting SDE differs greatly from that of SGD, and even though the scaling rule regarding the learning rate differs, we recover for the momentum term $\rho$ the exact same scaling rule as for SGD.

We finish the discussion with the case of Adam, which leads once again to the same rule as for SGD.
Adam with EMA tracking of the network parameters iterates
\begin{align}
    \rvm_{k+1} &=\beta_1\rvm_k + (1 - \beta_1) \rvg_k, \enspace \text{with } \rvg_k=\nabla f(\rvtheta_k, \rvzeta_k) + \sigma \rvepsilon_k , \text{ and }\rvepsilon_k \sim \mathcal{Z}_\sigma(\rvtheta_k, \rvzeta_k),\\
    \rvv_{k+1} &= \beta_2 \rvv_k + (1 - \beta_2) \rvg_k^2 \\
    \tilde{\rvm}_{k+1} &= \rvm_{k+1} / (1 - \beta_1^{k+1})\\ 
    \tilde{\rvv}_{k+1} &= \rvv_{k+1} / (1 - \beta_2^{k+1})\\ 
    \rvtheta_{k+1} &= \rvtheta_k - \eta (\sqrt{\tilde{\rvv}_{k}} + \varepsilon)^{-1} \times \tilde{\rvm}_{k+1}\\
    \rvzeta_{k+1} &= \rho \rvzeta_k + (1- \rho) \rvtheta_k\enspace.
\end{align}
 Here, we use the same minor modification of the iterations as in \citet{DBLP:conf/nips/MalladiLPA22}, where we use $\rvv_k$ instead of $\rvv_{k+1}$ in the denominator of the $\rvtheta_k$ update.

We consider the following scaling for the hyperparameters
\begin{equation}
    \label{app:eq:scaling_adam}
    c_1 = (1 - \beta_1)/ \eta^2,\enspace c_2 = (1- \beta_2) / \eta^2,\enspace \sigma_0 = \sigma\eta, \enspace \varepsilon_0 = \varepsilon \eta, \text{ and } \beta_0 = (1 - \rho) / \eta^2,
\end{equation}
and $\gamma_1(t) = 1 - \exp(-c_1t)$, $\gamma_2(t) = 1 - \exp(-c_2t)$, and $\rvu_k = \rvv_k / \sigma^2 $. The SDE for Adam + EMA is given by
\begin{align}
    dM_t &=c_1\left((\nabla f(\Theta_t, Z_t) - M_t)dt + \sigma_0\Sigma(\Theta_t, Z_t)^{1/2}dW_t\right)\\
    dU_t &=c_2(\mathrm{diag}(\Sigma(\Theta_t, Z_t)) - U_t)dt\\
    d\Theta_t &= -  \frac{\sqrt{\gamma_2(t)}}{\gamma_1(t)}(\sigma_0\sqrt{U_t} + \varepsilon_0\sqrt{\gamma_2(t)})^{-1} \times M_t dt\\
    dZ_t &= \beta_0(\Theta_t - Z_t)dt.
\end{align}

This is once again the same SDE as in \citet{DBLP:conf/nips/MalladiLPA22} with the added EMA term. Like previously, this SDE hints at the fact that the scaling rule in \cref{app:eq:scaling_adam} should be used. In other words, given a set of hyperparameters $\beta_1,\beta_2, \eta$, and $\rho$, if the batch size goes from $B$ to $\kappa \times B$, then the noise level becomes $\hat{\sigma} = \sigma/ \sqrt{\kappa}$ and keeping quantities in \cref{app:eq:scaling_adam} constant means that we should use as new hyperparameters
$$
\hat{\beta}_1 = 1 - (1 - \beta_1)\times \kappa,\enspace 
\hat{\beta}_2 = 1 - (1 - \beta_2)\times \kappa,\enspace 
\hat{\eta} = \eta \times \sqrt{\kappa},\text{ and } \hat{\rho} = 1 - (1-\rho)\times \kappa.
$$
We once again recover a linear rule for $1 - \rho$ which is equivalent to the exponential scaling rule $\hat{\rho} = \rho^\kappa$ in the limit $\rho\to 0$.

\section{Additional proofs}

\subsection{Iterations of SGD + EMA}
\label{app:matrix-calculations}

Here we derive a critical component of the \gls{ema} Scaling Rule, 
the matrix equation of \Cref{eq:scalingRuleSummaryEquation} from which the \gls{ema} Scaling Rule (\Cref{def:ema-sr}) follows.
\begin{theorem}[Iterations of SGD + EMA]
\label{thm:app:sgd-ema-iterations}
Assuming that gradients change slowly 
over iterations of \gls{sgd} (\Cref{def:sgd}) and \gls{ema} (\Cref{def:ema}):
$\nabla_{\rvtheta}\Ls(x;\rvtheta_{t+j},\rvzeta_{t+j})\approx \nabla_{\rvtheta}\Ls(x;\rvtheta_{t},\rvzeta_{t})\approx \rvg$, for 
$j=1,2,\ldots,\kappa$ and
representative gradient $\rvg$,
iterating over $\kappa$ independent minibatches produces model states
\begin{align}
\begin{bmatrix}
\rvtheta_{t+\kappa}
\\
\rvzeta_{t+\kappa}
\\
\rvg
\end{bmatrix}
=
\begin{bmatrix}
1 & 0 & -\eta \\
1-\rho & \rho & 0\\
0 & 0 & 1
\end{bmatrix}^\kappa
\cdot 
\begin{bmatrix}
\rvtheta_{t}
\\
\rvzeta_{t}
\\
\rvg
\end{bmatrix}
=
\begin{bmatrix}
\rvtheta_{t}-\eta\,\kappa \,\rvg
\\
\rho^\kappa \, \rvzeta_{t}
+(1-\rho^\kappa) \, \rvtheta_t
+\mathcal O\left(\eta\times \beta_\rho\right)
\\
\rvg
\end{bmatrix}.
\end{align}
\end{theorem}

\begin{proof}
First note that
for matrices of the form
\begin{equation}
    \mA
    =
    \begin{bmatrix}
    1 & 0 & a_{0,2} \\
    1-a_{1,1} & a_{1,1} & 0\\
    0 & 0 & 1
    \end{bmatrix},
\end{equation}
their multiplication follows
\begin{align}
    \mA\,\mB
    &=
    \begin{bmatrix}
    1 & 0 & a_{0,2} \\
    1-a_{1,1} & a_{1,1} & 0\\
    0 & 0 & 1
    \end{bmatrix}
    \begin{bmatrix}
    1 & 0 & b_{0,2} \\
    1-b_{1,1} & b_{1,1} & 0\\
    0 & 0 & 1
    \end{bmatrix}\nonumber\\
    &=
    \begin{bmatrix}
    1 & 0 & a_{0,2}+b_{0,2} \\
    1 - a_{1,1}\,b_{1,1} & a_{1,1}\,b_{1,1} & (1-a_{1,1})\,b_{0,2}\\
    0 & 0 & 1
    \end{bmatrix},
\end{align}
and
\begin{align}
    \mA\,\mB\,\mC
    &=
    \begin{bmatrix}
    1 & 0 & a_{0,2}+b_{0,2} \\
    1 - a_{1,1}\,b_{1,1} & a_{1,1}\,b_{1,1} & (1-a_{1,1})\,b_{0,2}\\
    0 & 0 & 1
    \end{bmatrix}
    \begin{bmatrix}
    1 & 0 & c_{0,2} \\
    1-c_{1,1} & c_{1,1} & 0\\
    0 & 0 & 1
    \end{bmatrix}\nonumber\\
    &=
    \begin{bmatrix}
    1 & 0 & a_{0,2}+b_{0,2}+c_{0,2} \\
    1 - a_{1,1}\,b_{1,1}\,c_{1,1} & a_{1,1}\,b_{1,1}\,c_{1,1} & 
    (1-a_{1,1})\,b_{0,2}+(1-a_{1,1}\,b_{1,1})\,c_{0,2}\\
    0 & 0 & 1
    \end{bmatrix}.
\end{align}
By induction
\begin{align}
    \mA^\kappa=
    \begin{bmatrix}
    1 & 0 & \kappa \times a_{0,2} \\
    1 - a_{1,1}^\kappa & a_{1,1}^\kappa & 
    \delta(a_{0,2}, a_{1,1},\kappa)\\
    0 & 0 & 1
    \end{bmatrix},
\end{align}
where
\begin{align}
\delta(a_{0,2}, a_{1,1},\kappa)
=a_{0,2}\,\sum_{i=1}^{\kappa - 1}\left(1-a_{1,1}^i\right)
&=a_{0,2}\left(\kappa-\frac{1-a_{1,1}^\kappa}{1-a_{1,1}}\right), \quad \text{for} \, a_{1,1}\neq 1.
\end{align}
It follows that
\begin{align}
\begin{bmatrix}
1 & 0 & -\eta \\
1-\rho & \rho & 0\\
0 & 0 & 1
\end{bmatrix}^\kappa
=
\begin{bmatrix}
1 & 0 & -\kappa\,\eta \\
1-\rho^\kappa & \rho^\kappa & \delta(\eta,\rho,\kappa)\\
0 & 0 & 1
\end{bmatrix}
\end{align}
where the \gls{ema} Scaling Rule error
\begin{align}
    \delta(\eta,\rho,\kappa)
    &=
    \inParentheses{-\eta}\left(\kappa-\frac{1-\rho^\kappa}{1-\rho}\right)
    \approx
    \inParentheses{-\eta}\left(\kappa-\kappa +\mathcal{O}(\beta_\rho)\right)=0+\mathcal{O}(\eta\times\beta_\rho),
    \label{eq:scaling-error}
\end{align}
where $\beta_\rho\equiv 1 - \rho$ and the approximation is around $\rho=1$.
\end{proof}

\subsection{Limiting behavior of Polyak-Ruppert averaging}
\label{app:asymptoticAnalysis}
Here we sketch the asymptotic behavior of a target model $\rtheta$ and its \gls{ema} $\rzeta$.
Let us assume that $\rtheta$ converges to the stationary distribution $\lim_{t\to\infty}\rtheta_t=\rtheta^*$, $\rtheta^*\sim p_{\infty}(\rtheta)$.
We are interested in statistical properties of $\rzeta^*=\lim_{t\to\infty}\rzeta_t$,
as this will formalize the notion of how the \gls{ema} depends on the a time-horizon defined by its momentum $\rho$ as discussed in \Cref{tab:different-ema}.

As a warm-up, for $n$ independent random variables $\rx_1,\ldots,\rx_2$, we know that the sample mean $\bar x=\frac1n(x_1,x_2,\ldots,x_n)$
has the statistical properties
\begin{align}
    \E[\bar x]
    &=\mu,
    &
    \Var[\bar x]
    &=
    \frac{\sigma^2}{n},
    \label{eq:sum-rvs-1}
\end{align}
where $\mu$ and $\sigma$ are the population mean and variance.
This gives us an idea of what to expect.
As we will now show,
the expectation of $\rzeta^*$ should have no time-horizon dependence, whereas the variance of $\rzeta^*$ will depend on its time horizon (i.e. the number of samples it integrates over) which is defined by $\rho$.

In the case of a weighted sum
\begin{align}
    \bar x^{(w)}
    &=\sum_{i=1}^nw_i\,x_i,
\end{align}
then if the $x_i$ are \gls{iid}, then
\begin{align}
    \E[\bar x^{(w)}]
    &=\sum_{i=1}^nw_i\,\E[x_i]
    =n\,\bar w\,\mu,
    &
    \bar w
    &=
    \frac 1n\sum_{i=1}^nw_i,
    \label{eq:app-mean-iid}
\end{align}
and for the variance \citep{KishDesignEffect}
\begin{align}
    \Var[\bar x^{(w)}]
    &=
    n
    \cdot
    \widebar{w^2}
    \cdot
    \sigma^2
    &
    \widebar{w^2}
    &=
    \frac1n\sum_{i=1}^nw_i^2,
    &
    \sigma^2
    &=
    \Var[x_i].
\end{align}
We can verify that we reproduce the well-known result in \Cref{eq:sum-rvs-1} in the case where all weights are equal to $\frac{1}{n}$ as follows
\begin{equation}
    \forall i: w_i=\frac1n 
    \implies
    \widebar{w^2}=
    \frac{1}{n}
    \cdot
    \sum_{i = 1}^n
    \inParentheses{\frac{1}{n}}^2
    =
    \frac{1}{n^2}
    \implies
    \Var[\bar x^{(w)}]
    =
    n\cdot
    \frac{1}{n^2}
    \cdot
    \sigma^2
    =
    \frac{\sigma^2}{n}.
\end{equation}
In the case of an exponential moving average we have
    \begin{equation}
    \rzeta_{t+1}
    =
    \rho \,\rzeta_{t}
    +
    (1-\rho)\,\rtheta_{t}
    =
    \rho^t \,\rzeta_{1}
    +
    (1-\rho)
    \sum_{i=0}^{t-1}
    \rho^{i}
    \rtheta_{t-i}.
\end{equation}
Let's consider the specific case where we are at iteration $k$ which is sufficiently large that $\rzeta$ and $\rtheta$ have converged to their stationary distributions.
From $k$, the iterations unfold as
    \begin{equation}
    \rzeta_{t+1}
    =
    \rho^{t+1-k} \,\rzeta_{k}
    +
    (1-\rho)
    \sum_{i=0}^{t-k}
    \rho^{i}
    \rtheta_{t-i}.
\end{equation}
We rearrange for terms in $\rzeta$
\begin{align}
    \rzeta_{t+1}
    -\rho^{t+1-k}\,\rzeta_k
    =
    (1-\rho)
    \sum_{i=0}^{t-k}
    \rho^{i}\,
    \rtheta_{t-i},
    \label{eq:app-zeta-arranged}
\end{align}
and before proceeding to the final result, using $n=t+1-k$, we compute the convenient quantities
\begin{align}
    \bar\rho
    &=\frac1{n}\sum_{i=0}^{n-1}\rho^i
    =\frac1{n}\times\frac{1-\rho^n}{1-\rho}\\
    \widebar{\rho^2}
    &=\frac1{n}\sum_{i=0}^{n-1}\rho^{2i}
    =\frac1{n}\times\frac{1-\rho^{2n}}{1-\rho^2}.
\end{align}
Taking expectation of \Cref{eq:app-zeta-arranged} and setting statistics to their stationary values, we have
\begin{align}
    (1-\rho^{n})\,\E[\rzeta^*]
    =
    (1-\rho)\,n\,\bar\rho\,\E[\rtheta^*]=
    (1-\rho^n)\,\E[\rtheta^*],
\end{align}
where we have used the result in \Cref{eq:app-mean-iid}.
It follows that for $\rho\neq 1$ we have
\begin{align}
    \E[\rzeta^*]=\E[\rtheta^*],
\end{align}
independent of $\rho$.
Finally, we can take the variance of \Cref{eq:app-zeta-arranged}.
First the left hand side
\begin{align}
    \Var\left[
    \rzeta_{t+1}
    -\rho^{n}\,\rzeta_k\right]
    =
    \Var\left[
    \rzeta_{t+1}\right]
    +
    \rho^{2n}\,
    \Var\left[\rzeta_k\right]
    =
    \left(1+\rho^{2n}\right)\,\Var\left[\rzeta^*\right].
\end{align}
Next the right hand side
\begin{align}
    \Var\left[
    (1-\rho)
    \sum_{i=0}^{n-1}
    \rho^{i}\,
    \rtheta_{t-i}
    \right]
    =
    (1-\rho)^2
    \,
    \Var\left[
    \sum_{i=0}^{n-1}
    \rho^{i}\,
    \rtheta_{t-i}
    \right]
    =
    (1-\rho)^2
    \cdot
    \inParentheses{
        \frac{1 - \rho^{2n}}{1 - \rho^2}
    }
    \cdot
    {\Var[\rtheta^*]}
    .
\end{align}
Finally,
equating left and right hand sizes and rearranging for $\Var[\rzeta^*]$ gives
\begin{align}
    \Var\left[\rzeta^*\right]
    =
    \frac{1 - \rho^{2n}}{1 + \rho^{2n}}
    \cdot
    \frac{1 - \rho}{1 + \rho}
    \cdot
    \Var\left[\rtheta^*\right]
    \label{eq:app-var-zeta}
\end{align}
In the limit $t\to\infty$, the momentum-dependent prefactor becomes
\begin{align}
    \lim_{t\to\infty}
    \inParentheses{
        \frac{1 - \rho^{2n}}{1 + \rho^{2n}}
        \cdot
        \frac{1 - \rho}{1 + \rho}
    }
    &=
    \frac{1-\rho}{1+\rho}
    \implies
    \lim_{t\to\infty}
    \Var\left[\rzeta^*\right]
    =
    \frac{1-\rho}{1+\rho}
    \cdot
    \Var\left[\rtheta^*\right].
    \label{eq:app-var-zeta-prefactor}
\end{align}
\Cref{eq:app-var-zeta,eq:app-var-zeta-prefactor} validate our intuition.
When $\rho\rightarrow0$, then $\rzeta$ behaves like $\rtheta$ independent of $T$, with their variance and expectation matching.
When $\rho>0$, the momentum-dependent prefactor serves as an aggregator over the history when $t$ is sufficiently large compared to $k$, reducing the variance $\Var[\zeta^*]$ but preserving its expectation.
This formalizes the notion of time horizon discussed in \Cref{tab:different-ema}.

\section{Additional details and results for Polyak-Ruppert averaging}
\label{app:polyak}

\paragraph{Additional background} Polyak-Ruppert averaging (\Cref{def:polyak-ruppert-average}) is a simplification of \gls{swa} \citep{DBLP:conf/uai/IzmailovPGVW18} which uses a more complex multi-cycle schedule based weighting of the model parameters. Both \Cref{def:polyak-ruppert-average} and \gls{swa} present similar favorable properties like wider minima and better generalization \citep{DBLP:conf/uai/IzmailovPGVW18}. 
For example, \citet{DBLP:conf/cvpr/HeCXLDG22} observed that a supervised ViT-H/14 overfits on ImageNet1k 
\citep{DBLP:journals/corr/RussakovskyDSKSMHKKBBF14} without a model \gls{ema}, achieving an accuracy of 80.9\%.
Equipping a Polyak-Ruppert average ($\rho=0.9999$) alleviated overfitting and gave a 83.1\% accuracy.

\paragraph{Organization} In this appendix, we look at additional momenta for one-dimensional noisy parabola, as well as extensions to $D$-dimensions (\Cref{app:noisy-parabola}), provide a more detailed view of the results of \Cref{subsec:supervised-polyakking} (\Cref{app:subsec:polyak-image-classification}), and investigate the scenario where the \gls{ema} Scaling Rule (\Cref{def:ema-sr}) is applied to batch normalization \citep{DBLP:conf/icml/IoffeS15} coefficients (\Cref{subsec:polyak-bn}).

\subsection{Noisy parabola}
\label{app:noisy-parabola}

\paragraph{Additional one-dimensional examples}

\begin{figure}[ht]
     \centering
     \begin{subfigure}[b]{0.45\textwidth}
         \centering
         \includegraphics[width=\textwidth]{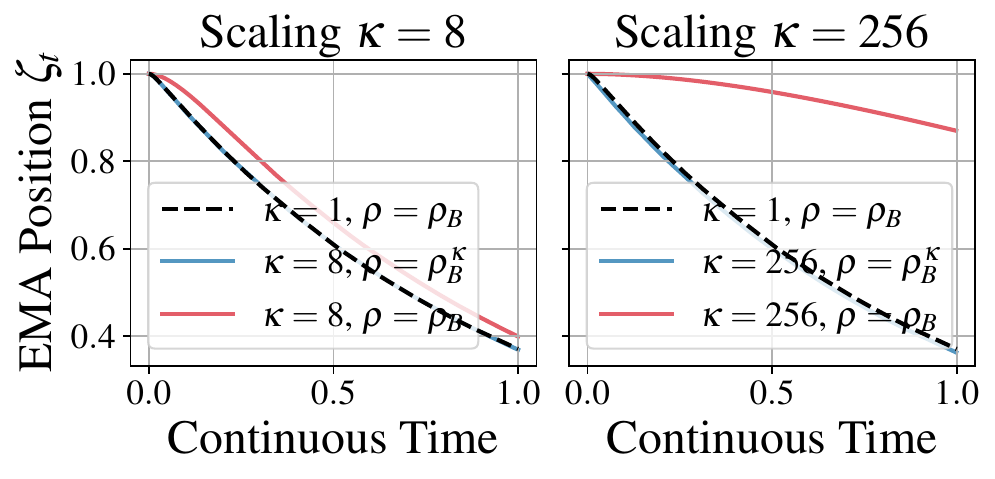}
         \caption{Trajectory of the model \gls{ema} $\rvzeta$ under different scalings $\kappa$, with $\rho_B=0.99$, $\eta_B=10^{-4}$.}
     \end{subfigure}
     \hfill
     \begin{subfigure}[b]{0.52\textwidth}
         \centering
         \includegraphics[width=\textwidth]{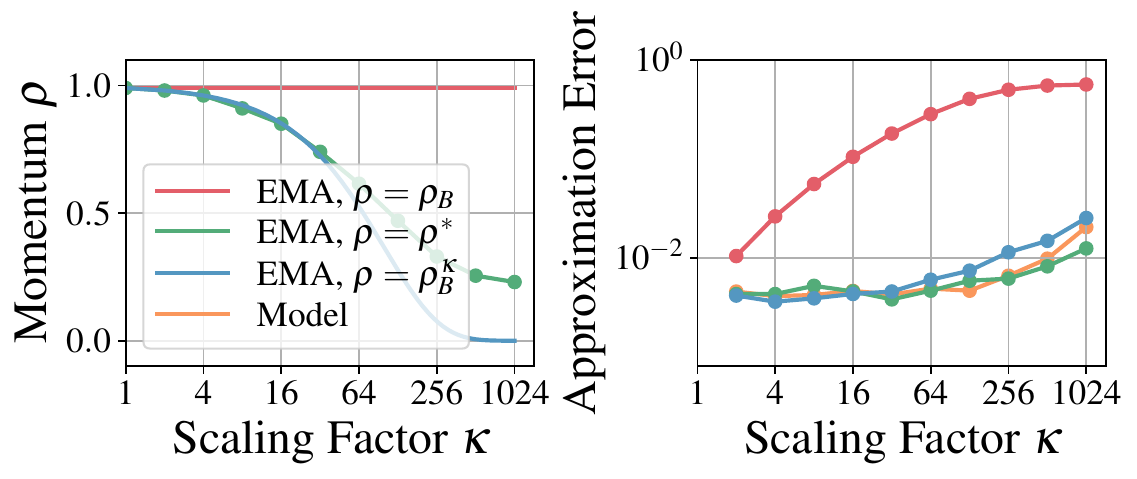}
         \caption{Choices for momentum (left) with corresponding approximation errors (\Cref{eq:optimal-momentum}) (right).}
     \end{subfigure}
     \caption{(a) We show the effect of scaling by comparing model \gls{ema} trajectories of the baseline 
     ($\kappa=1$, black dashed) to $\kappa=8$ (left) and $\kappa=256$ (right), 
     with ($\rho=\rho_B^\kappa$, blue) and without 
     ($\rho=\rho_B$, red) the EMA Scaling Rule.
     (b, left) The momentum according for different scaling rules and the empirically optimal $\rho^*$ (\Cref{eq:optimal-momentum}).
     (b, right) The approximation error (\Cref{eq:optimal-momentum}) of trajectories in (b, left) and the target model (orange). 
     Error for $\rho^*$ is computed using a hold-out to mitigate overfitting.
     }
     \label{fig:app-toy-experiment-1d-0-99}
\end{figure}

\begin{figure}[ht]
     \centering
     \begin{subfigure}[b]{0.45\textwidth}
         \centering
         \includegraphics[width=\textwidth]{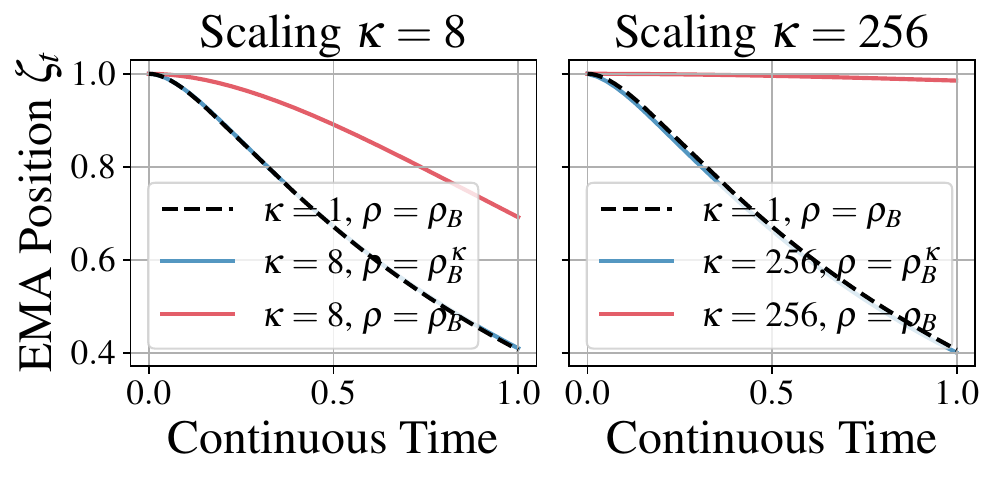}
         \caption{Trajectory of the model \gls{ema} $\rvzeta$ under different scalings $\kappa$, with $\rho_B=0.999$, $\eta_B=10^{-4}$.}
     \end{subfigure}
     \hfill
     \begin{subfigure}[b]{0.52\textwidth}
         \centering
         \includegraphics[width=\textwidth]{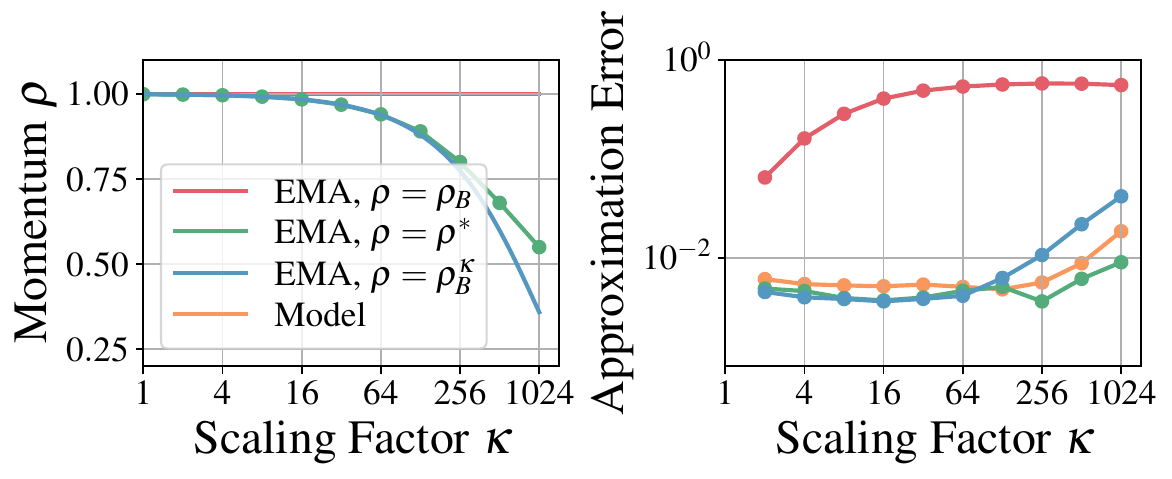}
         \caption{Choices for momentum (left) with corresponding approximation errors (\Cref{eq:optimal-momentum}) (right).}
     \end{subfigure}
     \caption{(a) We show the effect of scaling by comparing model \gls{ema} trajectories of the baseline 
     ($\kappa=1$, black dashed) to $\kappa=8$ (left) and $\kappa=256$ (right), 
     with ($\rho=\rho_B^\kappa$, blue) and without 
     ($\rho=\rho_B$, red) the EMA Scaling Rule.
     (b, left) The momentum according for different scaling rules and the empirically optimal $\rho^*$ (\Cref{eq:optimal-momentum}).
     (b, right) The approximation error (\Cref{eq:optimal-momentum}) of trajectories in (b, left) and the target model (orange). 
     Error for $\rho^*$ is computed using a hold-out to mitigate overfitting.
     }
     \label{fig:app-toy-experiment-1d-0-999}
\end{figure}

First we consider additional one-dimensional examples, investigating the effect of modifying the base momentum $\rho_B$.
We present $\rho_B=0.99$ in \Cref{fig:app-toy-experiment-1d-0-99}, and $\rho_B=0.999$ in \Cref{fig:app-toy-experiment-1d-0-999}.
The results for $\rho_B=0.9999$ are presented in main text in \Cref{fig:toy-experiment}.

As described by the scaling error term in \Cref{eq:scaling-error}, the approximation error at a given $\kappa$ is higher for lower momenta $\rho$.
For a large range of scalings $\kappa$, the \gls{ema} Scaling Rule and the optimal momenta $\rho^*$ are consistent.
In summary, we see the synthetic experiments validate the results of \Cref{subsec:toy-experiment} for a range of momenta $\rho$.

\begin{figure}[ht]
     \centering
     \begin{subfigure}[b]{0.45\textwidth}
         \centering
         \includegraphics[width=\textwidth]{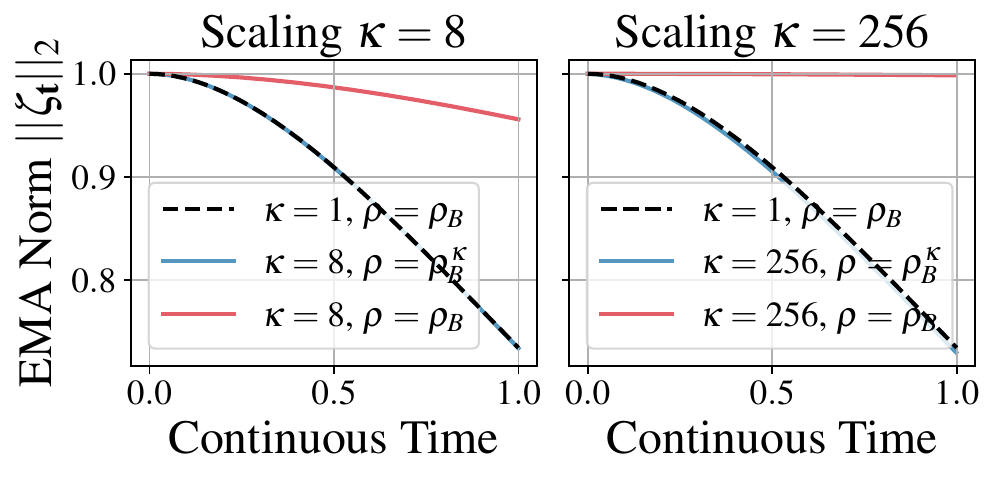}
         \caption{Norm of the model \gls{ema} $\rvzeta$ under different scalings $\kappa$, with $\rho_B=0.9999$, $\eta_B=10^{-4}$, $D=2$.}
     \end{subfigure}
     \hfill
     \begin{subfigure}[b]{0.52\textwidth}
         \centering
         \includegraphics[width=\textwidth]{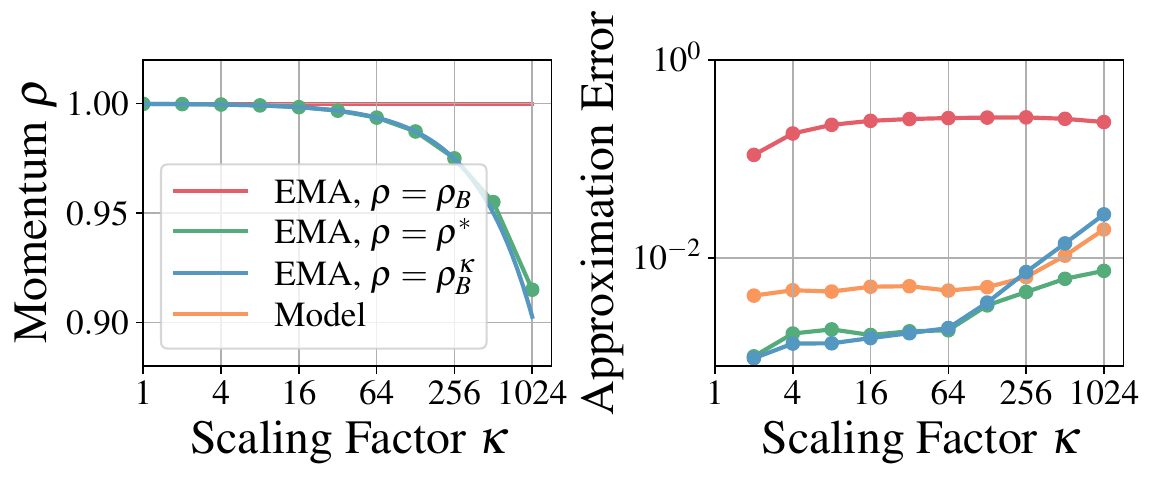}
         \caption{Choices for momentum (left) with corresponding approximation errors (\Cref{eq:optimal-momentum}) (right).}
     \end{subfigure}
     \caption{(a) We show the effect of scaling by comparing model \gls{ema} trajectories of the baseline 
     ($\kappa=1$, black dashed) to $\kappa=8$ (left) and $\kappa=256$ (right), 
     with ($\rho=\rho_B^\kappa$, blue) and without 
     ($\rho=\rho_B$, red) the EMA Scaling Rule.
     (b, left) The momentum according for different scaling rules and the empirically optimal $\rho^*$ (\Cref{eq:optimal-momentum}).
     (b, right) The approximation error (\Cref{eq:optimal-momentum}) of trajectories in (b, left) and the target model (orange). 
     Error for $\rho^*$ is computed using a hold-out to mitigate overfitting.
     }
     \label{fig:app-toy-experiment-2d-0-9999}
\end{figure}

\begin{figure}[ht]
     \centering
     \begin{subfigure}[b]{0.45\textwidth}
         \centering
         \includegraphics[width=\textwidth]{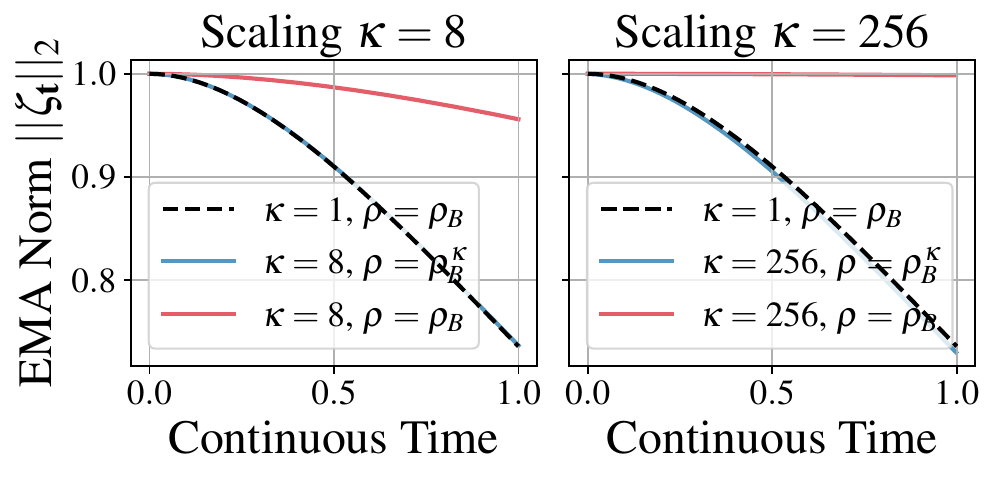}
         \caption{Norm of the model \gls{ema} $\rvzeta$ under different scalings $\kappa$, with $\rho_B=0.9999$, $\eta_B=10^{-4}$, $D=16$.}
     \end{subfigure}
     \hfill
     \begin{subfigure}[b]{0.52\textwidth}
         \centering
         \includegraphics[width=\textwidth]{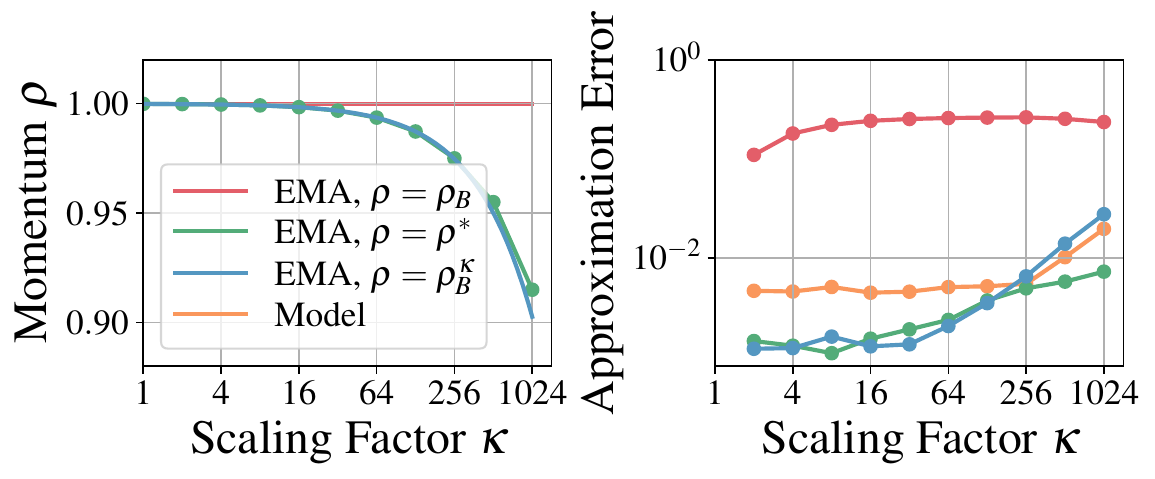}
         \caption{Choices for momentum (left) with corresponding approximation errors (\Cref{eq:optimal-momentum}) (right).}
     \end{subfigure}
     \caption{(a) We show the effect of scaling by comparing model \gls{ema} trajectories of the baseline 
     ($\kappa=1$, black dashed) to $\kappa=8$ (left) and $\kappa=256$ (right), 
     with ($\rho=\rho_B^\kappa$, blue) and without 
     ($\rho=\rho_B$, red) the EMA Scaling Rule.
     (b, left) The momentum according for different scaling rules and the empirically optimal $\rho^*$ (\Cref{eq:optimal-momentum}).
     (b, right) The approximation error (\Cref{eq:optimal-momentum}) of trajectories in (b, left) and the target model (orange). 
     Error for $\rho^*$ is computed using a hold-out to mitigate overfitting.
     }
     \label{fig:app-toy-experiment-16d-0-9999}
\end{figure}

\begin{figure}[ht]
     \centering
     \begin{subfigure}[b]{0.45\textwidth}
         \centering
         \includegraphics[width=\textwidth]{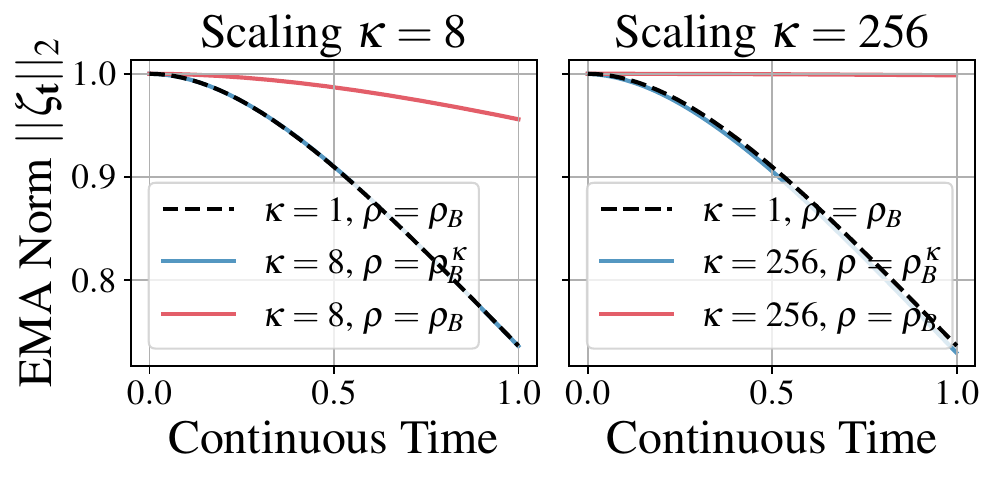}
         \caption{Norm of the model \gls{ema} $\rvzeta$ under different scalings $\kappa$, with $\rho_B=0.9999$, $\eta_B=10^{-4}$, $D=100$.}
     \end{subfigure}
     \hfill
     \begin{subfigure}[b]{0.52\textwidth}
         \centering
         \includegraphics[width=\textwidth]{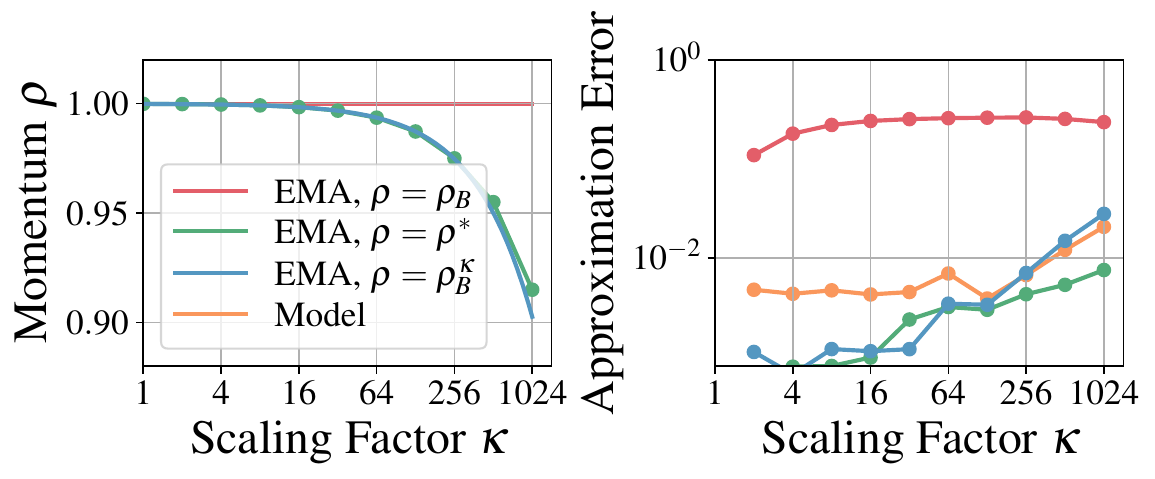}
         \caption{Choices for momentum (left) with corresponding approximation errors (\Cref{eq:optimal-momentum}) (right).}
     \end{subfigure}
     \caption{(a) We show the effect of scaling by comparing model \gls{ema} trajectories of the baseline 
     ($\kappa=1$, black dashed) to $\kappa=8$ (left) and $\kappa=256$ (right), 
     with ($\rho=\rho_B^\kappa$, blue) and without 
     ($\rho=\rho_B$, red) the EMA Scaling Rule.
     (b, left) The momentum according for different scaling rules and the empirically optimal $\rho^*$ (\Cref{eq:optimal-momentum}).
     (b, right) The approximation error (\Cref{eq:optimal-momentum}) of trajectories in (b, left) and the target model (orange). 
     Error for $\rho^*$ is computed using a hold-out to mitigate overfitting.
     }
     \label{fig:app-toy-experiment-100d-0-9999}
\end{figure}

\paragraph{Examples in higher dimensions}
Our final use of the synthetic \emph{noisy} parabola will consider an extension to $D$ dimensions.
Consider the optimization of $\rvtheta\in\R^D$ in a \emph{noisy parabola} at the origin:
\begin{align}
    \Ls(\rvtheta)
    &=\frac a2\,\rvtheta^\intercal\rvtheta,
    &\rvtheta_{k+1} &= \rvtheta_{k} - \eta \,\rvg_k, & \rvg_k&=a\,\rvtheta_k + \rvepsilon_k,
    & \rvepsilon_k\sim \mathcal{N}\left(\mathbf 0, \tfrac{b \,\rvg_k^2 + c}\kappa \right),
\end{align}
for curvature $a>0$,
scaled additive $b>0$,
and additive $c>0$ noise coefficients.
The scaling factor $\kappa$ in the covariance denominator implements the reduction in gradient noise as the scaling (i.e., the batch size) increases \citep{DBLP:journals/corr/abs-1711-04623}.
Let $\rvtheta\in\R^D$ be optimized with \gls{sgd} (\Cref{def:sgd}) and let there be a Polyak-Ruppert average (\Cref{def:polyak-ruppert-average}) $\rvzeta\in\R^D$ with momentum $\rho=1-\beta$ for $\rvtheta$.
We consider dimensionalities $D=2$ (\Cref{fig:app-toy-experiment-2d-0-9999}), $D=16$ (\Cref{fig:app-toy-experiment-16d-0-9999}), and $D=100$ (\Cref{fig:app-toy-experiment-100d-0-9999}).
We observe no significant differences in the \gls{ema} scaling behavior as we vary dimensions.

\paragraph{Compute} The compute usage for the noisy parabola experiments is relatively small, with each run taking less than one minute on a single CPU, and so we do not detail this compute usage as we do in the other experimental sections. 

\clearpage

\subsection{Image Classification}
\label{app:subsec:polyak-image-classification}

\paragraph{Hyperparameters} We present the base hyperparameters for our image experiments in \Cref{tab:sup-r50-recipe}.

\paragraph{Data} For large scale vision evaluation, we use the ImageNet1k dataset \citep{DBLP:journals/corr/RussakovskyDSKSMHKKBBF14}, 
a widely used dataset containing approximately 1.2 million labeled images, distributed almost uniformly across 1000 different object classes, like animals, plants, and vehicles.

The images in ImageNet1k are are not consistent in resolution.
To handle this, they are resized and cropped to a standard size (in our case, $224\times 224$), before further processing.
This is part of the standard ImageNet augmentation stack for convolutional networks mentioned in 
\Cref{tab:sup-r50-recipe}.

\begin{table}[t]
  \caption{Supervised ResNetv2-50 hyperparameters used in Polyak-Ruppert Averaging experiments.}
  \label{tab:sup-r50-recipe}
  \centering
  \small
  \begin{tabular}{lc}
    \toprule
    & Supervised ResNetv2-50 \\
    \midrule
    ImageNet1k Test Top-1 & $76.27 \pm 0.10 \%$ \\
    ImageNet1k EMA Test Top-1 & $76.55 \pm 0.07 \%$ \\
    \midrule
    Weight initialization & \texttt{kaiming\_normal(relu)}  \\
    Backbone normalization    & BatchNorm  \\
    Synchronized BatchNorm over replicas & No \\ 
    Learning rate schedule & Multi step: $\times 0.1$ at $[30, 60, 80]$ epochs \\    
    Learning rate warmup (epochs) & 5 \\    
    Learning rate minimum value & $1\times 10^{-6}$  \\    
    Training duration (epochs) & 90 \\
    Optimizer & SGD + Momentum \\    
    SGD momentum & 0.9 \\    
    Optimizer scaling rule & Linear \\
    Base learning rate & 0.4  \\
    Base batch size & 1024  \\
    Base Polyak momentum & 0.9999 \\    
    Weight decay & $1\times 10^{-4}$ \\
    Weight decay scaling rule & None \\
    Weight decay skip bias & Yes \\
    Numerical precision & \texttt{bf16} \\
    Augmentation stack & ImageNet \\   
    Label smoothing rate & 0.1 \\
    \bottomrule
  \end{tabular}
\end{table}

\ifthenelse{\equal{\anonymous}{0}}{
\FloatBarrier

\paragraph{Compute usage}
The compute usage image classification Polyak-Ruppert averaging is summarized in   \Cref{tab:polyak-image-compute}.

\begin{table}[ht]
  \caption{
  Compute usage for image classification Polyak-Ruppert averaging in \Cref{fig:r50-polyak,fig:r50-polyak-full-bn}. 
  The three runs for the batch size 1,024 baseline correspond to three seeds, and the nine runs for all other batch sizes correspond to using and not using the \gls{ema} Scaling Rule shown in \Cref{fig:r50-polyak}, and its application to Batch Normalization shown in \Cref{fig:r50-polyak-full-bn}. All experiments conducted are using 80Gb A100s.}
  \label{tab:polyak-image-compute}
  \centering
  \small
\begin{tabular}{cccccc}
\toprule
 Batch Size &  GPUs &  Time (h) &  Compute/Run (GPUh) &  Runs &  Compute (GPUh) \\
\midrule
        512 &     8 &      35.3 &               282.4 &     9 &          2,541.6 \\
       1,024 &     8 &      17.1 &               137.0 &     3 &           410.9 \\
       2,048 &     8 &      13.3 &               106.7 &     9 &           960.6 \\
       4,096 &     8 &       4.2 &                33.5 &     9 &           301.9 \\
       8,192 &    16 &       2.8 &                44.8 &     9 &           403.6  \\ \midrule 
\multicolumn{5}{l}{All other compute, e.g. code development, runs with errors, and debugging} & 25,768.3 \\ \midrule        
      \textbf{Total} &&&&& \textbf{30386.8} \\
\bottomrule
\end{tabular}
\end{table}

}{\compute}

\paragraph{Additional results}

\begin{figure}[ht]
    \centering
    \includegraphics[width=0.99\textwidth]{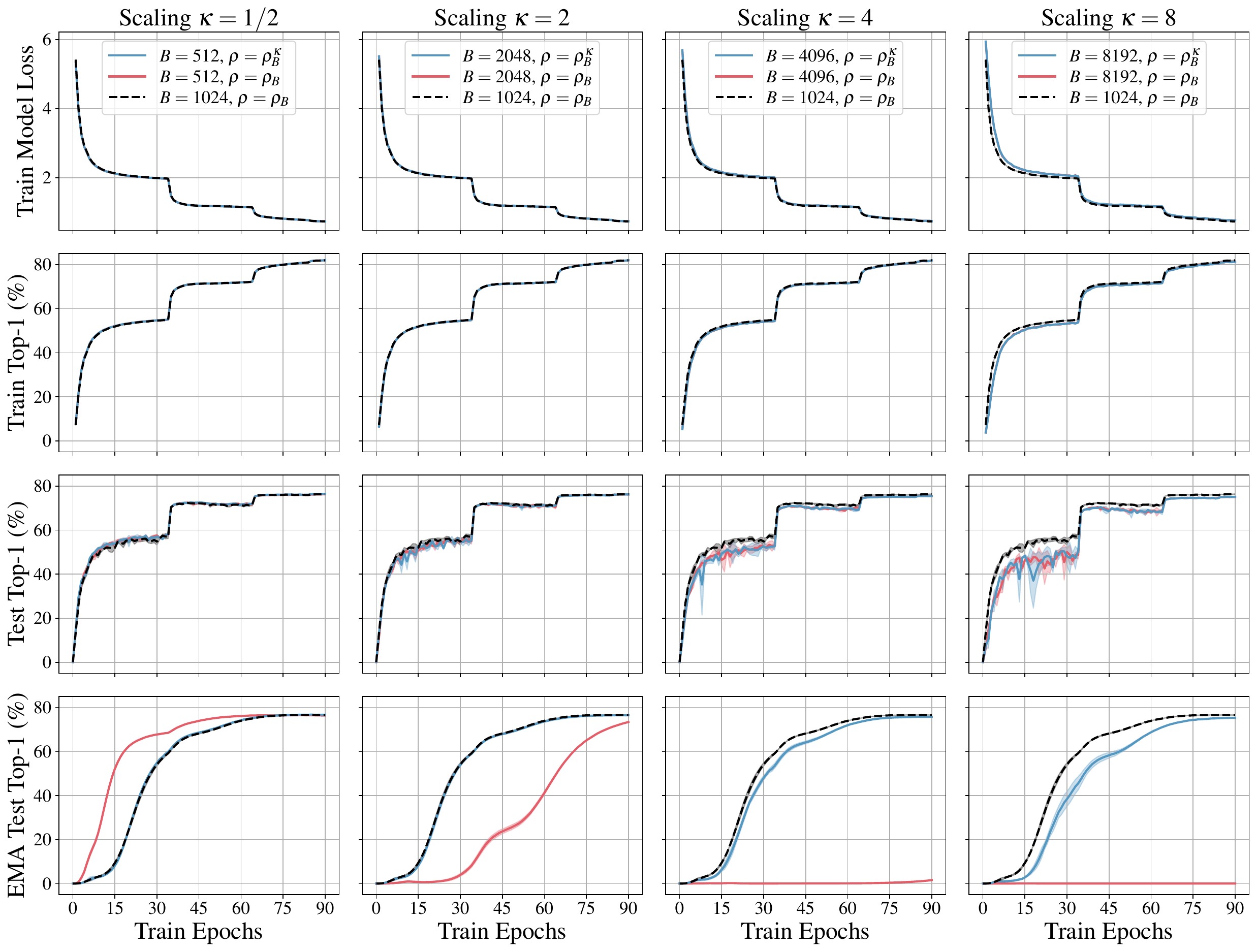}
    \caption{
    \emph{ResNetv2-50 Polyak-Ruppert averaging on ImageNet1k} for different scalings $\kappa$.
    The baseline model ($\kappa=1$, black dashed) uses batch size 1024 and momentum $\rho_B=0.9999$,
    is scaled down to a batch size of 512 (left), and up to a batch size of 4096 (right) with (blue, $\rho=\rho_B^\kappa$) and without (red, $\rho=\rho_B$) the EMA Scaling Rule (\Cref{def:ema-sr}). Bands indicate the mean and standard deviation across three runs.}
    \label{fig:r50-polyak-full}
\end{figure}

In \Cref{fig:r50-polyak-full} we present a more detailed view of the results in \Cref{subsec:supervised-polyakking}.
First, we see that for all train metrics, model trajectories match,
and that a learning rate step schedule after warmup is present.
As discussed in \Cref{fig:r50-polyak-full}, a gap in \gls{ema} Test Top-1 trajectories begins at scaling $\kappa=4$, with a more pronounced effect visible at $\kappa=8$.
From \Cref{fig:r50-polyak-full} it is clear that the (non-\gls{ema}) Test Top-1 performance trajectory is no longer matching at these scalings, demonstrating that the problem is not due to a breakdown of the \gls{ema} Scaling Rule, but rather, that the model is overfitting at larger batch sizes due to batch normalization \citep{DBLP:conf/icml/IoffeS15}.

\FloatBarrier

\subsection{Applying the EMA Scaling Rule to Batch Normalization}
\label{subsec:polyak-bn}

In 
\Cref{subsec:supervised-polyakking} and 
\Cref{app:subsec:polyak-image-classification},
we investigated a range of scalings $\kappa$, \emph{with} and \emph{without} applying the \gls{ema} Scaling Rule to the Polyak momentum.
In those experiments, we maintained Batch Normalization \citep{DBLP:conf/icml/IoffeS15} coefficients of $\rho_{\text{BN}}=0.9$ throughout\footnote{In many \gls{ml} frameworks, this value is defined using $\beta_\rho=1-\rho$, i.e. the default is $0.1$ and corresponds to $\beta_{\text{BN}}$ rather than $0.9$ corresponding to $\rho_{\text{BN}}$. We use $\rho_{\text{BN}}$ to maintain consistency across this work.}, i.e. the \gls{ema} Scaling Rule is not applied.
The running statistics of Batch Normalization \emph{are} an \gls{ema} with values determined by $\rho_{\text{BN}}$ and so it is reasonable to suspect we should apply the \gls{ema} Scaling Rule to $\rho_{\text{BN}}$ also.

\begin{figure}[th]
    \centering
    \includegraphics[width=0.99\textwidth]{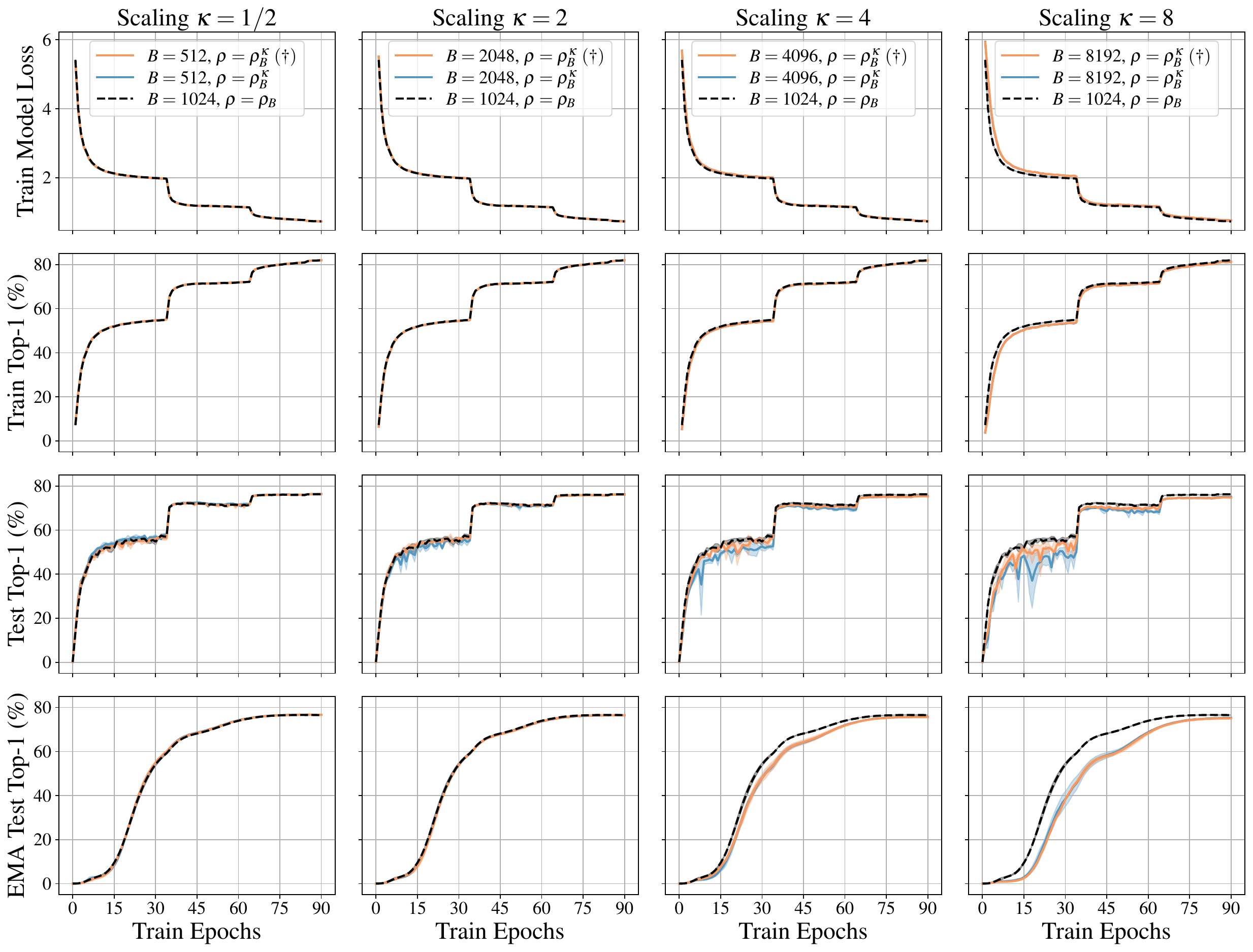}
    \caption{
    \emph{ResNetv2-50 Polyak-Ruppert averaging on ImageNet1k} for different scalings $\kappa$.
    The baseline model ($\kappa=1$, black dashed) uses batch size 1024 and momentum $\rho_B=0.9999$,
    is scaled down to a batch size of 512 (left), and up to a batch size of 4096 (right) with the EMA Scaling Rule applied to \emph{only} model parameters (blue, $\rho=\rho_B^\kappa$),
    and model parameters \emph{and} buffers
    (orange, $\rho=\rho_B^\kappa$ $(\dagger)$). Bands indicate the mean and standard deviation across three runs.}
    \label{fig:r50-polyak-full-bn}
\end{figure}

In \Cref{fig:r50-polyak-full-bn} we investigate the effect of applying the \gls{ema} Scaling Rule to Batch Normalization coefficients, using $\hat\rho_{\text{BN}}=\rho_{\text{BN}}^\kappa$.
We observe that the Test Top-1 trajectories \emph{with} the \gls{ema} Scaling Rule applied are slightly closer to the reference trajectories for scalings $\kappa\geq 2$ than those trajectories \emph{without} the \gls{ema} Scaling Rule.
As the effect is not particularly large, at least in this setup, we do pursue further ablating applications of the \gls{ema} Scaling Rule to batch normalization coefficients, and always use $\rho_{\text{BN}}=0.1$ for Batch Normalization, independent of $\kappa$.

\FloatBarrier

\section{Additional details and results for Automatic Speech Recognition (ASR)}
\label{app:speech}
In this section we provide additional details for the speech recognition experiments in both the supervised and semi-supervised case.

\paragraph{Data} We use the \librispeech{} dataset~\citep{panayotov2015librispeech} which is a dataset of audio-transcription pairs. 
For supervised Polyak-Ruppert averaging experiments, we use \tco{} as training data, and for semi-supervised pseudo-labeling experiments, we use \tco{} as the labeled and \tct{} and \tof{} as the unlabeled data.
The standard \librispeech{} validation sets (\devclean{} and \devother{}) are used to tune all hyperparameters, as well as to select the best models. 
Test sets (\testclean{} and \testother{}) are only used for reporting final model performance, measured in \gls{wer} without an external language model.
We maintain the original 16kHz sampling rate, and compute log-mel filterbanks with 80 coefficients for a 25ms sliding window, strided by 10ms, later normalized to zero mean and unit variance for each input sequence.

\paragraph{Acoustic model} 
We employ a vanilla encoder-based only transformer model trained with the \gls{ctc} loss~\citep{graves2006connectionist}. 
We use the training configuration from~\citet{likhomanenko2020slimipl}, which has three stages: 
i) 1D convolutions to perform striding (kernel of 7 with stride of 3); 
ii) a Transformer encoder with 36 layers, 
post-LayerNorm, 
four attention heads, 
an embedding dimension of 768, 
an MLP dimension of 3072, 
a dropout frequency of 0.3,
and a layer drop frequency of 0.3; and
iii) a linear layer to map to the target vocabulary\footnote{The token set of this vocabulary consists of the 26 English alphabet letters augmented with the apostrophe and a word boundary token.}. 
To reduce model training time by a factor of approximately $2-3\times$, and to reduce memory footprint, we use CAPE positional embeddings~\citep{likhomanenko2021cape} instead of relative positional embeddings~\citep{shaw2018self}: both models perform similarly.

\begin{table}[t!]
  \caption{Hyperparameters summary for speech recognition task for supervised (left) and semi-supervised pseudo-labeling (right) training with a vanilla transformer.
  The $0.3\rightarrow 0.1$ in the dropout and layer drop rates indicates that a rate of 0.3 is used during pre-training, and a rate of 0.1 is used during pseudo-labeling.}
  \label{tab:speech-transformer-hparams}
  \centering
  \small
  \begin{tabular}{lcc}
    \toprule
    &  Supervised & Pseudo-Labeling \\
    \midrule
    Librispeech test-clean / test-other WER & 7.8/19.1 & 4.8/11.5 \\
    \midrule
    Optimizer & Adam & Adam \\    
    Optimizer scaling rule & Adam & Adam \\
    Base ($\beta_1, \beta_2$) & (0.995, 0.999) & (0.995, 0.999) \\
    Base learning rate & $0.0001$ & $0.0001$  \\
    Base learning rate warmup (steps) & 64k & 64k \\  
    Learning rate schedule & Fixed (no decay) & Fixed (no decay) \\    
    Learning rate minimum value & 0 & 0  \\    
    Base training duration (steps) & 400k & 500k \\
    Base batch size (dynamic) & $8\times 290s$ & $8\times 290s$  \\
    Base teacher momentum & 0.99995 & 0.9999 \\    
    Weight decay & None & None \\
    Numerical precision & \texttt{bf16} & \texttt{bf16} \\
    Augmentation stack & \texttt{SpecAug} & \texttt{SpecAug} \\   
    Dropout & $0.3$ & $0.3\rightarrow 0.1$ \\
    Layer drop & $0.3$ & $0.3\rightarrow 0.1$ \\
    Gradient clipping & $1$ & $1$ \\
    Labeled:unlabeled data ratio  & N/A & 1:3 \\
    Base pre-training steps & N/A & 20k \\
    Base start of EMA accumulation (steps) & N/A & 19k \\
    \bottomrule
  \end{tabular}
\end{table}

\paragraph{Training} 
Here we discuss our training procedure for base batch size $B=8\times 290s$, which is adapted from~\citet{likhomanenko2020slimipl}, and is summarized in \Cref{tab:speech-transformer-hparams}.
We use SpecAugment~\citep{park2019specaug} activated after 5k steps of training: two frequency masks with frequency mask parameter $F=30$, ten time masks with maximum time-mask ratio $p=0.1$ and time mask parameter $T=50$ are used; time warping is not used. 

One difference in setup is we use the Adam optimizer, whereas \citet{likhomanenko2020slimipl} used Adagrad~\citep{DBLP:conf/colt/DuchiHS10}.
Even though Adagrad can be viewed as a particular limit ($\beta_1=0$ and $\beta_2\to 1$) of Adam \citep{DBLP:journals/corr/KingmaB14}, we were unable to produce reasonable optimization in practice when applying the Adam Scaling Rule of \citet{DBLP:conf/nips/MalladiLPA22} in this limit.
As a consequence, we chose to work with the Adam optimizer, where its scaling rule has been shown to work \citep{DBLP:conf/nips/MalladiLPA22}, and we take $\beta_1=0.995$, $\beta_2=0.999$, and $\eps=10^{-8}$.
We obtained similar results for $\beta_1=0.99$.
Finally, we use a linear learning rate warmup (64k steps) after which the learning rate is kept constant until convergence.
This performance can be improved further by using a step decay schedule as shown in prior work. 
We also apply gradient clipping of 1, and do not use weight decay. 

\paragraph{Pseudo-Labeling}
The pseudo-labeling process comprises of two stages:
i) The pre-training phase, where we train model on labeled data for 20k steps with model EMA accumulation starting after 19k steps; and ii) the pseudo-labeling phase, where we involve unlabeled data by generating pseudo-labels from the model EMA (teacher) and provide them to the model (student) as if they were ground-truth labels. 
Pseudo-labels are generated without any dropout applied to the teacher, and no data augmentation is applied for the corresponding inputs. 
To produce the pseudo-label, we use \emph{hard transcription} (\Cref{def:hard-transcription})
\begin{definition}[Hard Transcription] 
For a sequence of frames, select the most probable token per frame, removing repetitions \emph{and} the CTC blank token. For example, ``h\#\#eelll\#\#ll\#\#\#oo'' is transformed into ``hello'', where ``\#'' is the CTC blank token.
\label{def:hard-transcription}
\end{definition}
These hard transcriptions are then used as transcription for student optimization.
We use a 1:3 proportion of labeled to unlabeled data as this was found to be optimal in~\citet{likhomanenko2020slimipl}, and we decrease model dropout and layer drop rates to 0.1 after pre-training phase.
As we have access to the ground-truth labels on the data being treated as unlabeled, we can track pseudo-label quality by computing pseudo-labels on this data, and compute the \gls{wer} against their ground-truth. 
Pseudo-label quality is the primary metric to evaluate progress on unlabeled data, as loss on pseudo-labeled data is unreliable when a teacher model and pseudo-labels are evolving with each time step.

{\bf Scaling of batch size}~~~
Sequential data is typically processed using dynamic batching as it is more computationally efficient than using a fixed number of sequences~\citep{ott2019fairseq}. 
In our work, we use dynamic batching of $\sim$290s audio per GPU.
Moreover, for \gls{ctc} we do not apply any additional sequence normalization.
We experimented with fixed batching, but did not observe any significant differences in conclusions compared with the dynamic batching. 

We note that dynamic batching is a more challenging setting for achieving systematic scaling, as the number of independent sequences in any given batch may change, and the \gls{iid} assumption does not hold at the frame level.
Despite these violations of the assumptions of \Cref{subsec:ema-sdes},
our results demonstrate that the Adam Scaling Rule (\Cref{def:adam-sr}, \cite{DBLP:conf/nips/MalladiLPA22}) holds in the case of dynamic batches, as does our \gls{ema} Scaling Rule (\Cref{def:ema-sr}).

The base batch size is set to $B=8\times 290s$, and in our experiments we scale down to batch size of $B=2\times 290s$ and up to batch size of $B=128\times 290s$.
The number of warmup and pre-training steps, steps before SpecAugment is turn on and model EMA is accumulated are scaled according to \Cref{sec:app-sde-perspective}.

\ifthenelse{\equal{\anonymous}{0}}{{\bf Compute}~~~All experiments are done using A100 80GB 8GPU nodes with \texttt{bfloat16} precision training. While for supervised training evaluation of different EMA decay values is done in parallel during a single run, for pseudo-labeling every EMA decay value needs separate training. Final models training compute is detailed in~\Cref{tab:asr-sup-compute,tab:asr-pl-compute}. Total compute, including e.g. code development, runs with errors, and debugging, is {\bf 61k} GPUh.

\begin{table}[ht]
  \caption{
  Compute usage for supervised model for speech recognition task in \Cref{fig:app-speech-polyak}. 
  Values \emph{include} node allocation times (typically a small \% of corresponding total runtime), giving a practical   estimate of reproduction cost. 
  All experiments conducted are using 80Gb A100s with fast interconnect.}
  \label{tab:asr-sup-compute}
  \centering
  \small
  \begin{tabular}{ccccccc}
\toprule
 Batch Size &  GPUs &  Time (h) &  Compute/Run (GPUh) &  Runs &  Compute (GPUh) \\
\midrule
       2 $\times$ 290s &      2 &     222      &       444         &       1 &    222       \\
       4 $\times$ 290s &      4 &       108    &       432         &       1 &    432       \\
       8 $\times$ 290s &      8 &     64      &         512       &       1 &      512     \\
       16 $\times$ 290s &      16 &      54     &       864         &       1 &       896    \\
       32 $\times$ 290s &      32 &      37     &       1,184         &       1 &     1,184      \\
      \midrule 
      \textbf{Total} &&&&& \textbf{3,436} \\
\bottomrule
  \end{tabular}
\vspace{-0.5cm}
\end{table}

\begin{table}[ht]
  \caption{
  Compute usage for continuous pseudo-labeling for the speech recognition task in \Cref{fig:app-speech-pl-9999}. 
  Values \emph{include} node allocation times (typically a small \% of corresponding total runtime), giving a practical estimate of reproduction cost. 
  All experiments conducted are using 80Gb A100s with fast interconnect.}
  \label{tab:asr-pl-compute}
  \centering
  \small
  \begin{tabular}{ccccccc}
\toprule
 Batch Size &  GPUs &  Time (h) &  Compute/Run (GPUh) &  Runs &  Compute (GPUh) \\
\midrule
       2 $\times$ 290s &      2 &     225      &       550         &       2 &    1,110       \\
       4 $\times$ 290s &      4 &       120    &       480         &       2 &    960       \\
       8 $\times$ 290s &      8 &     72      &         576       &       1 &      576     \\
       16 $\times$ 290s &      16 &      45     &       720         &       2 &       1,440    \\
       32 $\times$ 290s &      32 &      33     &       1,056         &       4 &     4,224      \\
       64 $\times$ 290s &      64 &      25     &       1,600         &       2 &     3,200      \\
      \midrule 
      \textbf{Total} &&&&& \textbf{11,510} \\
\bottomrule
  \end{tabular}
\end{table}
\vspace{-0.2cm}
}{\compute}

\ifthenelse{\equal{\anonymous}{0}}{}{\FloatBarrier}
\FloatBarrier
\subsection{Additional experimental settings and detailed metrics}
\label{subsec:speech-detailed}

\begin{figure}[ht]
    \centering
    \includegraphics[width=0.99\textwidth]{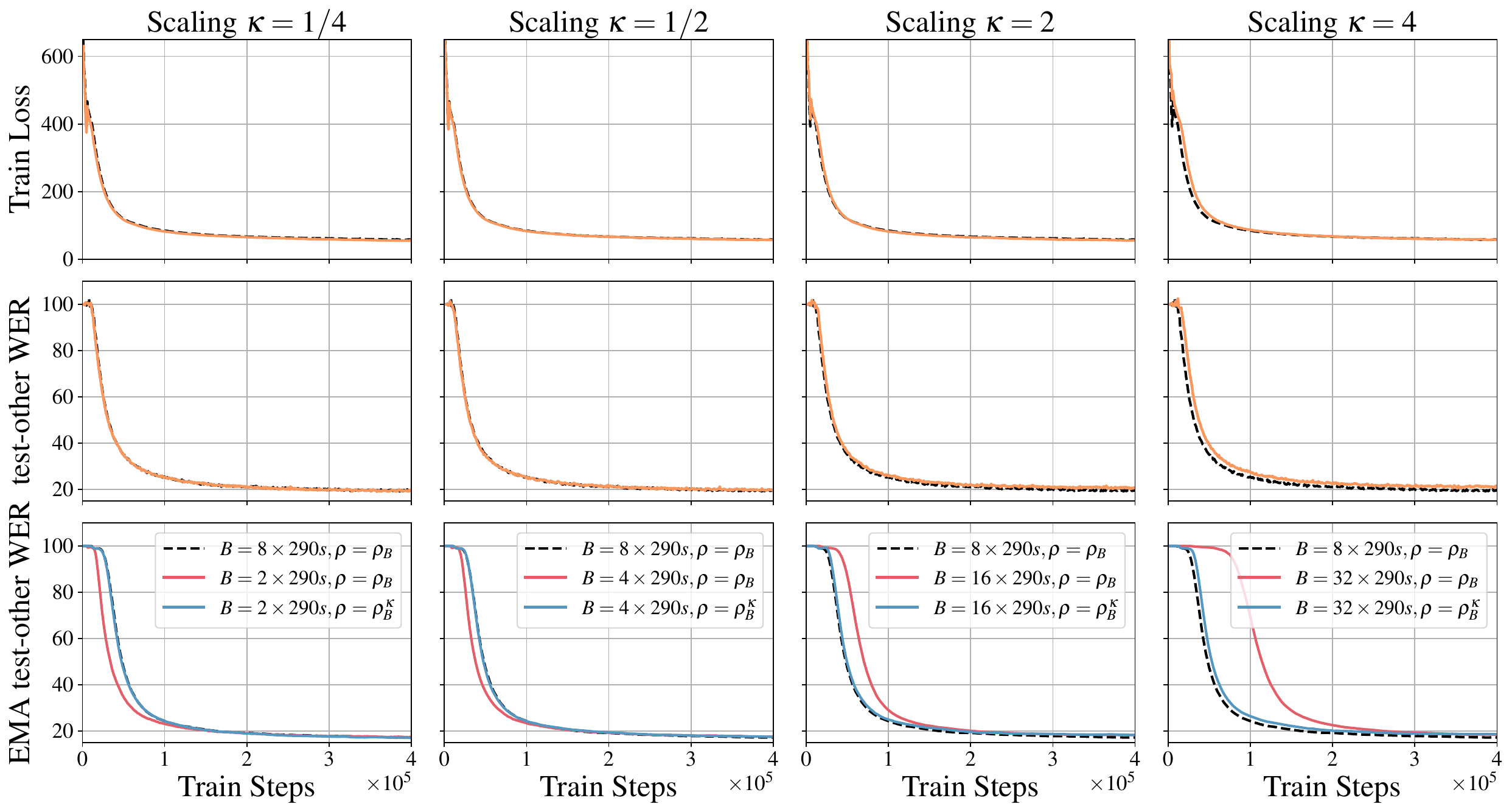}
    \caption{
    \emph{Transformer Polyak-Ruppert averaging on LibriSpeech (trained on \tco{})} with different scalings $\kappa$.
    The baseline ($\kappa=1$, black dashed)
    is trained with Adam and momentum $\rho_B=0.99995$ at a \emph{dynamic batch size} $B=8\times 290s$, which corresponds to a single train step on the $x$-axis. 
    We investigate dynamic batch sizes down to 
    $B=2\times 290s$ (left) and up to 
    $B=32\times 290s$ (right), 
    with (blue, $\rho=\rho_B^\kappa$), and without (red, $\rho=\rho_B$) the \gls{ema} Scaling Rule (model non-EMA is marked by orange).
    The Adam Scaling Rule (\citet{DBLP:conf/nips/MalladiLPA22}, \Cref{def:adam-sr}) is used throughout. 
    For momentum $\rho_B=0.9999$ we observe similar trajectories for all models.
    }
    \label{fig:app-speech-polyak}
    \vspace{-0.3cm}
\end{figure}

\begin{figure}[h!]
    \centering
    \includegraphics[width=0.99\textwidth]{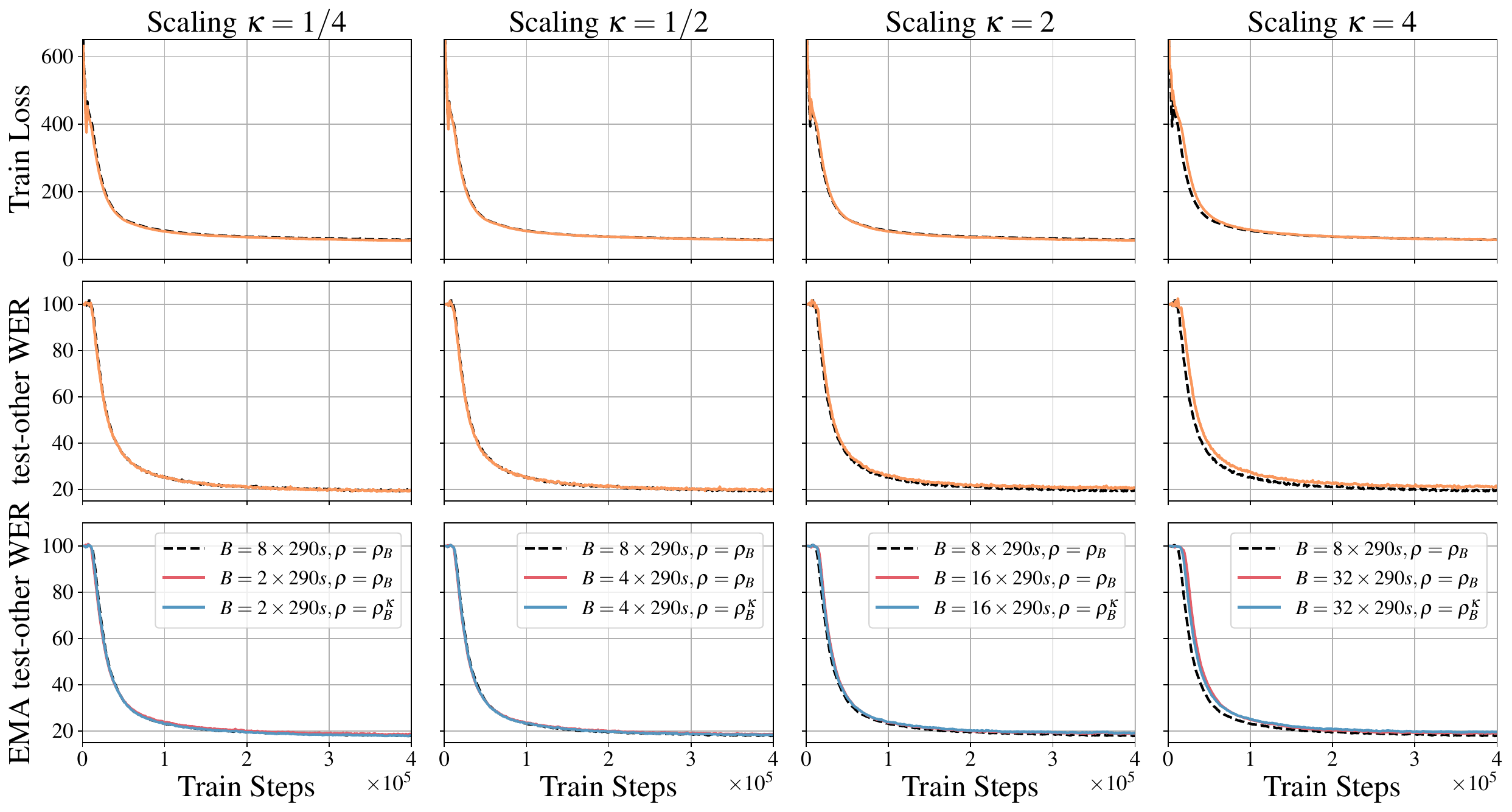}
    \caption{
    \emph{Transformer Polyak-Ruppert averaging on LibriSpeech (trained on \tco{})} with different scalings $\kappa$.
    The baseline ($\kappa=1$, black dashed)
    is trained with Adam and momentum $\rho_B=0.999$ at a \emph{dynamic batch size} $B=8\times 290s$, which corresponds to a single train step on the $x$-axis. 
    We investigate dynamic batch sizes down to 
    $B=2\times 290s$ (left) and up to 
    $B=32\times 290s$ (right), 
    with (blue, $\rho=\rho_B^\kappa$), and without (red, $\rho=\rho_B$) the \gls{ema} Scaling Rule (model non-EMA is marked by orange).
    The Adam Scaling Rule (\citet{DBLP:conf/nips/MalladiLPA22}, \Cref{def:adam-sr}) is used throughout. 
    If momentum $\rho_B$ is small and accumulation history is short we observe no any significant difference between models which all are matching the reference trajectory despite scaling $\kappa$.
    }
    \label{fig:app-speech-polyak-noinf}
    \vspace{-0.5cm}    
\end{figure}

\begin{figure}[ht]
    \centering
    \includegraphics[width=0.99\textwidth]{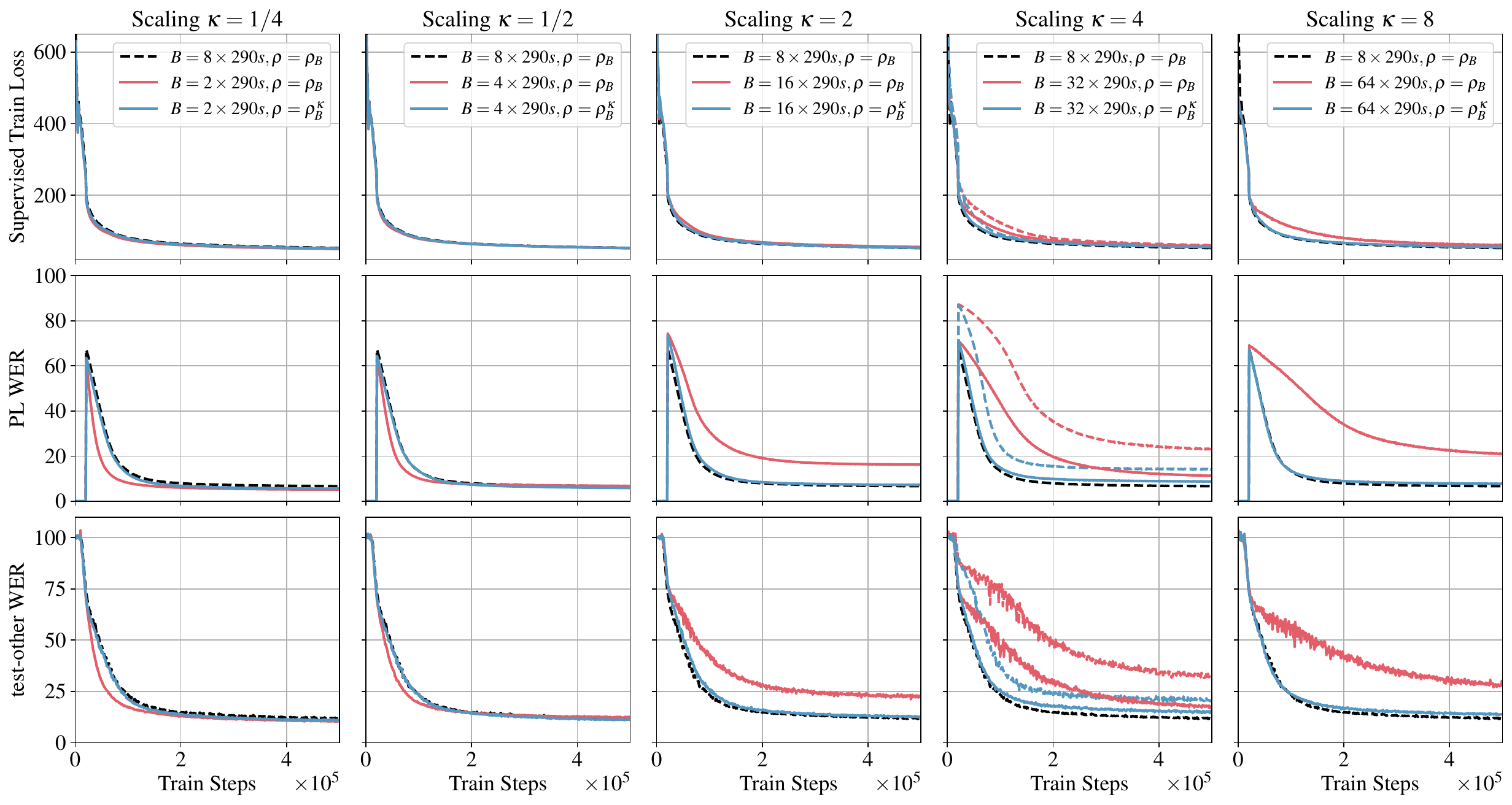}
    \caption{
    \emph{Transformer pseudo-labeling on LibriSpeech (trained on \tco{} as labeled and the rest of LibriSpeech as unlabeled)} with different scalings $\kappa$. 
    The baseline ($\kappa=1$, black dashed)
    is trained with Adam at a \emph{dynamic batch size} of $8\times 290$ seconds, which corresponds to a single train step on the $x$-axis.
    The model \gls{ema} (\emph{teacher}) is updated with momentum $\rho_B=0.9999$.
    We investigate dynamic batch sizes down to $B=2\times 290s$ (left) and up to $B=64\times 290s$ (right),
    with (blue, $\rho=\rho_B^\kappa$) and without (red, $\rho=\rho_B$) the \gls{ema} Scaling Rule.
    The Adam Scaling Rule (\citet{DBLP:conf/nips/MalladiLPA22}, \Cref{def:adam-sr}) is used throughout.
    For $\kappa\leq2$, we start pseudo-labeling after $20\text{k}/\kappa$ training steps; while for $\kappa>2$, we start when pre-training WER matches the baseline WER ($24\text{k}/\kappa$ for $\kappa=4$ and $29\text{k}/\kappa$ for $\kappa=8$). For $\kappa=4$ we experimented with both variants: we start pseudo-labeling after $20\text{k}/\kappa$ (dashed) and when pre-training WER matches the baseline WER (solid, $24\text{k}/\kappa$).
    }
    \label{fig:app-speech-pl-9999}
    \vspace{-0.2cm}
\end{figure}

\begin{figure}[h!]
    \centering
    \includegraphics[width=0.72\textwidth]{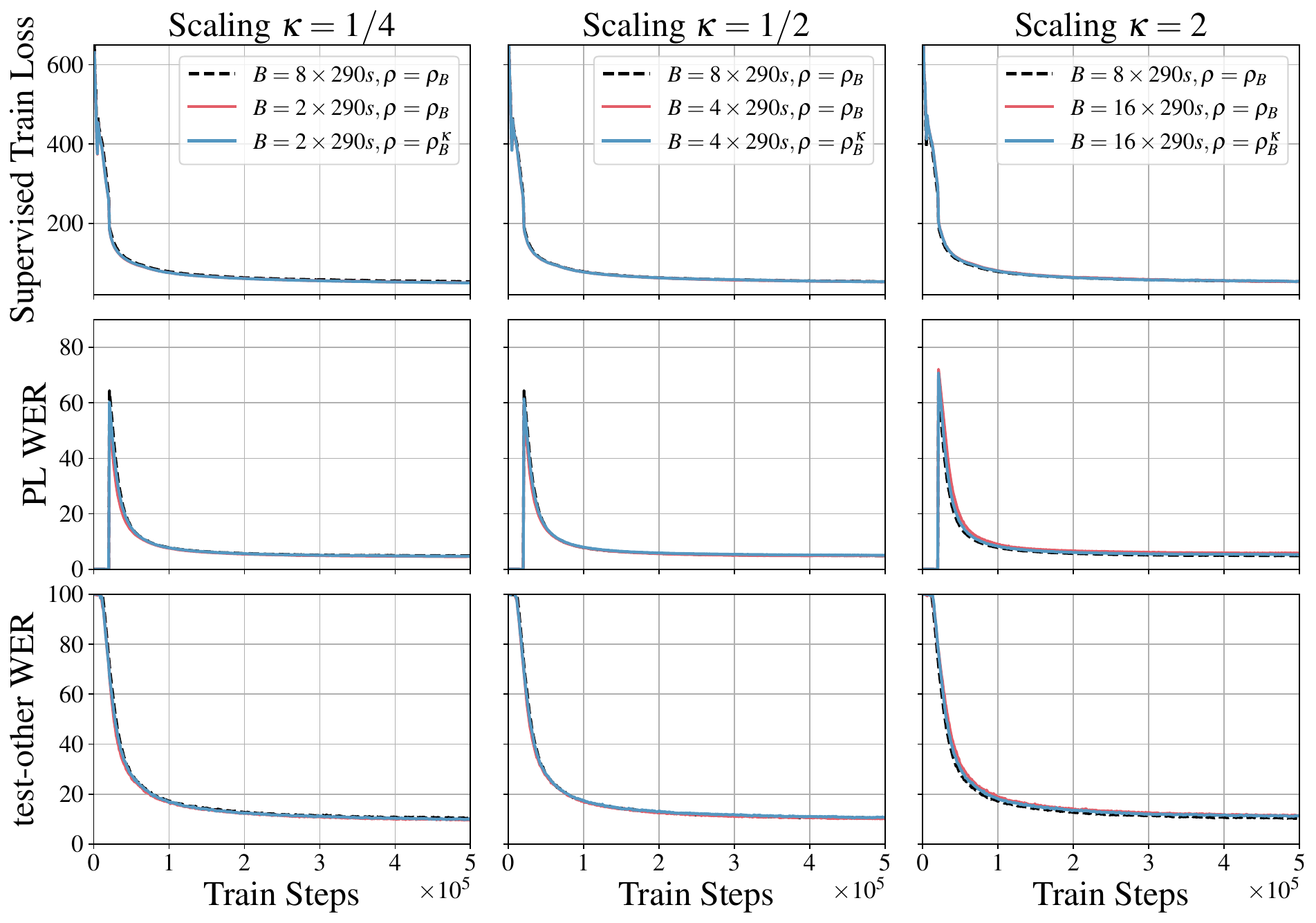}
    \caption{
    \emph{Transformer pseudo-labeling on LibriSpeech (using \tco{} as labeled)} with different scalings $\kappa$. 
    The baseline ($\kappa=1$, black dashed)
    is trained with Adam at a \emph{dynamic batch size} of $8\times 290$ seconds, which corresponds to a single train step on the $x$-axis.
    The model \gls{ema} (\emph{teacher}) is updated with momentum $\rho_B=0.999$.
    We investigate dynamic batch sizes down to $B=2\times 290s$ (left) and up to $B=16\times 290s$ (right),
    with (blue, $\rho=\rho_B^\kappa$) and without (red, $\rho=\rho_B$) the \gls{ema} Scaling Rule.
    The Adam Scaling Rule is used throughout.
    }
    \label{fig:app-speech-pl-999}
    \vspace{-0.5cm}
\end{figure}

We present detailed comparison between models trained with and without EMA Scaling Rule in~\Cref{fig:app-speech-polyak,fig:app-speech-polyak-noinf} for supervised training and in~\Cref{fig:app-speech-pl-9999,fig:app-speech-pl-999} for semi-supervised training.

First, we observe that if the Adam Scaling Rule does not hold perfectly\footnote{See \citet{DBLP:conf/nips/MalladiLPA22} for a discussion on scenarios that lead to a breakdown of the Adam Scaling Rule.} (there is a mismatch between trajectories for the model before pseudo-labels are involved) the EMA Scaling Rule also gives discrepancies with the reference trajectory, however they are negligible compared to models trained without EMA Scaling Rule.
For the semi-supervised training, to alleviate the difficulties with a breakdown of the Adam Scaling Rule for large $\kappa$ we postpone the pseudo-labeling process until the model reaches similar \gls{wer} as the baseline. 
This allows us to align the initial model conditions for pseudo-labeling.
In this scenario we are able to match the reference trajectory up to $\kappa=8$.

We note that this result reveals that errors for the Adam Scaling Rule \emph{and} the EMA Scaling Rule are contributing, 
although the way in which they contribute is different, and one can dominate the other.
We observe in \Cref{fig:app-speech-pl-9999} that if the initial conditions of the models are similar (attained by using the same \gls{wer} as a condition to begin pseudo-labeling) then the error from the EMA Scaling Rule dominates over that of the Adam Scaling Rule, causing a divergence in training dynamics.

Second, we observe in practice that the EMA Scaling Rule holds for both fixed batching (a sequence length in the batch can vary significantly) and for dynamic batching (when total number of frames in the batch is fixed, though padding still is accounted to the this amount). 
This shows that EMA Scaling Rule is applicable to sequential data too.

Third, we observe in \Cref{fig:app-speech-polyak-noinf,fig:app-speech-pl-999} that for smaller values of $\rho_B$, scaling with or without EMA Scaling Rule behave similarly, and reference trajectories match in the supervised and semi-supervised cases.
However, if the momentum is too large, the \emph{teacher} moves slowly and is uninformative, whereas if the momentum is too low, the \emph{teacher} and the \emph{student} are effectively be the same model, implying: i) the student will self-predict with high confidence, removing any benefits of distillation\footnote{\citet{He2020Revisiting} alleviated the problem with the proper amount of noise during \textit{student} model training, whilst~\citet{xu2020iterative} used beam-search decoding with a language model.}; and ii) training instability or model divergence will happen in the low-resource settings \citep{likhomanenko2020slimipl,higuchi2022momentum}. 

\FloatBarrier

\subsection{Scaling to $\kappa=16$ with Progressive Scaling}\label{app:speech-progressive}
\begin{figure}[h]
    \centering
    \includegraphics[width=0.85\textwidth]{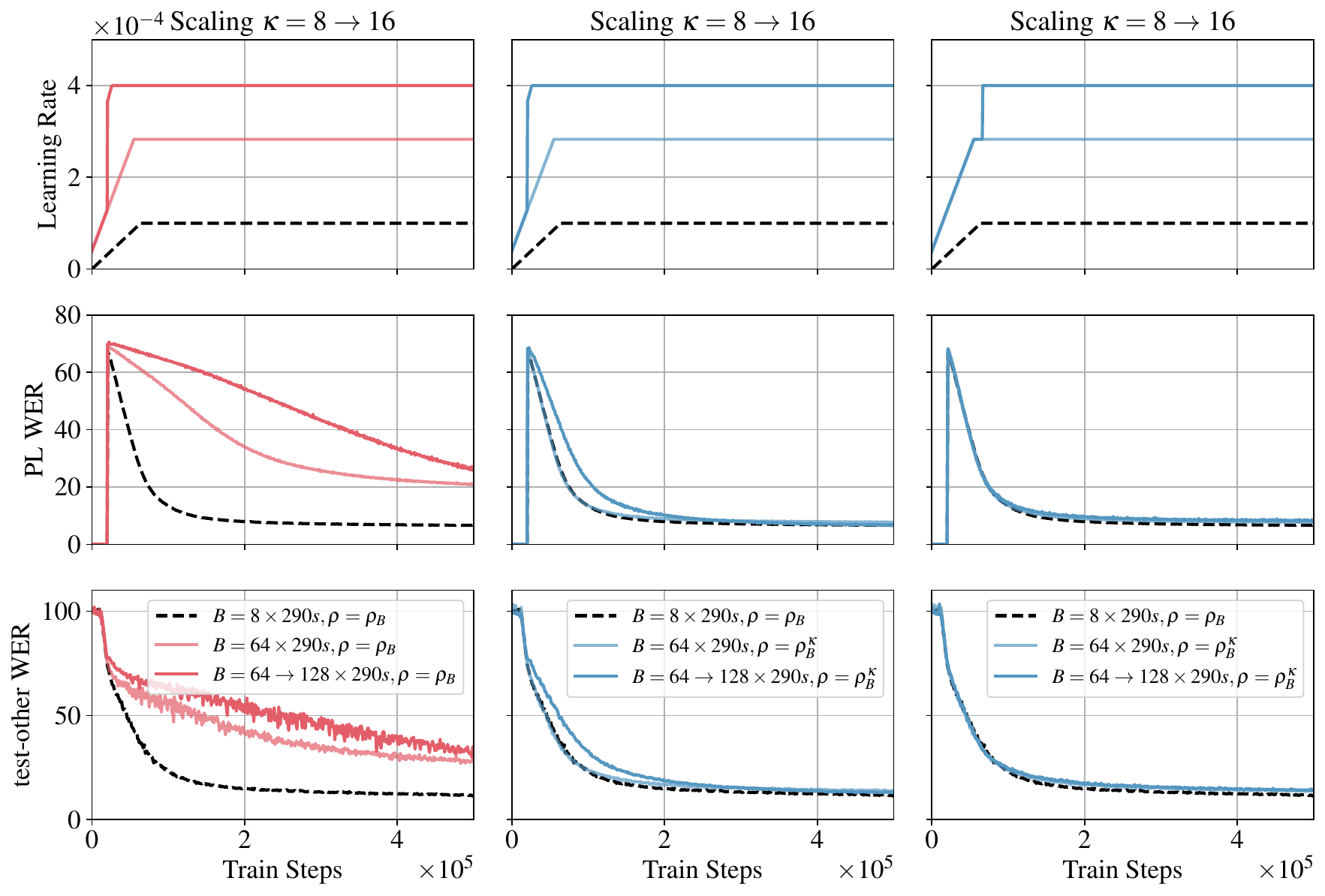}
    \caption{
    \emph{Transformer pseudo-labeling on LibriSpeech (trained on \tco{} as labeled and the rest of LibriSpeech as unlabeled)} with different Progressive Scaling from $\kappa=8$ to $\kappa=16$ ($\kappa=8\rightarrow 16$). 
    The baseline ($\kappa=1$, black dashed)
    is trained with Adam at a \emph{dynamic batch size} of $8\times 290$ seconds, which corresponds to a single train step on the $x$-axis.
    The model \gls{ema} (\emph{teacher}) is updated with momentum $\rho_B=0.9999$.
    The scaling with $\kappa=8$ is shown with lighter color for reference from~\Cref{fig:app-speech-pl-9999}.
    We investigate dynamic batch sizes progressively from $B=64\times 290s$ to $B=128\times 290s$,
    with (blue, $\rho=\rho_B^\kappa$) and without (red, $\rho=\rho_B$) the \gls{ema} Scaling Rule.
    For reference (top) we show the learning rate schedule with Progressive Scaling.
    The Adam Scaling Rule (\citet{DBLP:conf/nips/MalladiLPA22}, \Cref{def:adam-sr}) is used throughout.
    Left and middle correspon to Progressive Scaling with scale from $\kappa=8$ to $\kappa=16$ at 29k steps, while right corresponds to 75k steps.
    }
    \label{fig:app-speech-pl-9999-progressive}
    \vspace{-0.4cm}
\end{figure}

Finally, we aim to scale semi-supervised pseudo-labeling further to $\kappa=16$. 
In this case we observe that Adam Scaling Rule does not hold in the pre-training phase and there is no model convergence.
To overcome this, we apply Progressive Scaling (\Cref{def:progressive-scaling}). 
We pre-train models on supervised data with $\kappa=8$ for 29k of reference steps 
(model EMA accumulation starts at 28k steps). We then scale to $\kappa=16$ and begin pseudo-labeling.
We see in \Cref{fig:app-speech-pl-9999-progressive} that Progressive Scaling enables us to scale pseudo-labeling to $\kappa=16$ with (middle) and without (left) the EMA Scaling Rule.
Second, models \emph{with} the \gls{ema} Scaling Rule track the baseline much closer than models without the \gls{ema} Scaling Rule, although a small gap is present.
We further experimented with Progressive Scaling, postponed the transition condition to the $\kappa=16$ until 75k reference steps.
In \Cref{fig:app-speech-pl-9999-progressive} (right), we see this scaled model tracks the reference trajectory, and so using a combination of the \gls{ema} Scaling Rule and Progressive Scaling, we are able to scale pseudo-labeling to $\kappa=16$, corresponding to a dynamic batch size of $128\times 290s$.

\FloatBarrier

\section{Additional details and results for self-supervised image representation learning}
\label{app:ssl}

\paragraph{Organization} 
This appendix is structured into three sections.
We first give an overview of our chosen 
\gls{ssl} method \gls{byol} (\Cref{app:sec-components-ssl}),
our recipe for training \gls{byol} using ResNet 18s (\Cref{subsec:byol-r18}), 
our recipe for training \gls{byol} using \glspl{vit} (\Cref{app:byol-vit}), 
ablations of normalization approaches that lead to the development of this recipe (\Cref{app:byol-vit-ln-vs-bn}), 
and additional results corresponding to longer training duration (\Cref{app:byol-waterfall}) and further understanding the impact of Progressive Scaling (\Cref{app:byol-progressive-scaling-regimes}).

Second, we demonstrate that the \gls{ema} Scaling Rule combined with Progressive Scaling can scale a ResNet-50 BYOL model trained with LARS to batch size 32,768 without performance drop, demonstrating the empirical utility of the tools we provide outside of their theoretical validity (\Cref{subsec:byol-additional}).

Finally, we show that it is possible to systematically scale DINO \citep{DBLP:journals/corr/abs-2104-14294} using a combination of Progressive Scaling and the \gls{ema} Scaling Rule,
providing a solution for researchers and practitioners wanting to train DINO at scale.

\subsection{Components of self-supervised learning}
\label{app:sec-components-ssl}

First, a key component of many \gls{ssl} methods is the \emph{stop-gradient} 
or $\text{StopGrad}$ (\Cref{def:stop-grad}).
\begin{definition}[Stop Gradient/$\text{StopGrad}(\,\cdot\,)$] 
     The \emph{stop-gradient} operator $\text{StopGrad}(\,\cdot\,)$ prevents the flow of gradient information
    \begin{equation}
    \frac{df(\text{StopGrad}(h(x ; \rvomega)); \rvtheta)}{d\rvomega}\equiv 0
    \end{equation}
    \label{def:stop-grad}
    for all parametric functions $h$ and $f$ and for all parameters $\rvtheta$ and $\rvomega$.
\end{definition}
Applying a \emph{stop-gradient} is sometimes called \emph{detaching} in the literature. 
Now, we 
introduce the update rule of our representative \gls{ssl} method \gls{byol} in \Cref{def:byol}.
\begin{definition}[BYOL Update] 
     \gls{byol} learns unsupervised features by minimizing the cosine distance between the predictions of a student backbone
     $f(\,\cdot\,;\rvtheta)$ (typically a ResNet or Vision Transformer),
     projected through 
     $h(\,\cdot\,;\rvomega$) (typically a \gls{mlp}), and the predictions of an \gls{ema} teacher $f(\,\cdot\,;\rvzeta)$ \citep{DBLP:conf/nips/GrillSATRBDPGAP20}. 
     The update for the parameters of \gls{byol} is then
    \begin{align}
    (\rvtheta_{t+1},\rvomega_{t+1})
    &=
    (\rvtheta_t,\rvomega_t) - \eta \times \frac1B
    \sum_{x\in\sB} \nabla_{(\rvtheta,\rvomega)} \Ls(x;\rvtheta_{t},\rvomega_{t},\rvzeta_{t})\\
    \rvzeta_{t+1}
    &=
    \rho \,\rvzeta_t + (1-\rho)\,\rvtheta_{t+1} \label{eq:byol-ema-update}\\
    \text{with} \;\; 
    \Ls(x;\rvtheta_{t},\rvomega_{t},\rvzeta_{t})
    &=
    \frac12
    \cos \big[ h( f(x_1;\rvtheta_t);\rvomega_t), \text{StopGrad}(f(x_2;\rvzeta_t))  \big]  + (x_1\leftrightarrow x_2),
    \end{align}
    where $\cos (\va,\vb)\equiv 1-\va\cdot \vb/(||\va||\,||\vb||)$ is the cosine distance, 
    and
    $x_1$ and $x_2$ are two views of a single variate $x$, often produced by augmentations,
    and
    $x_1\leftrightarrow x_2$ denotes symmetrization over $x_1$ and $x_2$.
    \label{def:byol}
\end{definition}
As noted in \Cref{subsec:self-supervised},the  \gls{byol} \gls{ema} update (\Cref{eq:byol-ema-update}) uses $\rvtheta_{t+1}$ instead of our analyzed $\rvtheta_{t}$ (\Cref{eq:scalingRuleSummaryEquation}).
The effect upon the overall \gls{ema} update is $\mathcal{O}(\eta\times\beta_\rho)$ and so is captured by the \gls{ema} Scaling Rule (\Cref{def:ema-sr}).

One more piece of technology typically employed in \gls{ssl} is a \emph{tracking probe} (\Cref{def:linear-probe}) which we will use to evaluate the performance of \gls{byol} on downstream tasks of interest, for example, image classification.
\begin{definition}[Tracking Probe/Linear Probe] 
    When optimizing model parameters $\rvomega_t$ of an \gls{ssl} method, 
    simultaneously optimize the parameters $\rvxi$ of a probe model $r(\,\cdot\,;\rvxi)$
    under a downstream objective $\Ls^{(d)}$.
    For example, in classification, with data $x$ and samples $y$
    \begin{align}
    \Ls^{(d)}(x,y,\rvtheta_{t},\rvxi_t)&= - \log P(y|r(\text{StopGrad}(h(x;\rvomega_t));\rvxi))\\
    \Ls^{(\text{total})}(x,y;\rvtheta_{t},\rvomega_{t},\rvzeta_{t},\rvxi_t) & = \Ls(x;\rvtheta_{t},\rvomega_{t},\rvzeta_{t})+\Ls^{(d)}(x,y,\rvomega_{t},\rvxi_t),
    \end{align}
    The is a probe for the teacher, which is typically the better choice due to Polyak-Ruppert averaging effects (see \Cref{subsec:supervised-polyakking}).
    \label{def:linear-probe}
    When the $r$ is a linear model, the tracking probe is called a linear probe.
\end{definition}
It is also typical to use a Batch Normalization layer \emph{without} trainable affine terms before this linear layer as in \citet{DBLP:conf/cvpr/HeCXLDG22} to stabilize probe training.
In this case, the running statistics can be absorbed into a definition of the linear layer weights and biases, and so this is still a \emph{linear probe}, although we will call this a \emph{pre-bn linear probe} to remove ambiguity.

\subsection{A ResNet-18 recipe for BYOL}
\label{subsec:byol-r18}

\paragraph{Hyperparameters} We present the base hyperparameters for training \gls{byol} with a ResNet-18 backbone using \gls{sgd} in \Cref{tab:byol-r18-recipe}.
This recipe was developed by starting from a well-known BYOL ResNet-50 recipe \citep{DBLP:conf/nips/GrillSATRBDPGAP20}, adapting the input augmentations for CIFAR10, and performing a search over learning rate choices for an SGD optimizer.

\begin{table}[t]
  \caption{BYOL ResNet-18 hyperparameters for CIFAR10}
  \label{tab:byol-r18-recipe}
  \centering
  \small
  \begin{tabular}{lc}
    \toprule
    & ResNet-18 \\
    \midrule
    Weight initialization & \texttt{kaiming\_uniform} \citep{DBLP:conf/iccv/HeZRS15} \\
    Backbone normalization    & BatchNorm  \\
    Head normalization    & BatchNorm  \\
    Synchronized BatchNorm over replicas & Yes \\     
    Learning rate schedule & Single Cycle Cosine \\    
    Learning rate warmup (epochs) & 20 \\    
    Learning rate minimum value & $0$  \\    
    Training duration (epochs) & 100 \\
    Optimizer & SGD \\    
    Optimizer scaling rule & SGD \\
    Optimizer momentum & $0.9$ \\
    Gradient clipping & $0.1$ \\
    Base learning rate & $0.02$  \\
    Base batch size & $1024$  \\
    Base teacher momentum & $0.992$ \\    
    Weight decay & $1\times 10^{-6}$ \\
    Weight decay scaling rule & None \\
    Weight decay skip bias & Yes \\
    Numerical precision & \texttt{tf32} \\
    Augmentation stack & \texttt{BYOL CIFAR10} \\   
    \bottomrule
  \end{tabular}
\end{table}

\FloatBarrier

\subsection{A Vision Transformer recipe for BYOL}
\label{app:byol-vit}

\paragraph{Hyperparameters} We present the base hyperparameters for training \gls{byol} with a ViT-B/16 backbone in \Cref{tab:byol-recipe}.
This recipe was developed by starting from a well-known supervised ViT-B/16 recipe \citep{DBLP:conf/cvpr/HeCXLDG22} and performing a search over weight decay and learning rate hyperparameter choices. 
We find that \gls{byol} performs well with heavy weight decay ($\lambda=0.3$)
and a low learning rate
($\eta=10^{-3}$)
at a base batch size $B=4096$.
The AdamW optimizer
is used, and so for scaling to other batch sizes $\hat B=\kappa B$ we use the Adam Scaling Rule (\Cref{def:adam-sr})\footnote{We note that Adam \citep{DBLP:journals/corr/KingmaB14} and AdamW  \citep{DBLP:conf/iclr/LoshchilovH19} are equivalent in the limit of zero weight decay, and that
the Adam Scaling Rule (\Cref{def:adam-sr}) was derived with zero weight decay \citep{DBLP:conf/nips/MalladiLPA22}.
}
We use a pre-bn linear probe as discussed in \Cref{app:sec-components-ssl}.
Finally, the performance of \gls{byol} can be further improved by employing multicrop \citep{DBLP:conf/nips/CaronMMGBJ20} by $\approx$ +2\% in absolute test top-1 performance on ImageNet1k compared to without multicrop, however, as this is not our focus, we omit this from the presented recipe.

\begin{table}[t]
  \caption{BYOL ViT-B/16 hyperparameters.}
  \label{tab:byol-recipe}
  \centering
  \small
  \begin{tabular}{lc}
    \toprule
    & BYOL ViT-B/16 \\
    \midrule
    ImageNet1k Linear Probe Test Top-1 & 74.47\% (\Cref{fig:vitb-byol-ln-vs-bn}) \\
    \midrule
    Weight initialization & \texttt{trunc\_normal(.02)}  \\
    Backbone normalization    & LayerNorm  \\
    Head normalization    & BatchNorm  \\
    Synchronized BatchNorm over replicas & No \\     
    Learning rate schedule & Single Cycle Cosine \\    
    Learning rate warmup (epochs) & 40 \\    
    Learning rate minimum value & $1\times 10^{-6}$  \\    
    Training duration (epochs) & 480 \\
    Optimizer & AdamW \\    
    Optimizer scaling rule & Adam \\
    Base ($\beta_1, \beta_2$) & (0.9, 0.95) \\
    Base learning rate & $1\times 10^{-3}$  \\
    Base batch size & 4096  \\
    Base teacher momentum & 0.99 \\    
    Weight decay & 0.3 \\
    Weight decay scaling rule & None \\
    Weight decay skip bias & Yes \\
    Numerical precision & \texttt{bf16} \\
    Augmentation stack & 
    \texttt{BYOL} \citep{DBLP:conf/nips/GrillSATRBDPGAP20} \\   
    Stochastic depth & 0.1 \\
    \bottomrule
  \end{tabular}
\end{table}

\paragraph{Additional background} 
Achieving large scale \gls{ssl} training with \glspl{vit} to large scale \gls{ssl} training has been a long standing goal in the community. 
MoCo-v3 \citep{DBLP:conf/iccv/ChenXH21} enables the use of \gls{vit}s with contrastive learning, but achieves this through modifications of the \gls{vit} training procedures, including gradient freezing on the image patching layer, and re-introducing Batch Normalization to post-attention MLP layers.
Despite these modifications, MoCo-v3 was only trained up to a batch size of 6144, where model performance begins to suffer \citep{DBLP:conf/iccv/ChenXH21}. 
In \Cref{fig:vitb-byol} we demonstrate that combining dynamic batch scaling (\Cref{subsec:dynamic-batch-scaling}) with the \gls{ema} Scaling Rule (\Cref{def:ema-sr}) enables \gls{byol} to be trained using \glspl{vit} to batch sizes of 24,576 without any drop in performance compared to the reference batch size of 4096. 
We emphasize that the piecewise transitions in the schedules are important for preserving training dynamics.

\subsection{The role of Batch Normalization and Layer Normalization in BYOL with ViTs}
\label{app:byol-vit-ln-vs-bn}
\begin{figure}[th]
    \centering
    \includegraphics[width=0.99\textwidth]{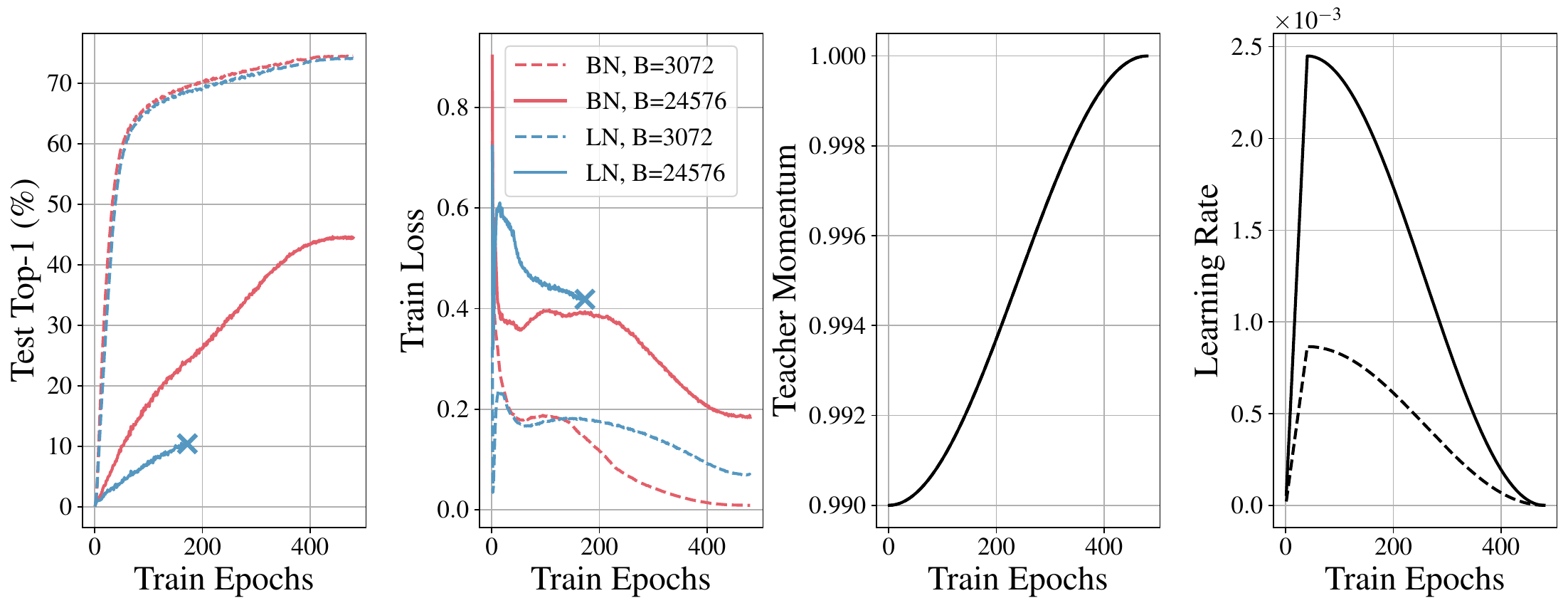}
    \caption{
    \emph{BYOL ViT-B/16 on ImageNet1k} for different scalings $\kappa$.
    We present runs comparing LayerNorm (blue) to BatchNorm (red) in the projection and prediction heads of BYOL \gls{vit} models for batch size 3072 (dashed) and 24,576 (solid) \emph{without the EMA Scaling Rule}.
    $\kappa=1$ corresponds to $B=4096$.
    In all scenarios the transformer backbone \emph{only} uses LayerNorm. 
    We truncate the training of the large batch size LayerNorm variant to preserve compute (indicated by $\times$).
    }
    \label{fig:vitb-byol-ln-vs-bn}
\end{figure}

Here we compare the roles of Batch Normalization (BatchNorm, \cite{DBLP:conf/icml/IoffeS15}) and  Layer Normalization (LayerNorm, \cite{DBLP:journals/corr/BaKH16})
in the projection and prediction heads of \gls{byol} \citep{DBLP:conf/nips/GrillSATRBDPGAP20} using \glspl{vit}.

It has been observed that BatchNorm 
plays a critical role in \gls{byol} predictor and projector dynamics \citep{Fetterman_Albrecht_2020}, 
and using either LayerNorm or \emph{no normalization} significantly decreases model performance.
Subsequently, it was demonstrated \citep{DBLP:journals/corr/abs-2010-10241} that competitive \gls{byol} performance could be achieved through a combination of Group Normalization (GroupNorm, \cite{DBLP:conf/eccv/WuH18}) and Weight Standardization \citep{DBLP:journals/corr/abs-1903-10520}.
Additionally, \citet{DBLP:journals/corr/abs-2010-10241} showed that if BatchNorm is used in the backbone, one can use LayerNorm or \emph{no normalization} in the predictor and projector without any performance drop.

In this work, we we show it is possible to train \gls{byol} \gls{vit} using \emph{only LayerNorm} across the backbone, projector and predictor (see \Cref{fig:vitb-byol-ln-vs-bn}),
decoupling \gls{byol}'s reliance on batch statistics, a desirable trait for a representation learning algorithm \citep{DBLP:conf/icml/BrockDSS21}. 
At batch size 3072, using LayerNorm in the predictor and projector achieves competitive performance (74.10\%), performing slightly worse than using BatchNorm (74.47\%). 
At the larger batch size of 24,576, runs perform significantly worse as the \gls{ema} Scaling Rule was not applied.

\subsection{Longer training duration with incremental Progressive Scaling}
\label{app:byol-waterfall}
\begin{figure}[ht]
    \centering
    \includegraphics[width=0.99\textwidth]{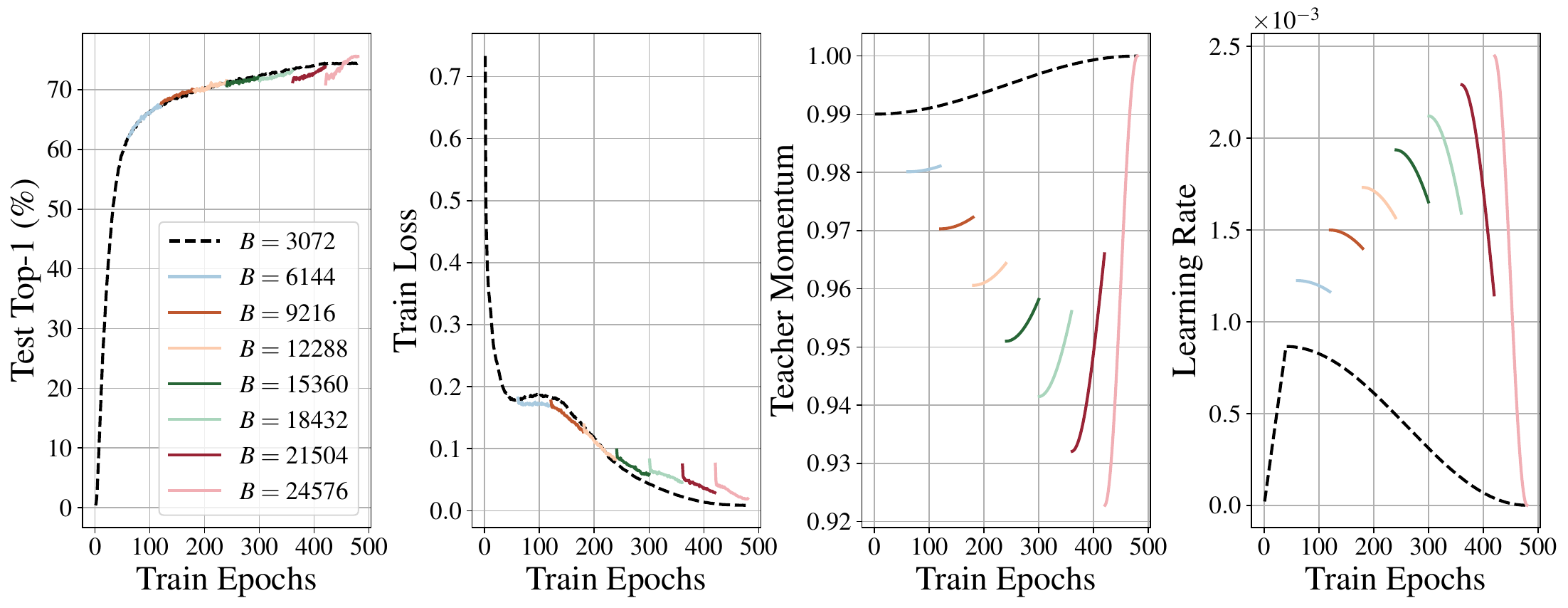}
    \caption{
    \emph{BYOL ViT-B/16 on ImageNet1k} for different scalings $\kappa$.
    The baseline model ($\kappa=0.75$, black dashed) uses batch size 3072 and teacher momentum
    $\rho_B=0.99$. 
    We increment the batch size by 3072 every 60 epochs to a final batch size of 24,576 using Progressive Scaling (\Cref{def:progressive-scaling}). 
    }
    \label{fig:vitb-byol-waterfall}
\end{figure}

Here we use the same base hyperparameters as \Cref{tab:byol-recipe}, except that we train for 480 instead of 300 epochs.
To mitigate the student impulse phenomena discussed in \Cref{subsec:self-supervised}, in \Cref{fig:vitb-byol-waterfall} we investigate increasing the batch size every 60 epochs using Progressive Scaling  (\Cref{def:progressive-scaling}).
We observe that this more gradual procedure enables closer tracking of the baseline train loss trajectory.
Additionally, this procedure results in a scaled linear probe performance that outperforms the baseline (75.64\% compared to the baseline performance of 74.47\%).
The same procedure can be applied to the LayerNorm variant discussed in \Cref{app:byol-vit-ln-vs-bn}, which produces a similar result (75.09\% compared to the baseline performance of 74.10\%).

\subsection{Building intuition around Progressive Scaling and momentum sensitivity}
\label{app:byol-progressive-scaling-regimes}
\begin{figure}[ht]
    \centering
    \includegraphics[width=0.99\textwidth]{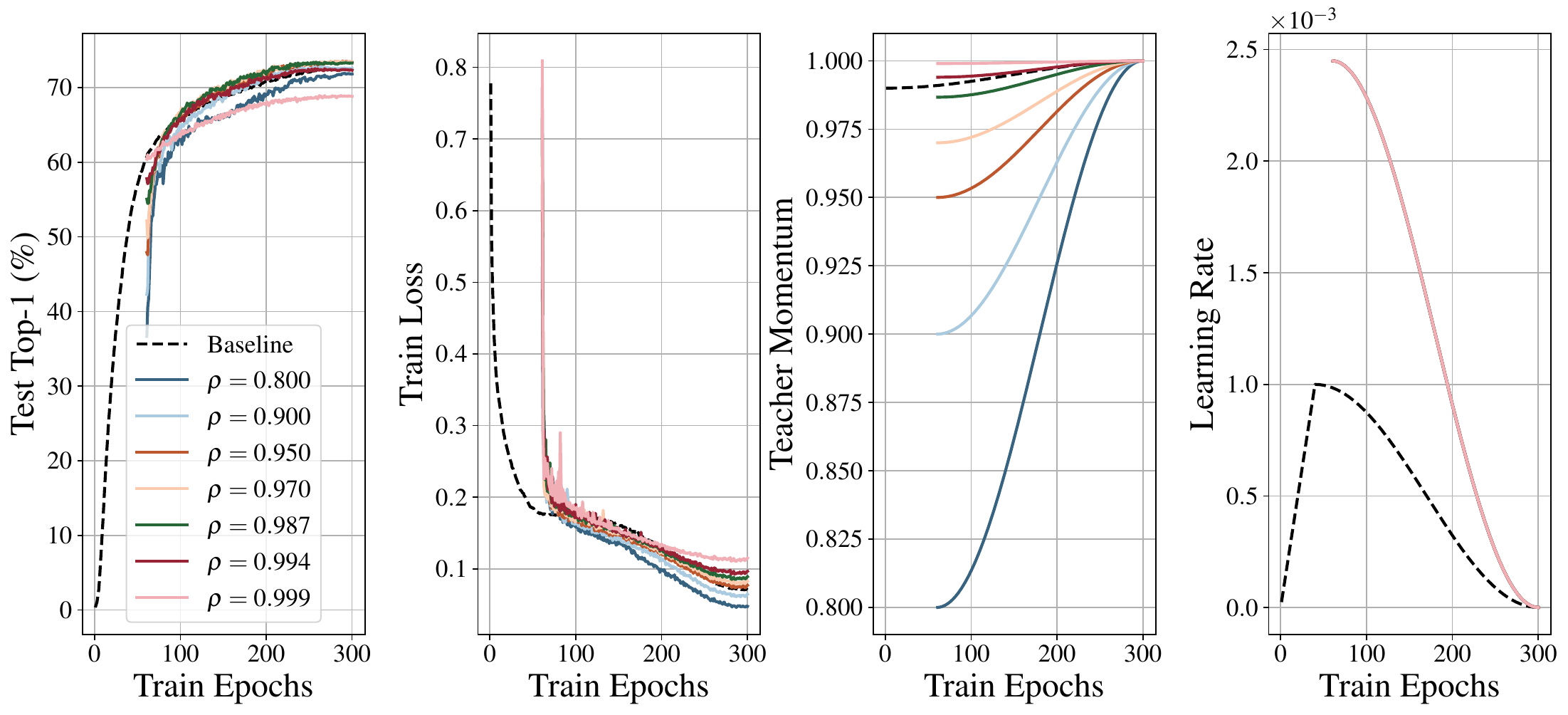}
    \caption{
    \emph{BYOL ViT-B/16 on ImageNet1k} for different momenta $\rho$.
    The baseline model ($\rho=0.99$, black dashed) uses batch size 4096. At the 60th epoch we apply Progressive Scaling (\Cref{def:progressive-scaling}) and transition to batch size 24576.
    We train for a further 240 epochs without EMA scaling for a range of momenta: $\rho \in \{0.9, 0.95, 0.97, 0.9867, 0.994\}$.
    }
    \label{fig:vitb-byol-rho-ablations}
\end{figure}
Our final \gls{byol} \gls{vit} results are to help build intuition around Progressive Scaling (\Cref{def:progressive-scaling}),
as well as when the \gls{ema} Scaling Rule is most important.
In \Cref{fig:vitb-byol-rho-ablations} we explore transitioning from the baseline batch size 4096 model to batch size 24,576 in a \emph{single transition} after 60 epochs.
After this transition, we continue training for 240 epochs for a range of momenta: $\rho \in \{0.8, 0.9, 0.95, 0.97, 0.9867, 0.994, 0.999\}$ \emph{without} the EMA Scaling Rule. 

We observe that after the transition, any $0.9\leq \rho\leq0.994$ produces a linear probe performance that matches or outperforms the baseline at the end of training. 
This indicates that after the initial training period, BYOL becomes less sensitive to the choice of teacher momentum. 
Note that without the initial 60 epochs of training with batch size 4096, \emph{all models}, including those employing the \gls{ema} Scaling Rule diverge (see $B=24,576$ in \Cref{fig:vitb-byol}).

\begin{figure}[ht]
    \centering
    \includegraphics[width=0.4\textwidth]{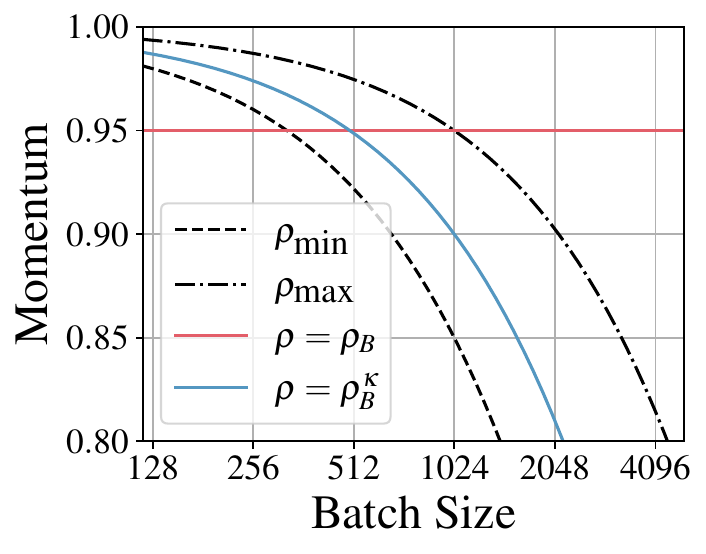}
    \caption{
    \emph{A hypothetical scenario where there is an upper and lower limit for momenta qualitatively leading to the same result.}
    We assume at base batch size $B=1024$ there is an upper ($\rho_{\text{max}}$, black dashdot) and lower ($\rho_{\text{min}}$, black dashed) limit for \emph{valid momenta}.
    We show what happens if we start with $\rho_B=0.95$ at a batch size of 4096, and scale with ($\rho=\rho_B^\kappa$, blue) and without ($\rho=\rho_B$, red) the \gls{ema} Scaling Rule.
    }
    \label{fig:robustness-cartoon}
\end{figure}

We present an illustration for why this might happen in \Cref{fig:robustness-cartoon}.
First, we see that using the \gls{ema} Scaling Rule \emph{always} keeps the model within the acceptable momentum region.
We also wee that \emph{not} using the \gls{ema} Scaling Rule can keep the model  within the acceptable momentum region for a range of batch sizes, depending on how large wide in momenta the acceptable region is at the base batch size.
Finally, we see that the momentum value matters much more at low values of momenta (the acceptable momentum region shrinks), whereas at large momenta, this region of acceptability widens.

\ifthenelse{\equal{\anonymous}{0}}{
\subsection{Compute usage for ViT BYOL investigation}

We now summarize the compute usage for the BYOL 
\gls{vit} experiments.
First we detail cost of reproducing 
\Cref{fig:vitb-byol} in \Cref{tab:byol-vit-compute},
\Cref{fig:vitb-byol-ln-vs-bn} in \Cref{tab:byol-vit-compute-2},
and \Cref{fig:vitb-byol-waterfall} in \Cref{tab:byol-vit-compute-3}.

\begin{table}[ht]
  \caption{
  Compute usage for baseline ViT BYOL investigation in \Cref{fig:vitb-byol}. 
  Values \emph{include} node allocation times (typically a small \% of corresponding total runtime), giving a practical   estimate of reproduction cost. 
  All experiments conducted are using 80Gb A100s, and experiments indicated by $(\dagger)$ have a faster interconnect.}
  \label{tab:byol-vit-compute}
  \centering
  \small
  \begin{tabular}{ccccccc}
\toprule
 Batch Size &  GPUs &  Time (h) &  Compute/Run (GPUh) &  Runs &  Compute (GPUh) \\
\midrule
       4,096 &      32 &          16.6 &               531.4 &       2 &          1,062.7 \\
       8,192 &      48 &          14.1 &               678.0 &       2 &          1,356.1 \\
      16,384 &      96 &           8.3 &               800.3 &       2 &          1,600.6 \\
      24,576 &     128 &           6.6 &               850.1 &       4 &          3,400.4 \\
      32,768${}^{\dagger}$ &     176 &           4.1 &               721.7 &       2 &          1,443.4 \\ \midrule 
      \textbf{Total} &&&&& \textbf{8,863.1} \\
\bottomrule
  \end{tabular}
\vspace{-0.2cm}
\end{table}

\begin{table}[ht]
  \caption{
  Compute usage for ViT BYOL investigation into BatchNorm and LayerNorm variants in \Cref{fig:vitb-byol-ln-vs-bn}.
  Values \emph{include} node allocation times (typically a small \% of corresponding total runtime), giving a practical   estimate of reproduction cost. 
  All experiments conducted are using 80Gb A100s, and were run for 480 epochs, except those indicated by $(*)$ were truncated early (see \Cref{fig:vitb-byol-ln-vs-bn} for more details).}
  \label{tab:byol-vit-compute-2}
  \centering
  \small
\begin{tabular}{cccccc}
\toprule
 Batch Size & Normalization &  GPUs &  Time (h) &  Compute (GPUh) \\
\midrule
       3,072 &     BatchNorm &    16 &      47.9 &           766.0 \\
       3,072 &     LayerNorm &    16 &      48.0 &           768.7 \\
      24,576 &     BatchNorm &   128 &      14.8 &          1900.4 \\
      24,576${}^{*}$ &     LayerNorm &   128 &       3.5 &           451.1 \\ \midrule
      \textbf{Total} &&&& \textbf{3,886.2} \\
\bottomrule
\end{tabular}
\vspace{-0.2cm}
\end{table}

\begin{table}[ht]
  \caption{
  Compute usage for ViT BYOL investigation into incremental scaling in \Cref{fig:vitb-byol-waterfall}.
  Values \emph{include} node allocation times (typically a small \% of corresponding total runtime), giving a practical   estimate of reproduction cost. 
  All experiments conducted are using 80Gb A100s for 60 epochs. Stage 0 corresponding to the baseline in \Cref{fig:vitb-byol-waterfall} is the run detailed in the first row of \Cref{tab:byol-vit-compute-2}, using a batch size of 3,072, Batch Normalization, and 16 GPUs. Computing only the first 60 epochs of stage 0 corresponds to approximately 127.7 GPUh, which would bring the total cost of \Cref{fig:vitb-byol-waterfall} to 1,432.9 GPUh.}
  \label{tab:byol-vit-compute-3}
  \centering
  \small
\begin{tabular}{cccccc}
\toprule
 Stage &  Batch Size &    GPUs &  Time (h) &  Compute (GPUh) \\
\midrule
     1 &        6,144 &        32 &       3.5 &           113.0 \\
     2 &        9,216 &          48 &       3.1 &           149.8 \\
     3 &       12,288 &          64 &       2.8 &           176.0 \\
     4 &       15,360 &          80 &       2.3 &           186.5 \\
     5 &       18,432 &          96 &       2.1 &           202.9 \\
     6 &       21,504 &         112 &       2.1 &           235.8 \\
     7 &       24,576 &         128 &       1.9 &           241.3 \\ \midrule 
     \textbf{Total} &&&& \textbf{1,305.2} \\
\bottomrule
\end{tabular}
\vspace{-0.2cm}
\end{table}

\FloatBarrier

Next, the cost of a single momentum ablation presented in \Cref{fig:vitb-byol-rho-ablations}
is 240 epochs at batch size 24,576, which is $\approx 240/480 \times 1900.4\text{ GPUh}=950.2\text{ GPUh}$, giving a total cost over seven runs of $\mathbf{6651.4}\text{ GPUh}$.

Finally, providing a full view of the investigations carried out for the \gls{vit} \gls{byol} is given in \Cref{tab:byol-vit-compute-summary}.

\begin{table}[ht]
  \caption{
  Total compute usage for ViT BYOL investigations.}
  \label{tab:byol-vit-compute-summary}
  \centering
  \small
\begin{tabular}{lc}
\toprule
 &  Compute (GPUh) \\
\midrule
Baselines (\Cref{fig:vitb-byol} and \Cref{tab:byol-vit-compute}) & 8,863.1 \\
BatchNorm and LayerNorm (\Cref{fig:vitb-byol-ln-vs-bn} and \Cref{tab:byol-vit-compute-2}) & 3,886.2 \\
Incremental scaling (\Cref{fig:vitb-byol-waterfall} and \Cref{tab:byol-vit-compute-3}) & 1,305.2 \\
Momentum ablations (\Cref{fig:vitb-byol-rho-ablations}) & 6,651.4 \\
All other compute, e.g. code development, runs with errors, and debugging & 84,984.1 \\
\midrule 
     \textbf{Total} & \textbf{105,690.0} \\
\bottomrule
\end{tabular}
\vspace{-0.2cm}
\end{table}

}{\compute}

\FloatBarrier

\subsection{ResNet-18 hyperparameter sensitivity analysis}
\label{subsec:byol-sensitivity-analysis}

To demonstrate that the \gls{ema} Scaling Rule works for a broad range of optimization hyperparameters (i.e. \emph{beyond} those presented in 
\Cref{fig:r18-byol} and \Cref{subsec:self-supervised}), we provide a sensitivity analysis for base teacher momentum $\rho_B$ and base learning rate $\eta_B$ in the challenging setting of \gls{byol}.

\paragraph{Base teacher momentum}

\begin{figure}[ht]
    \centering
    \begin{subfigure}[b]{0.49\textwidth}
        \centering
        \includegraphics[width=\textwidth]{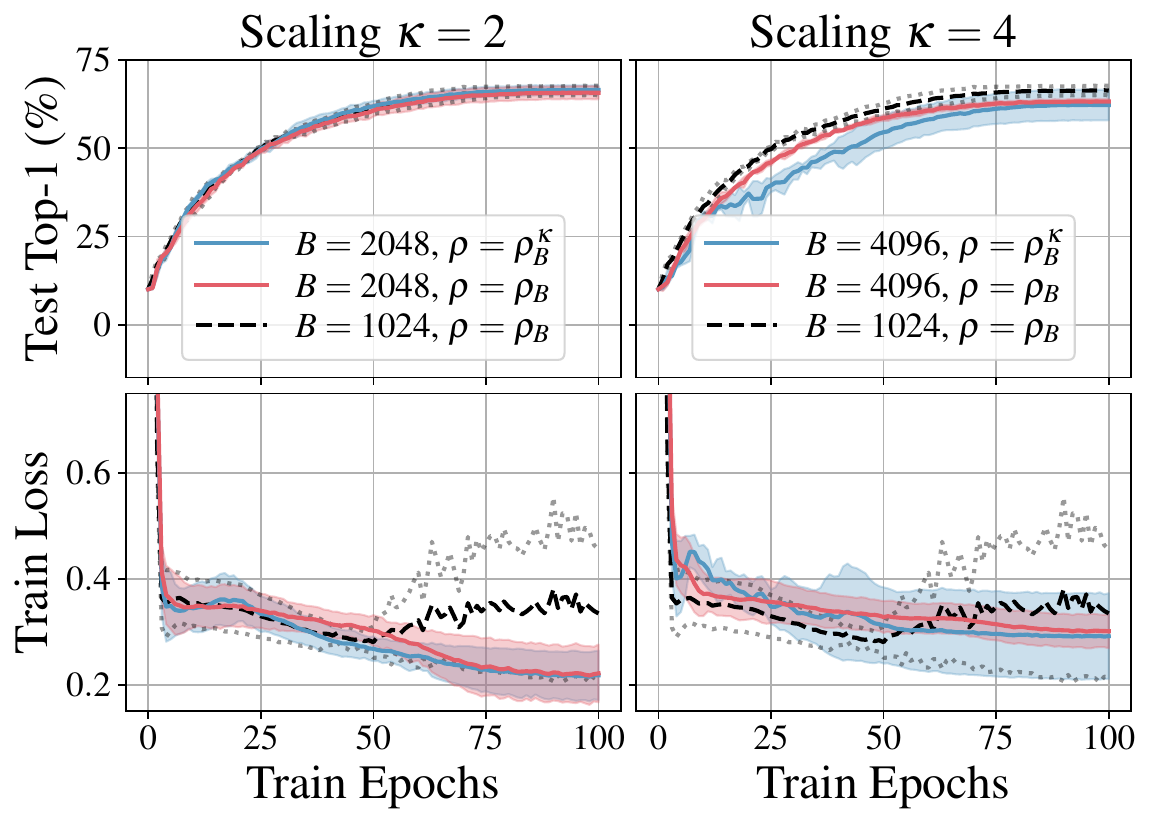}
        \caption{$\rho_B=0.841$}
    \end{subfigure}
    \hfill
    \begin{subfigure}[b]{0.49\textwidth}
        \centering
        \includegraphics[width=\textwidth]{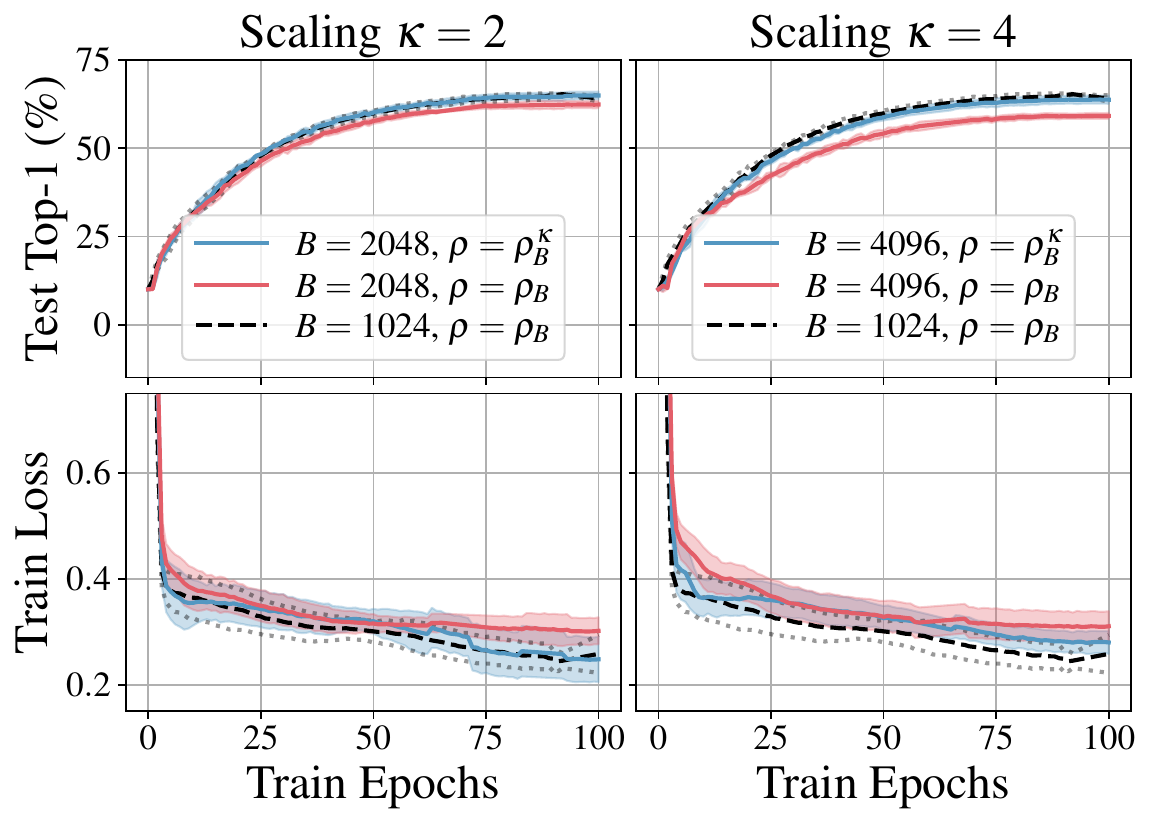}
        \caption{$\rho_B=0.946$}
    \end{subfigure}
    \\
    \begin{subfigure}[b]{0.49\textwidth}
        \centering
        \includegraphics[width=\textwidth]{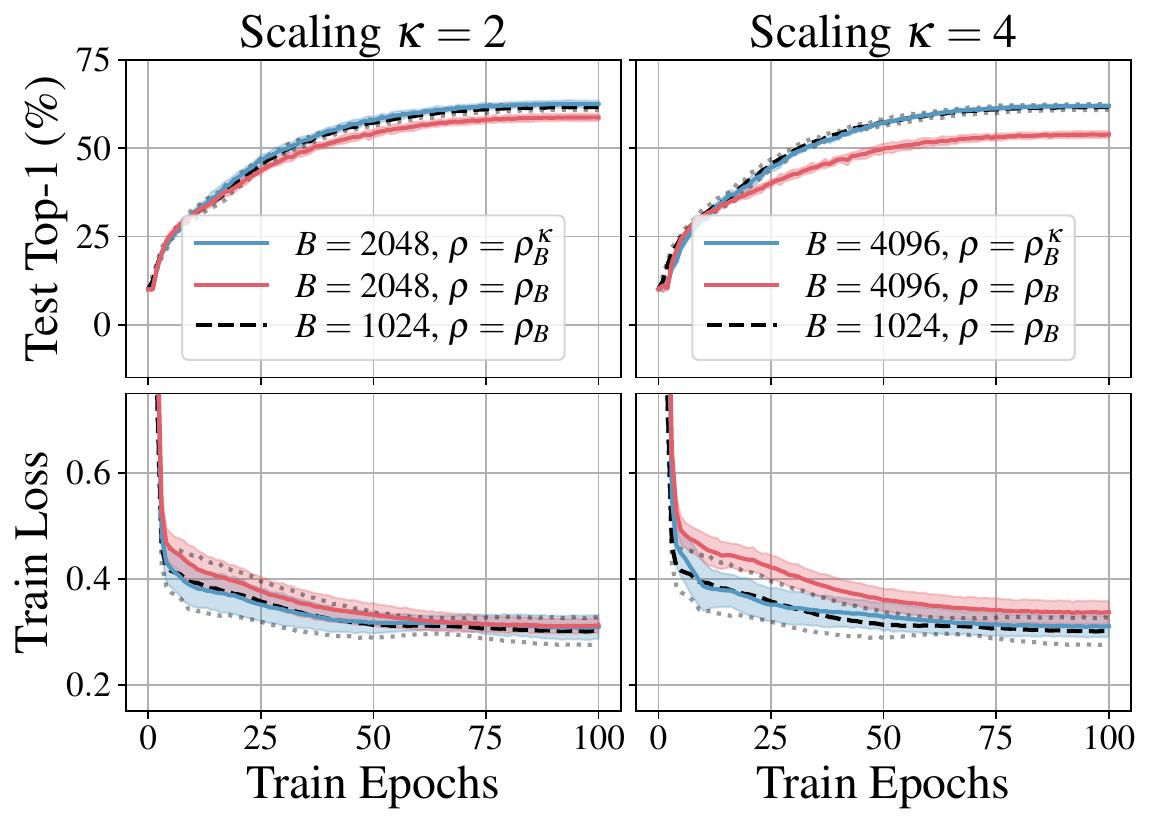}
        \caption{$\rho_B=0.974$}
    \end{subfigure}    
    \hfill
    \begin{subfigure}[b]{0.49\textwidth}
        \centering
        \includegraphics[width=\textwidth]{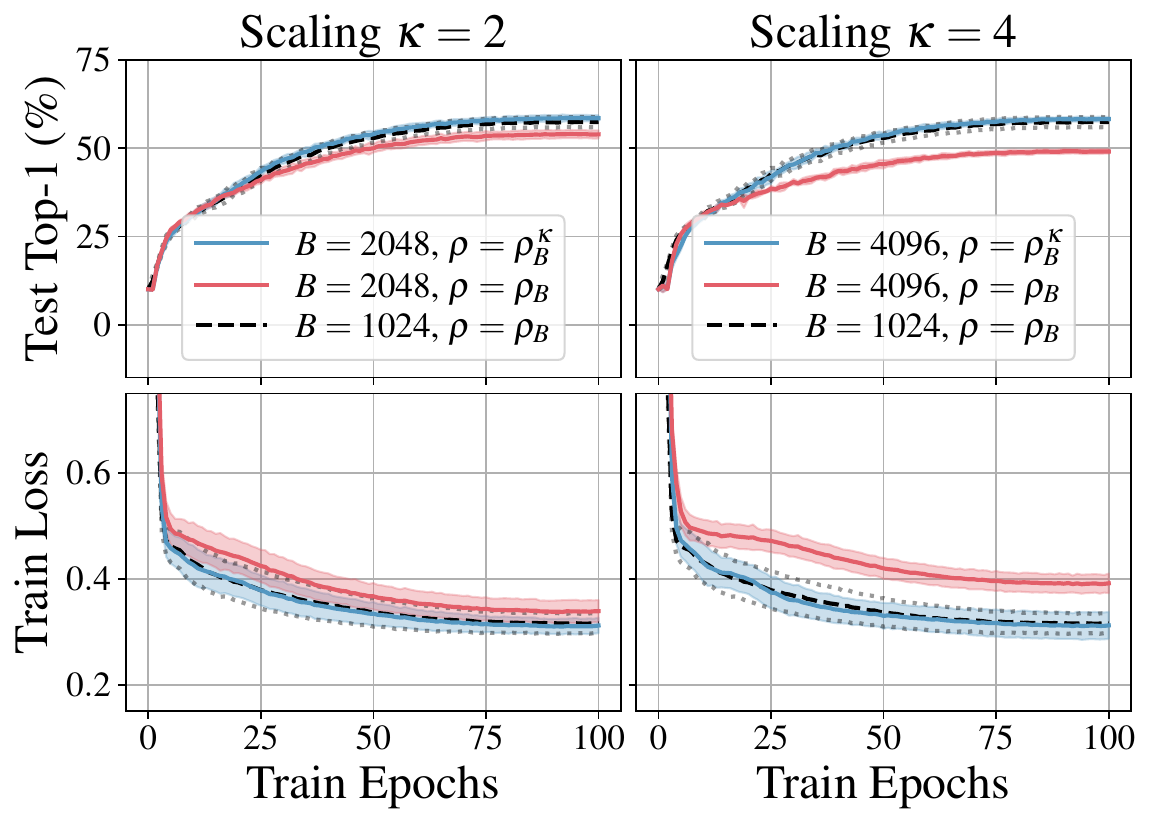}
        \caption{$\rho_B=0.987$}
    \end{subfigure}
    \\
    \begin{subfigure}[b]{0.49\textwidth}
        \centering
        \includegraphics[width=\textwidth]{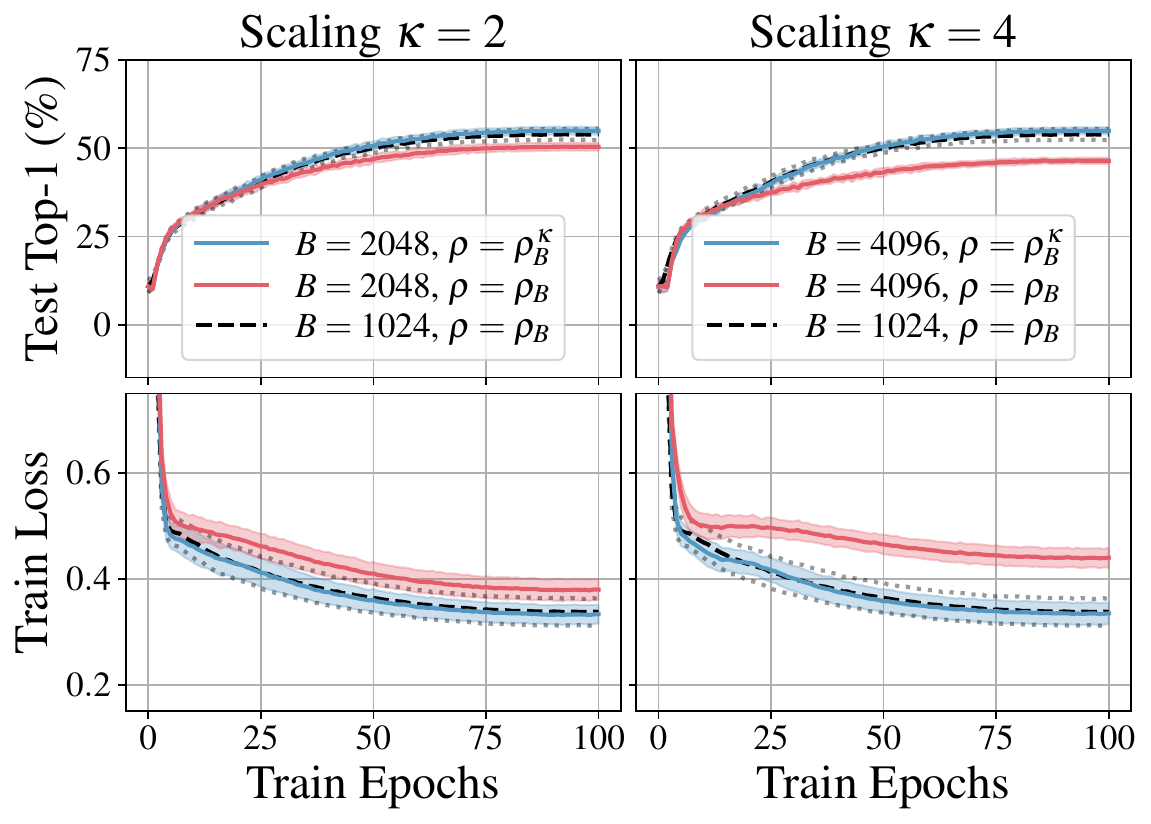}
        \caption{$\rho_B=0.992$}
    \end{subfigure}
    \hfill
    \begin{subfigure}[b]{0.49\textwidth}
        \centering
        \includegraphics[width=\textwidth]{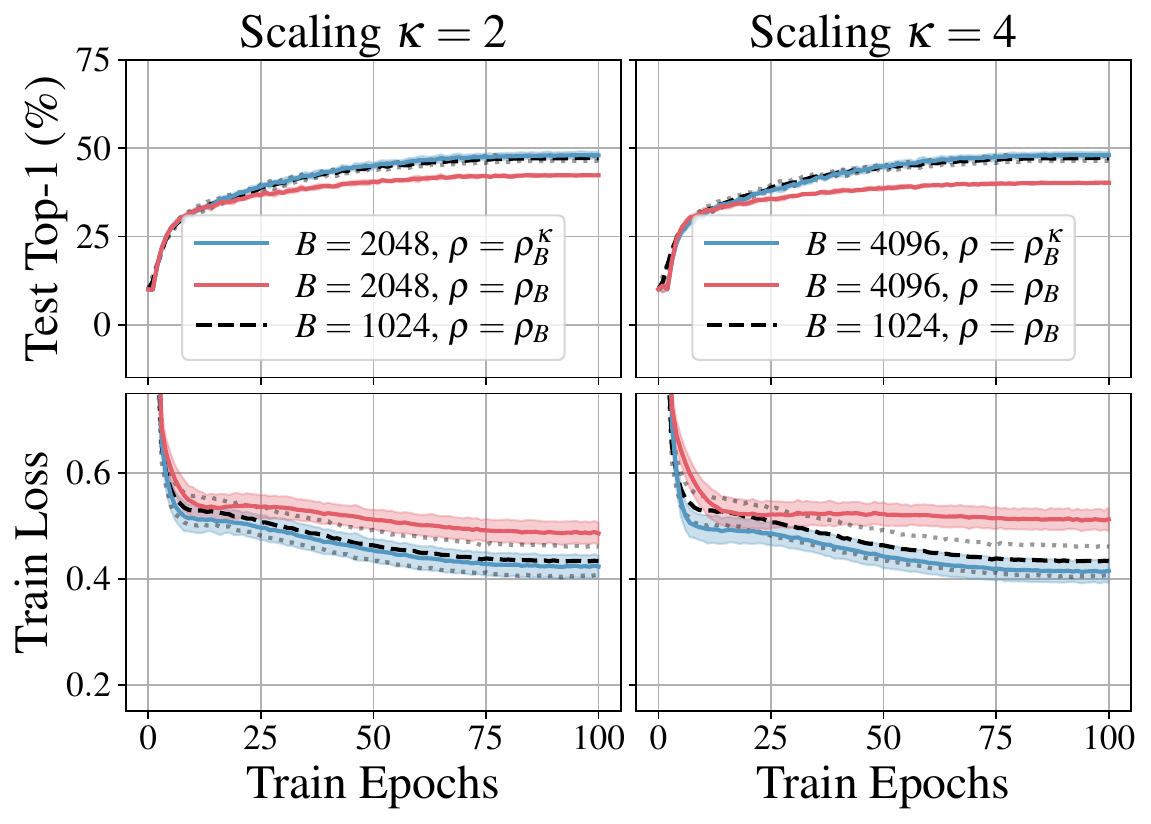}
        \caption{$\rho_B=0.997$}     
    \end{subfigure}
    \caption{
    \emph{ResNet-18 BYOL on CIFAR10 teacher momentum sensitivity}
    ($\eta_B=0.08$)
    for scalings $\kappa\in\{2,4\}$ and base teacher momenta $\rho_B\in\{0.841, 0.946, 0.974,0.987,0.992,0.997\}$ defined at $\kappa=1$.
    The baseline ($\kappa=1$, black dashed)
    uses batch size 1024,
    and is scaled from batch size 2048 (left) to 4096 (right) with (blue, $\rho=\rho_B^\kappa$) and without (red, $\rho=\rho_B$) the \gls{ema} Scaling Rule.
    Bands indicate mean and standard deviation across three runs.}
    \label{fig:momentum-ablation}
\end{figure}

In \Cref{fig:momentum-ablation} we show the effect of changing the base teacher momentum $\rho_B$, defined at batch size 1024.
The \gls{ema} Scaling Rule is robust to modifications of momentum down to $\rho_B\approx 0.946$ in this particular setting.
Below $\rho_B\approx 0.946$, matching is poor, 
although the smallest momentum in this setting corresponds to $0.841^4\approx 0.5$, which is a particularly small teacher momentum, and is unlike to provide utility over the using the target model (see \Cref{app:asymptoticAnalysis}).

\FloatBarrier

\begin{figure}[ht]
    \centering
    \begin{subfigure}[b]{0.49\textwidth}
        \centering
        \includegraphics[width=\textwidth]{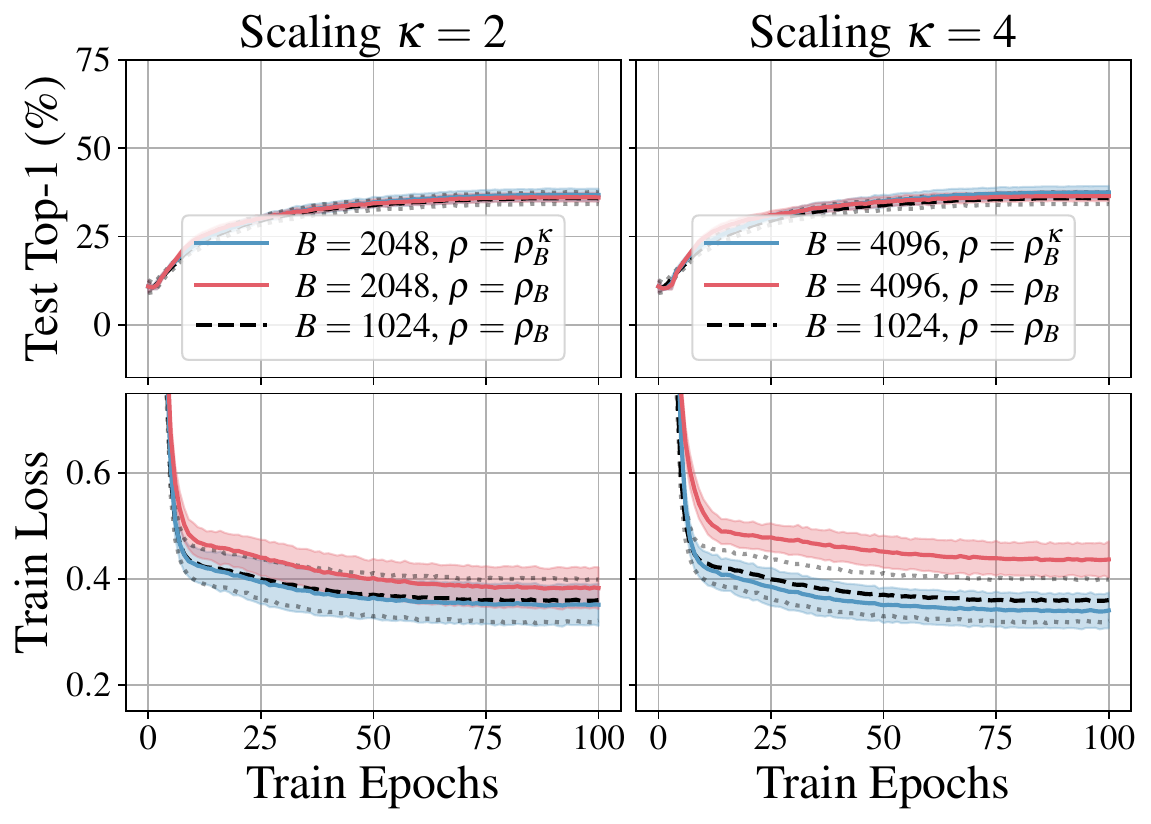}
        \caption{$\eta_B=0.01$}
    \end{subfigure}
    \hfill
    \begin{subfigure}[b]{0.49\textwidth}
        \centering
        \includegraphics[width=\textwidth]{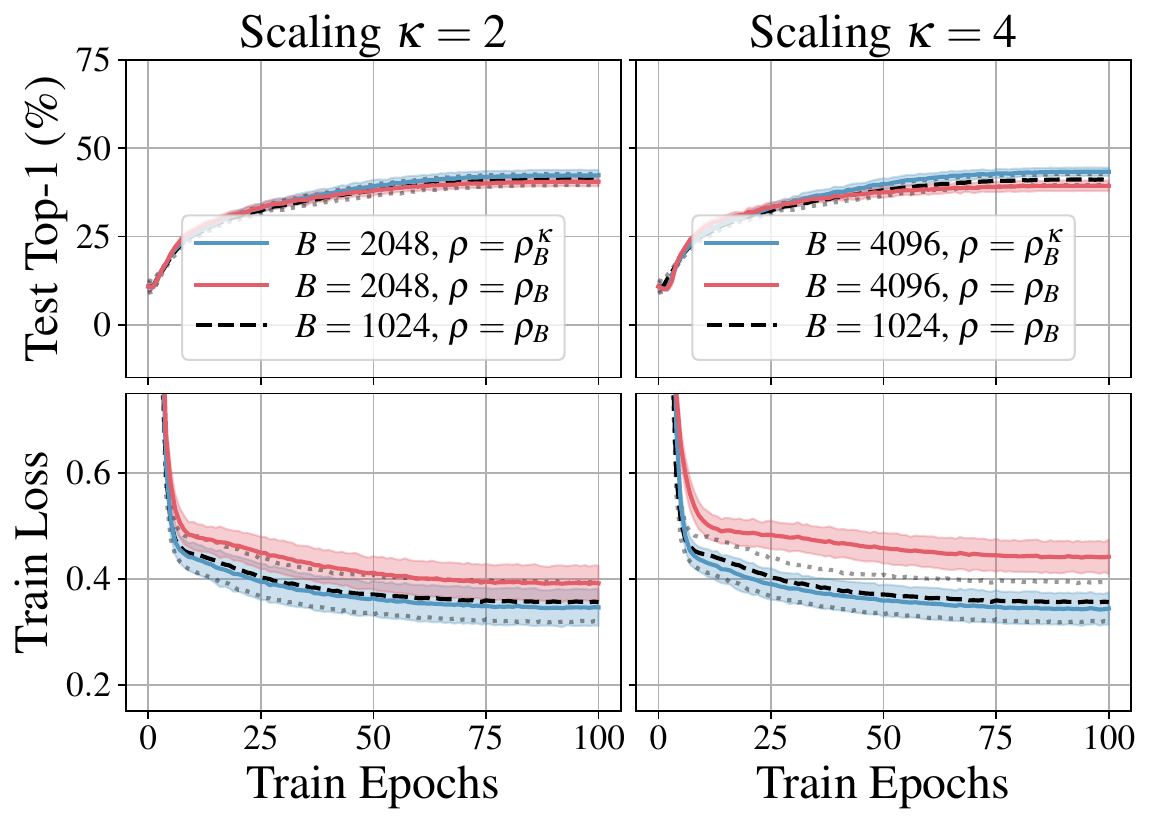}
        \caption{$\eta_B=0.02$}
    \end{subfigure}
    \\
    \begin{subfigure}[b]{0.49\textwidth}
        \centering
        \includegraphics[width=\textwidth]{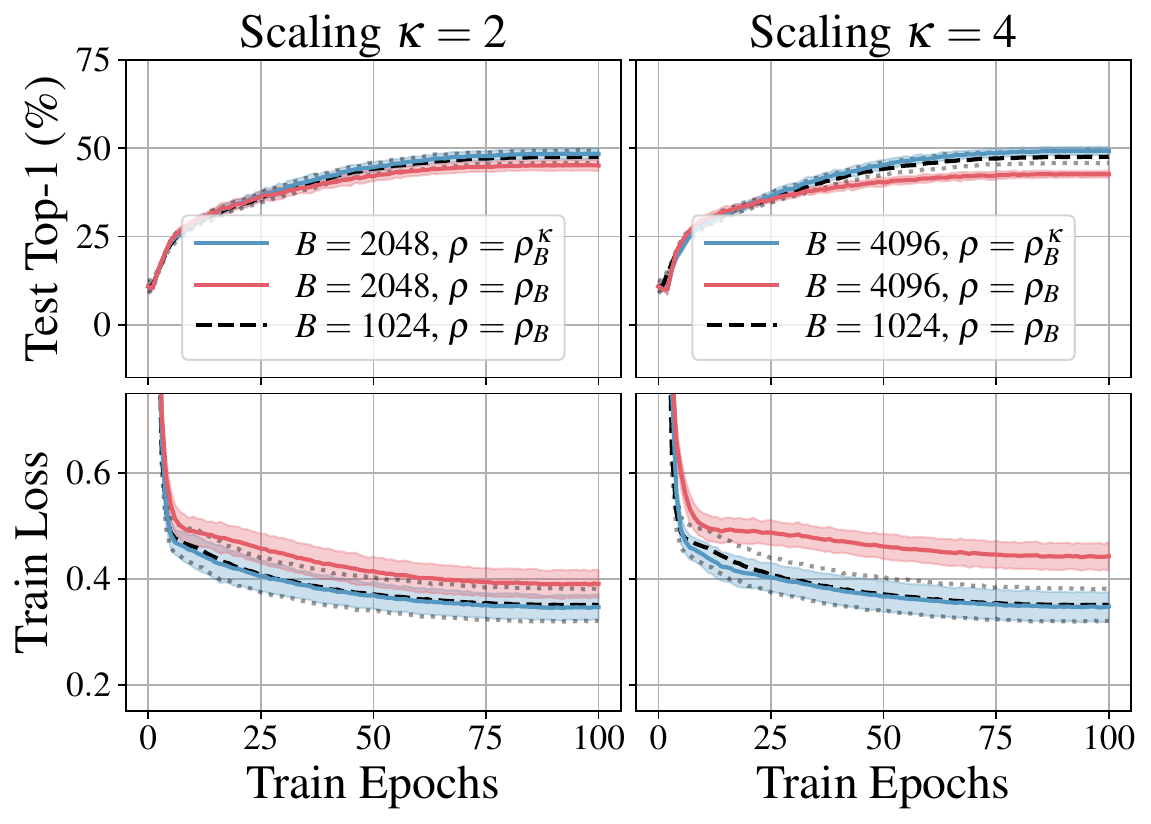}
        \caption{$\eta_B=0.04$}
    \end{subfigure}    
    \hfill
    \begin{subfigure}[b]{0.49\textwidth}
        \centering
        \includegraphics[width=\textwidth]{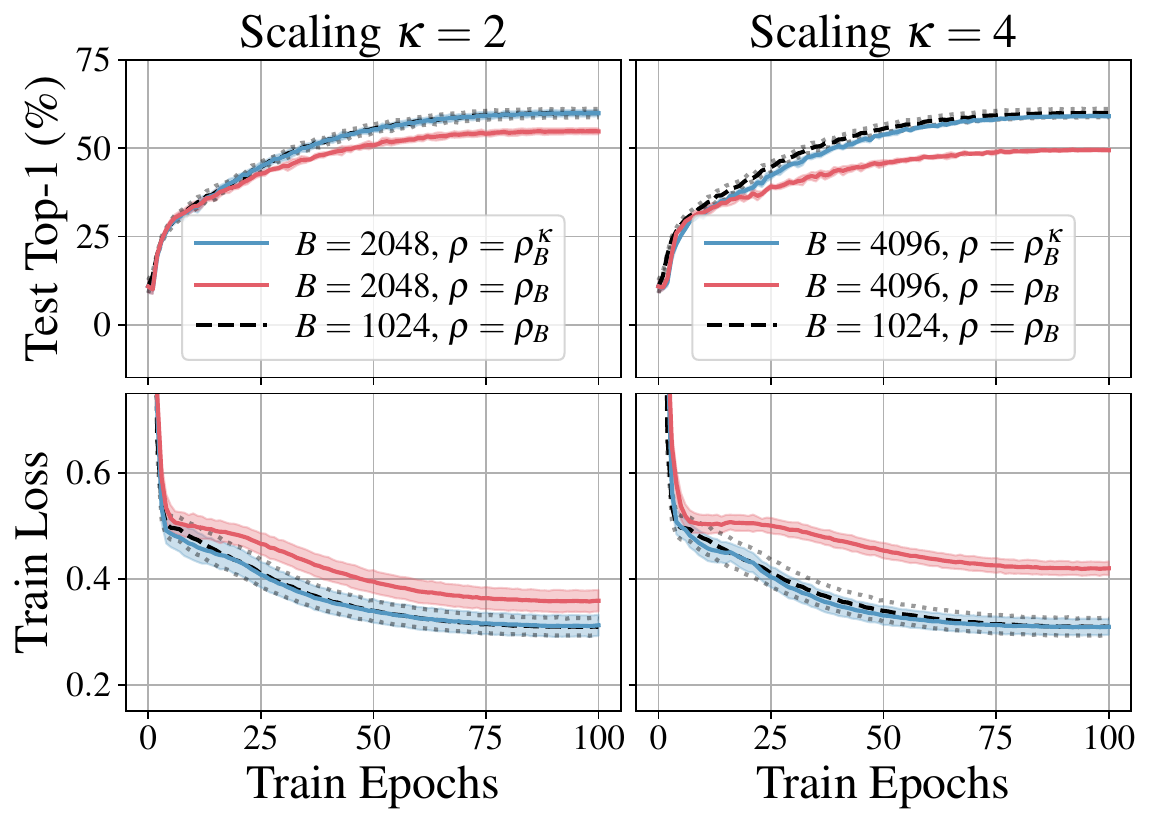}
        \caption{$\eta_B=0.15$}
    \end{subfigure}    
    \\    
    \begin{subfigure}[b]{0.49\textwidth}
        \centering
        \includegraphics[width=\textwidth]{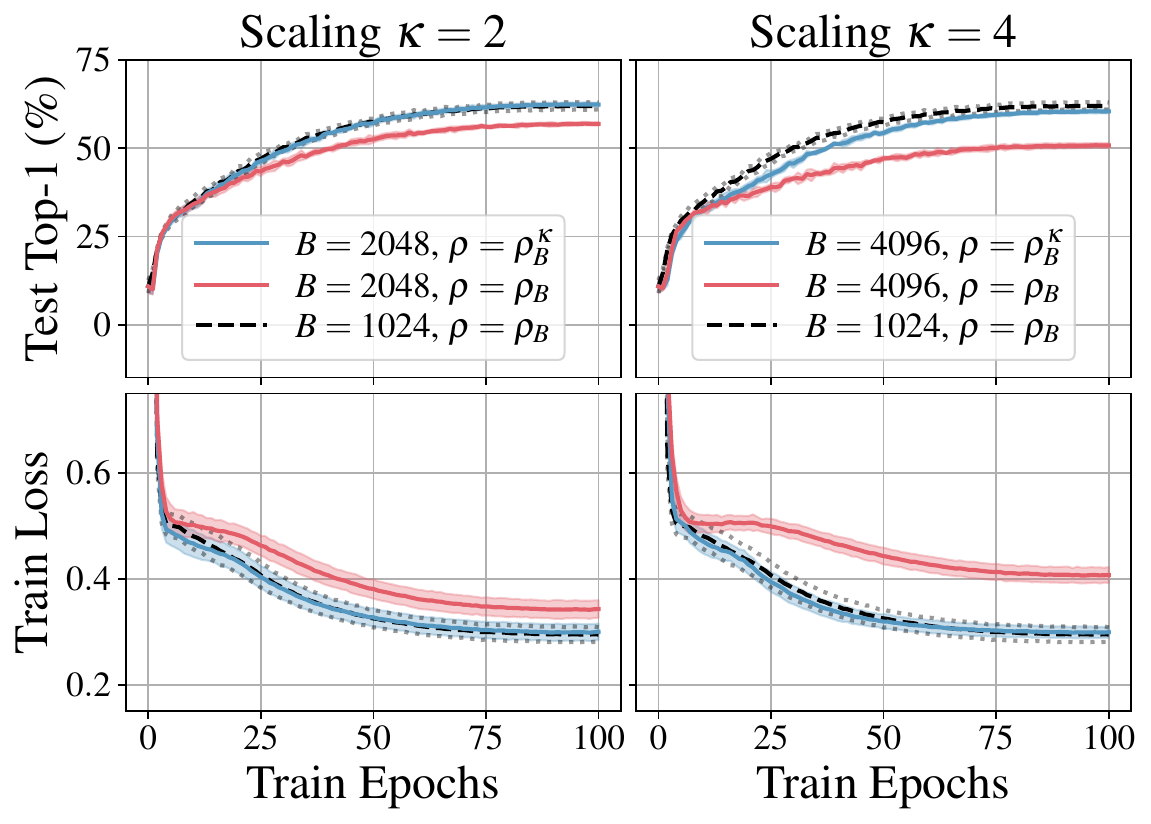}
        \caption{$\eta_B=0.2$}
    \end{subfigure}    
    \hfill
    \begin{subfigure}[b]{0.49\textwidth}
        \centering
        \includegraphics[width=\textwidth]{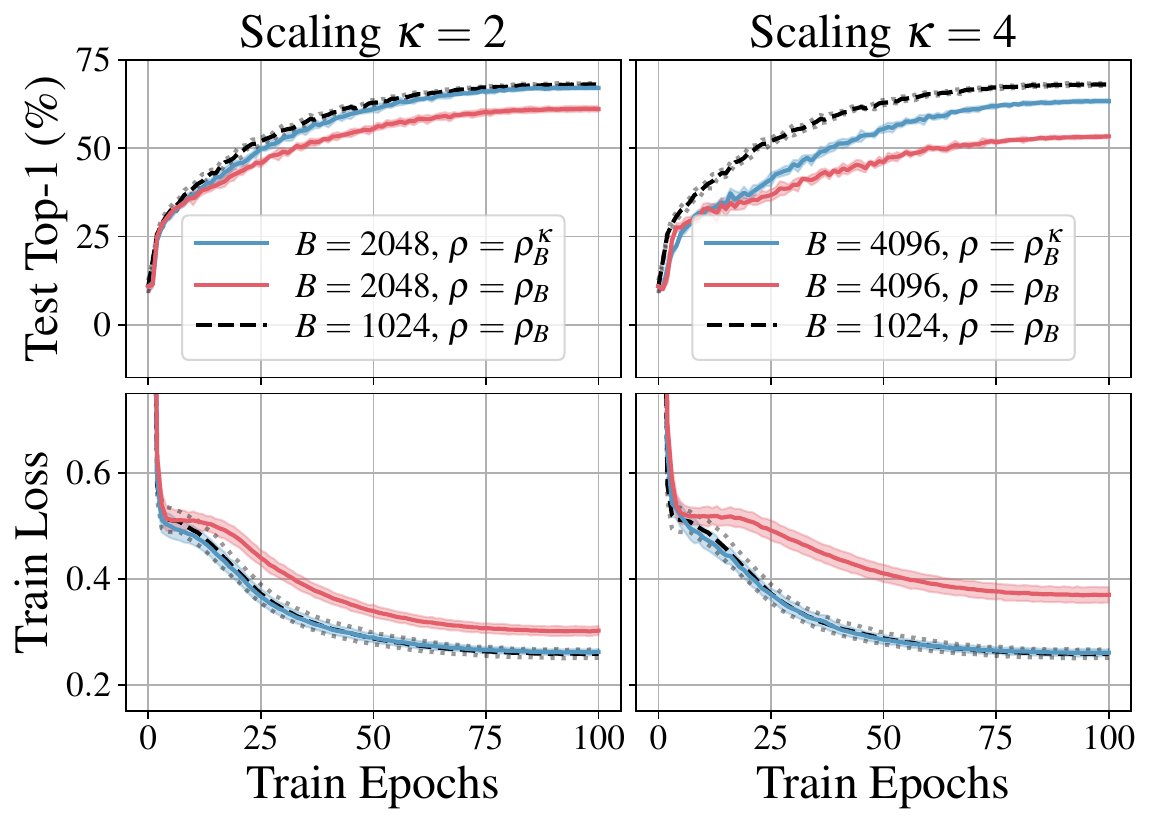}
        \caption{$\eta_B=0.5$}
    \end{subfigure}
    \caption{
    \emph{ResNet-18 BYOL on CIFAR10 learning rate sensitivity}
    ($\rho_B=0.992$)
    for scalings $\kappa\in\{2,4\}$ and base learning rates $\eta_B\in\{0.01,0.02,0.04,0.15,0.20,0.50\}$ defined at $\kappa=1$.
    The baseline ($\kappa=1$, black dashed)
    uses batch size 1024,
    and is scaled from batch size 2048 (left) to 4096 (right) with (blue, $\rho=\rho_B^\kappa$) and without (red, $\rho=\rho_B$) the \gls{ema} Scaling Rule.
    Bands indicate mean and standard deviation across three runs.}
    \label{fig:lr-ablation}
\end{figure}

\newpage
\paragraph{Base learning rate}
In \Cref{fig:lr-ablation} we show the effect of changing the base learning rate $\eta_B$, defined at batch size 1024.
The \gls{ema} Scaling Rule is robust over a wide range of learning rates.
At the largest learning rate $\eta_B=0.5$ matching starts to become poor at scaling $\kappa=4$.

\FloatBarrier

\subsection{ResNet-18 additional scaling analysis}
\label{subsec:byol-scaling-analysis}
\begin{figure}[th]
  \centering
  \includegraphics[width=0.82\textwidth]{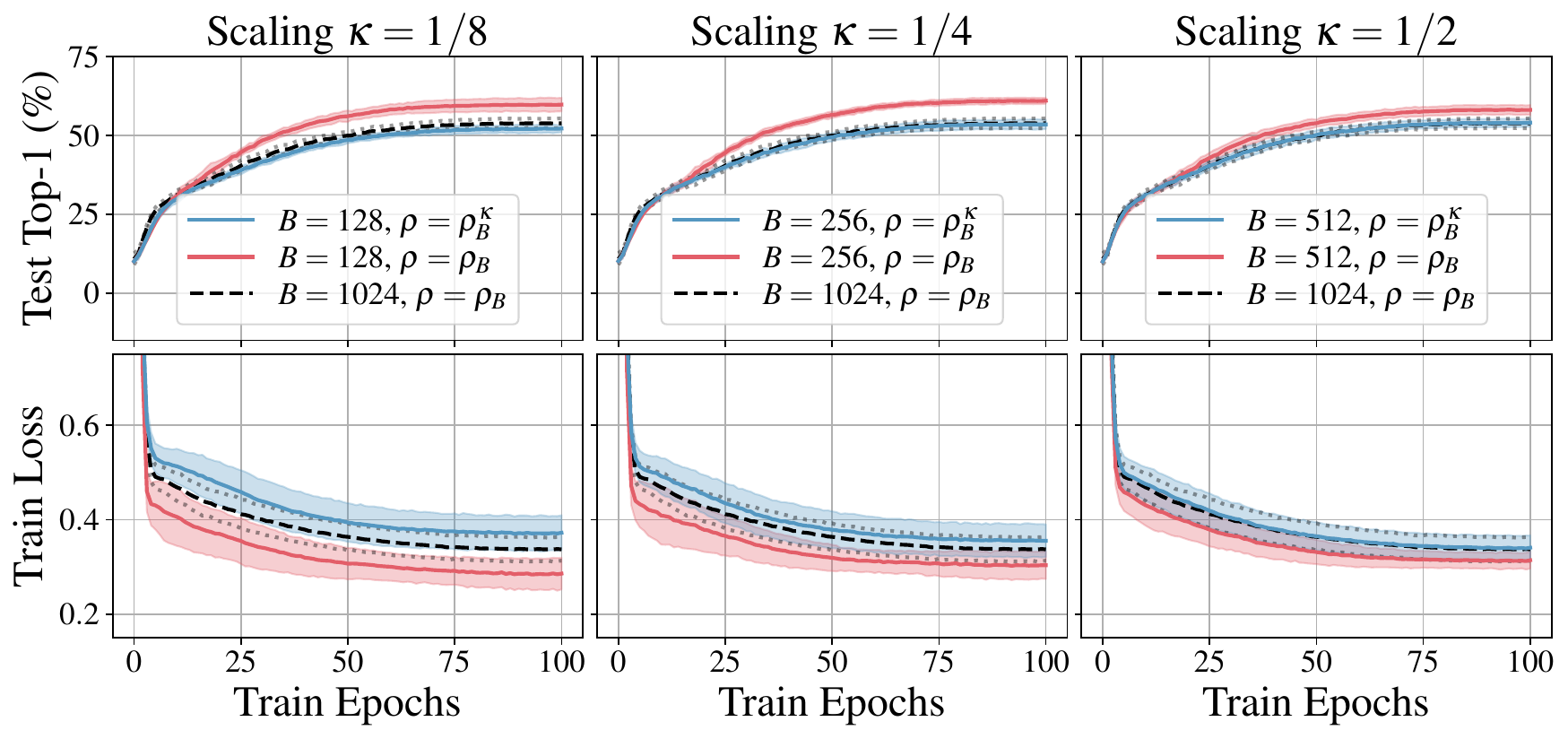}
  \label{fig:scaling-ablation}
  \caption{
      \emph{ResNet-18 BYOL on CIFAR10 lower scaling analysis}
    ($\eta_B=0.08$, $\rho_B=0.992$)
    for scalings $\kappa\in\{1/8,1/4,1/2\}$.
    The baseline ($\kappa=1$, black dashed)
    uses batch size 1024,
    and is scaled from batch size 128 (left) to 512 (right) with (blue, $\rho=\rho_B^\kappa$) and without (red, $\rho=\rho_B$) the \gls{ema} Scaling Rule.
    Bands indicate mean and standard deviation across three runs.}
  \label{fig:scaling-ablation-2}
\end{figure}

To demonstrate that the \gls{ema} Scaling Rule works for a broad range of scalings $\kappa$ (i.e. \emph{beyond} those presented in 
\Cref{fig:r18-byol} and \Cref{subsec:self-supervised}), we investigate scaling down to $\kappa=1/8$ in \Cref{fig:scaling-ablation-2}.
We see that the \gls{ema} Scaling Rule works well down to the small batch size of 128, although matching is not perfect.
We suspect this is due to the presence of Batch Normalization layers in the ResNet-18 architecture, which underperform at small batch sizes \citep{DBLP:conf/icml/IoffeS15}.
The synthetic analysis of \Cref{subsec:toy-experiment} instead demonstrated the \gls{ema} Scaling Rule holding for scalings spanning factors of $\kappa$ that differ by 1024, with scaling error insensitive to the value of $\kappa$ for sufficiently low $\kappa$ (see \Cref{fig:curve-approximation-error}).

\FloatBarrier

\subsection{Scaling a ResNet-50 BYOL using LARS and Progressive Scaling}
\label{subsec:byol-additional}

\begin{figure}[ht]
    \centering
    \includegraphics[width=0.75\textwidth]{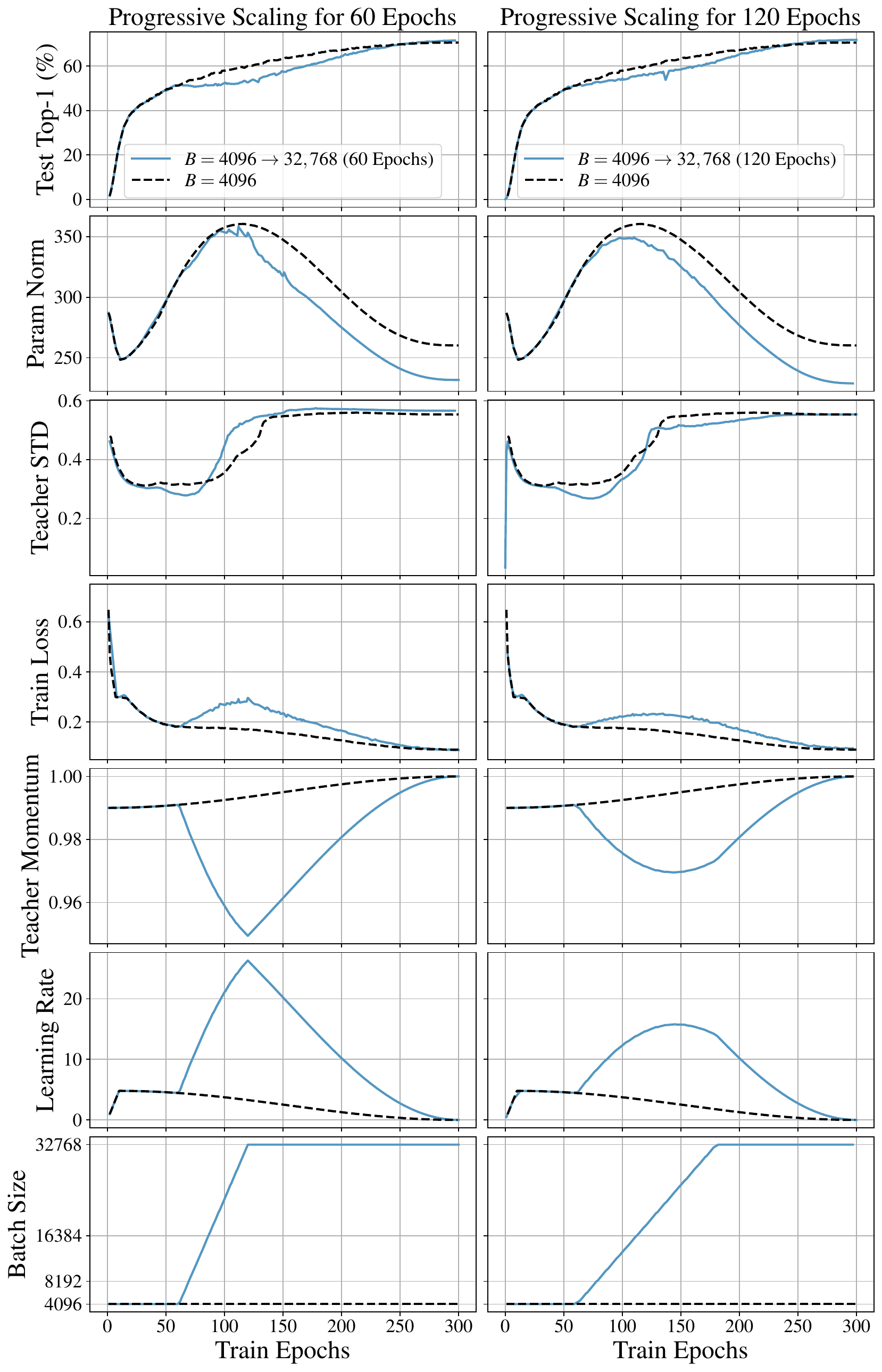}
    \caption{
    \emph{ResNet50 BYOL on ImageNet1k using LARS}
    for different configurations of progressive scaling.
    The baseline (black dashed)
    uses batch size 4096 and momentum $\rho_B=0.99$.
    We consider progressive scaling (blue) smoothly from epoch 60 for 60 epochs (left) and 120 epochs (right) up until batch size 32,768, scaling the learning rate linearly, and applying the \gls{ema} Scaling Rule.}
    \label{fig:r50-byol}
\end{figure}

Here we investigate whether Progressive Scaling and the \gls{ema} Scaling Rule can be used in practice where there is no known optimizer \gls{sde} approximation.
We use the default 300 epoch configuration for \gls{byol} \citep{DBLP:conf/nips/GrillSATRBDPGAP20} in \Cref{fig:r50-byol}.
We see that although trajectories during training do not match, we are able to match or surpass the linear probe performance of the \gls{byol} baseline at the larger batch size if 32,768.
\emph{This indicates that the contributions of our work have practical utility beyond the theoretical constraints.}

\ifthenelse{\equal{\anonymous}{0}}{
\FloatBarrier

The compute usage for the BYOL ResNet using LARS is detailed in  
\Cref{tab:byol-lars-compute}.

\begin{table}[ht]
  \caption{
  Compute usage for ResNet 50 LARS investigation in \Cref{fig:r50-byol}. 
  Values \emph{include} node allocation times (typically a small \% of corresponding total runtime), giving a practical estimate of reproduction cost. 
  All experiments conducted are using 80Gb A100s.}
  \label{tab:byol-lars-compute}
  \centering
  \small
\begin{tabular}{cccc}
\toprule
                              Batch Size &  GPUs &  Time (h) &  Compute (GPUh) \\
\midrule
$4,096\rightarrow 32,768$ (120 Epochs) &   128 &      14.1 &          1809.8 \\
 $4,096\rightarrow 32,768$ (60 Epochs) &   128 &      12.9 &          1655.9 \\
                               $4,096$ &    16 &      32.8 &           524.9 \\ \midrule 
\multicolumn{3}{l}{All other compute, e.g. code development, runs with errors, and debugging} & 17,654.6 \\ \midrule 
      \textbf{Total} &&& \textbf{21645.2} \\
\bottomrule
  \end{tabular}
\end{table}

}{\compute}

\FloatBarrier
\subsection{Preventing collapse phenomena in DINO at scale}
\label{subsec:dino}

Until now, our representatives \gls{ssl} method has been \gls{byol} for reasons discussed in \Cref{subsec:self-supervised}.
Here, we will turn our attention to \gls{dino} \citep{DBLP:journals/corr/abs-2104-14294},
which has the update rule presented in \Cref{def:dino}.
\begin{definition}[DINO Update] 
     \gls{dino} learns unsupervised features by 
     matching predictions over emergent pseudo-labels of a student backbone and head $f(\,\cdot\,;\rvtheta)$ 
     to those of an \gls{ema} teacher $f(\,\cdot\,;\rvzeta)$ through a 
     cross-entropy guided distillation procedure.
    \gls{dino} has a additional centering procedure, which is a form of batch normalization with momentum $\rho_c=0.9$ which we do not scale using the \gls{ema} Scaling Rule. 
    The update for the parameters of \gls{dino} is
    \begin{align}
    \rvtheta_{t+1}
    &=
    \rvtheta_t - \eta \times \frac1B
    \sum_{x\in\sB} \nabla_{\rvtheta} \Ls(x;\rvtheta_{t},\rvzeta_{t}, \rvc_t)
    \\
    \rvzeta_{t+1}
    &=
    \rho \,\rvzeta_t + (1-\rho)\,\rvtheta_{t+1}
    \\
    \rvc_{t+1}
    &= 
    \rho_{c} \, \rvc_t + (1 - \rho_{c}) \,\E_{x^\prime} \rvzeta(x^\prime) \\
    \text{with} \;\; \Ls(x;\rvtheta_{t},\rvzeta_{t}, \rvc_t)
    &=
    H\big( f(x_1, \rvtheta_{t}), f(x_2, \rvzeta_{t}) - \rvc_{t} \big)  + (x_1\leftrightarrow x_2),
    \end{align}
    where $H(\va,\vb)\equiv - \sum_{m=1}^M p_m(\va)\,\log p_m(\vb)$
    is the cross-entropy between categorical distributions over $M$ (emergent pseudo-)classes given logits $\va,\vb\in\R^{M}$, 
    $x_1$ and $x_2$ are two views of a single variate $x$, often produced by augmentations,
    and $x_1\leftrightarrow x_2$ denotes symmetrization over $x_1$ and $x_2$.
    \label{def:dino}
\end{definition}
In practice, \gls{dino} employs multi-crop \citep{DBLP:journals/corr/abs-2104-14294}.
We omit this detail for clarity of presentation, although we \emph{do} use multi-crop in the experiments that follow.

Our interest \gls{dino} is due to the  difficulty in its optimization\footnote{For an example, see
\url{https://github.com/facebookresearch/dino/issues/43\#issuecomment-881453515}.}, and in particular, preventing collapse phenomena in \gls{dino} at batch sizes above 1024, which is an open research problem.
In this section, we will show that a combination of the \gls{ema} Scaling Rule (\Cref{def:ema-sr}) and 
Progressive Scaling (\Cref{def:progressive-scaling}) enable training of \gls{dino} beyond batch size 1024 without sacrificing performance.

\paragraph{Hyperparameters} Base hyperparameters are presented in \Cref{tab:dino-hp}.

\begin{table}[t]
  \caption{\gls{dino} ViT-B/16 Training hyperparameters.}
  \label{tab:dino-hp}
  \centering
  \small
  \begin{tabular}{lcccc}
    \toprule
    & \gls{dino} ViT-B/16 \\
    \midrule
    CIFAR10 Linear Probe Top-1 ($\rho_B=0.996$) & 85.38\%  \\
    CIFAR10 Linear Probe Top-1 ($\rho_B=0.992$) & 86.96\%  \\
    \midrule
    Weight initialization & \texttt{trunc\_normal(.02)}  \\
    Normalization    & Layer Norm  \\
    Learning rate schedule & Single Cycle Cosine \\    
    Learning rate warmup (epochs) & 50 \\    
    Learning rate minimum value & $1\times10^{-6}$ \\        
    Training duration (epochs) & 280 \\    
    Optimizer & AdamW \\    
    Optimizer scaling rule & Adam \\
    Base ($\beta_1$, $\beta_2$) & (0.9, 0.95) \\
    Base learning rate & $3\times 10^{-4}$  \\
    Base batch size ($B$) & 1024  \\
    Base teacher momentum ($\rho_B$) & 0.992 or 0.996  \\
    Base weight decay & 0.04 \\
    Weight decay scaling rule & Linear \\
    Weight decay skip bias & Yes \\
    Center Momentum & 0.9 \\
    Center Momentum Scaling Rule & None \\
    Precision & \texttt{bf16} \\
    Augmentation stack & \texttt{DINO multi-crop} \citep{DBLP:conf/nips/CaronMMGBJ20} \\   
    \bottomrule
  \end{tabular}
\end{table}

\begin{figure}[ht]
    \centering
    \includegraphics[width=0.99\textwidth]{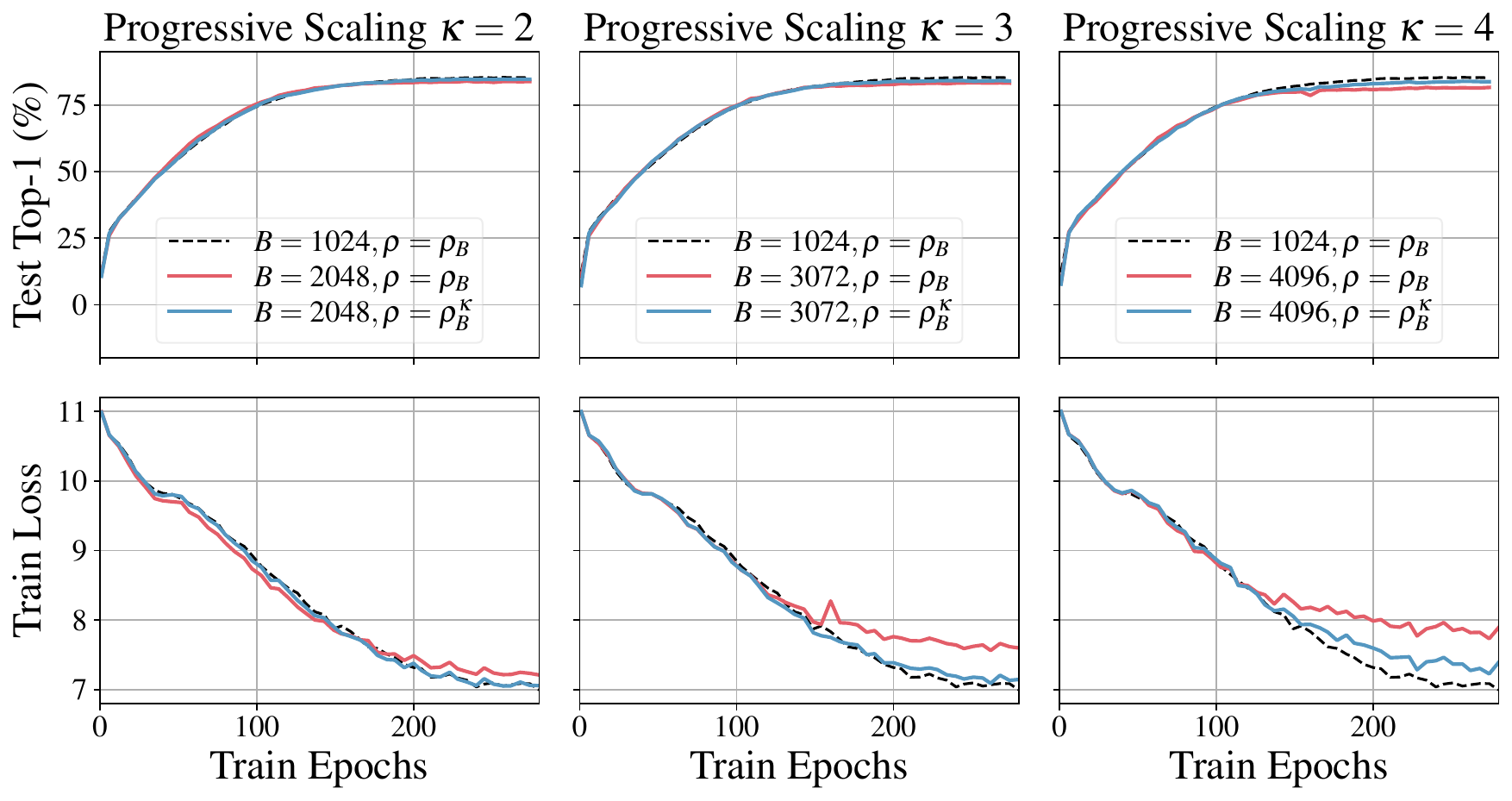}
    \caption{
    \emph{DINO ViT-B/16 on CIFAR-10} for different scalings $\kappa$ and base teacher momentum $\rho_B=0.996$.
    The baseline model ($\kappa=1$, black dashed) uses batch size 1024 and center momentum $\rho_c=0.9$, and is scaled up from batch size 2048 (left) to 4096 (right) with (blue, $\rho=\rho_B^\kappa$) and without (red, $\rho=\rho_B$) the \gls{ema} Scaling Rule.
    Between epochs 100 and 180 we scale the batch size using progressive scaling (\Cref{def:progressive-scaling}).
    }
    \label{fig:dino-cifar10-0.996}
\end{figure}

\begin{figure}[th]
    \centering
    \includegraphics[width=0.99\textwidth]{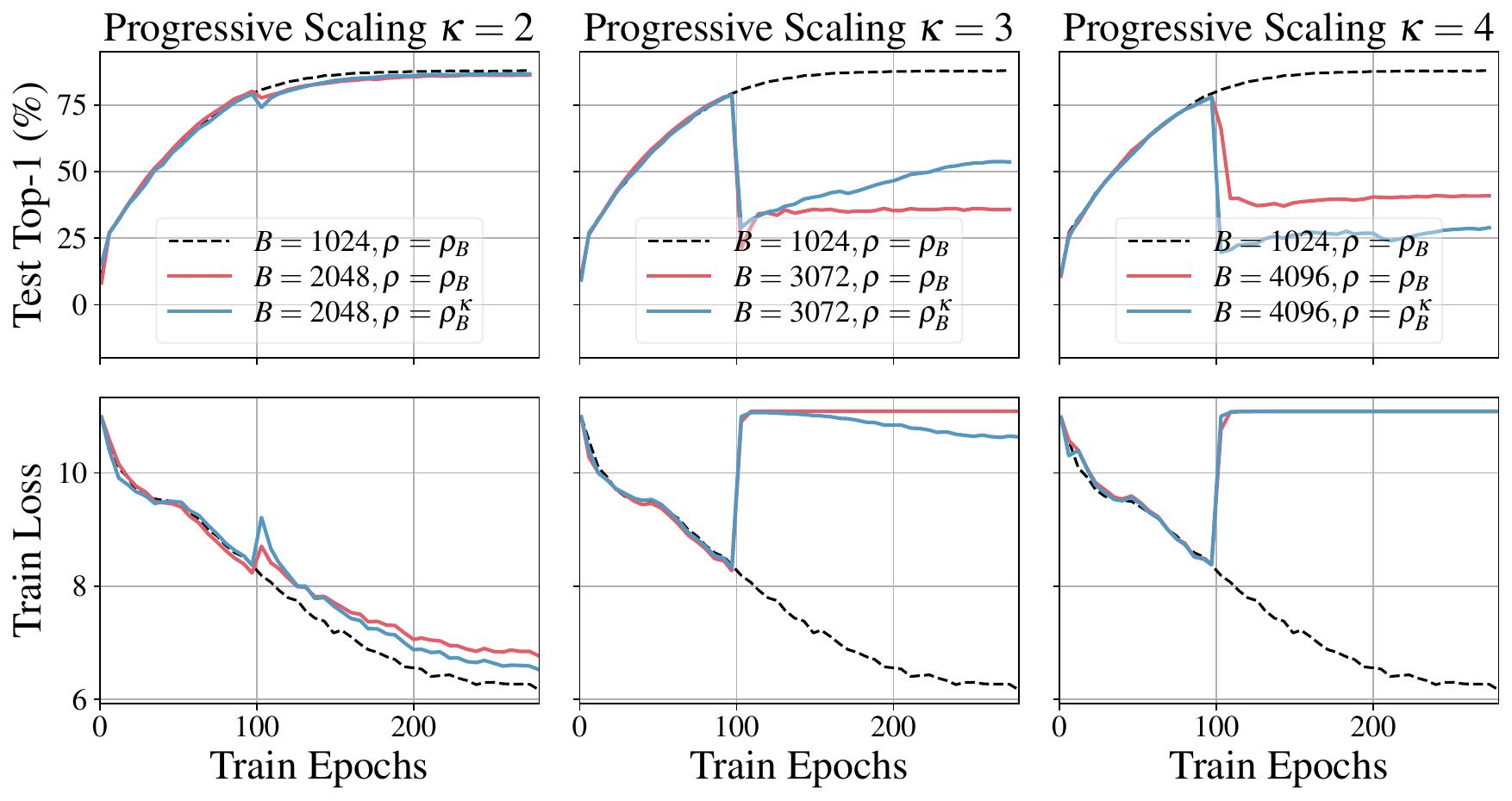}
    \caption{
    \emph{DINO ViT-B/16 on CIFAR-10} for different scalings $\kappa$ and base teacher momentum $\rho_B=0.992$.
    The baseline model ($\kappa=1$, black dashed) uses batch size 1024 and center momentum $\rho_c=0.9$, and is scaled up from batch size 2048 (left) to 4096 (right) with (blue, $\rho=\rho_B^\kappa$) and without (red, $\rho=\rho_B$) the \gls{ema} Scaling Rule.
    Between epochs 100 and 180 we scale the batch size using progressive scaling (\Cref{def:progressive-scaling}).
    }
    \label{fig:dino-cifar10-0.992}
\end{figure}

\paragraph{Results}
In \Cref{fig:dino-cifar10-0.996,fig:dino-cifar10-0.992} we show the results obtained training \gls{dino} on CIFAR-10 with $\rho_B=0.996$ and $\rho_B=0.992$ respectively at the reference batch size of 1024. 
We employ smooth Progressive Scaling (\Cref{def:progressive-scaling}) between epochs 120 and 180.

At batch size 2048, the training loss matches the reference \emph{only} when the \gls{ema} Scaling Rule is applied, whereas the run \emph{without} the scaling rule diverges from the reference. 
The impact of this divergence is emphasized as we consider the larger batch size of 4096.
Here. there is also a gap \emph{with} the \gls{ema} Scaling Rule, however is approximately three times smaller than the gap \emph{without} the \gls{ema} Scaling Rule.

Additionally, we observe that using $\rho_B=0.992$ yields higher Top-1 accuracy over $\rho_B=0.996$, and in our experiments, using the \gls{ema} Scaling Rule \emph{always} performs better in terms of linear probe performance than not using the scaling rule.

\FloatBarrier
\newpage

\begin{figure}[htb!]
    \centering
    \includegraphics[width=0.99\textwidth]{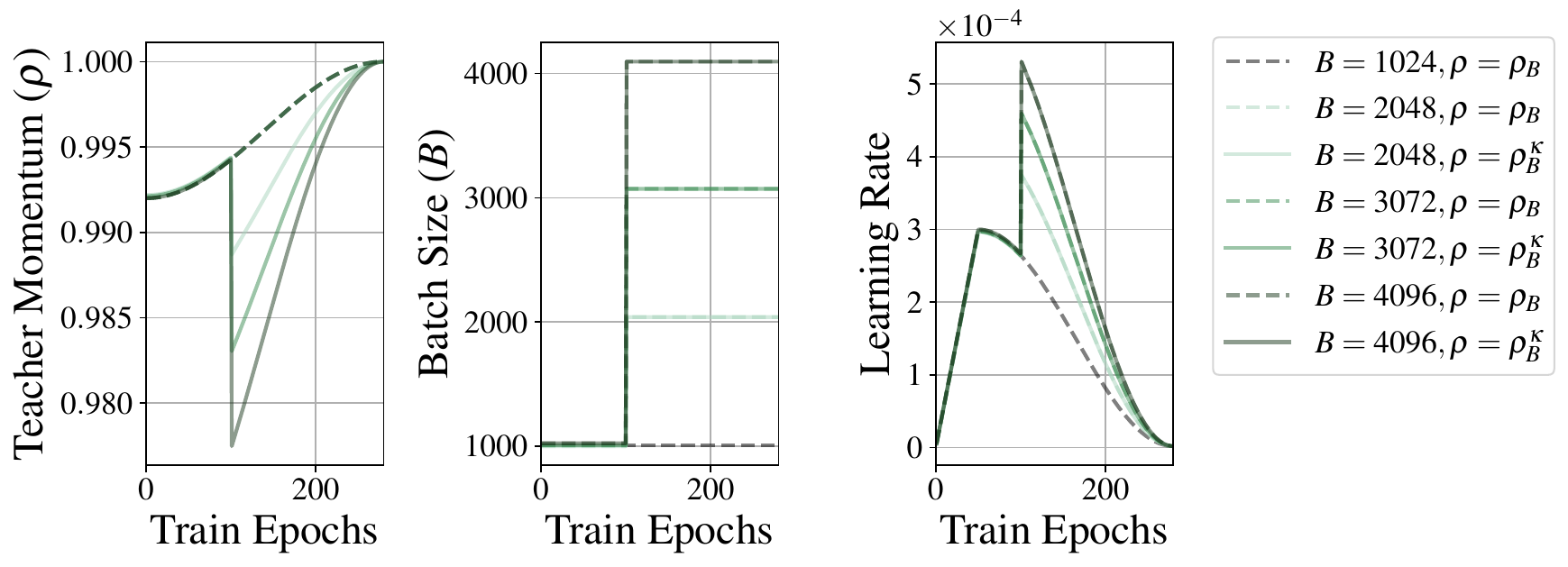}
    \caption{
    \emph{DINO ViT-B/16 on CIFAR-10} for different scalings $\kappa$ and base teacher momentum $\rho_B=0.992$.
    We show how the hyperparameters $\rho$, $B$ and learning rate change with the Progressive Scaling in \Cref{def:progressive-scaling}. 
    These hyperparameters correspond to the training runs in \Cref{fig:dino-cifar10-0.992}. 
    Those for \Cref{fig:dino-cifar10-0.996} are identical, with the exception of $\rho$ that starts at $0.996$ instead of $0.992$.
    }
    \label{fig:dino-cifar10-0.996-hp}
\end{figure}

In \Cref{fig:dino-cifar10-0.996-hp} we show how the hyperparameters $\rho$, $B$ and learning rate change with the progressive scaling in \Cref{def:progressive-scaling}.

\begin{figure}[htb!]
    \centering
    \includegraphics[width=0.99\textwidth]{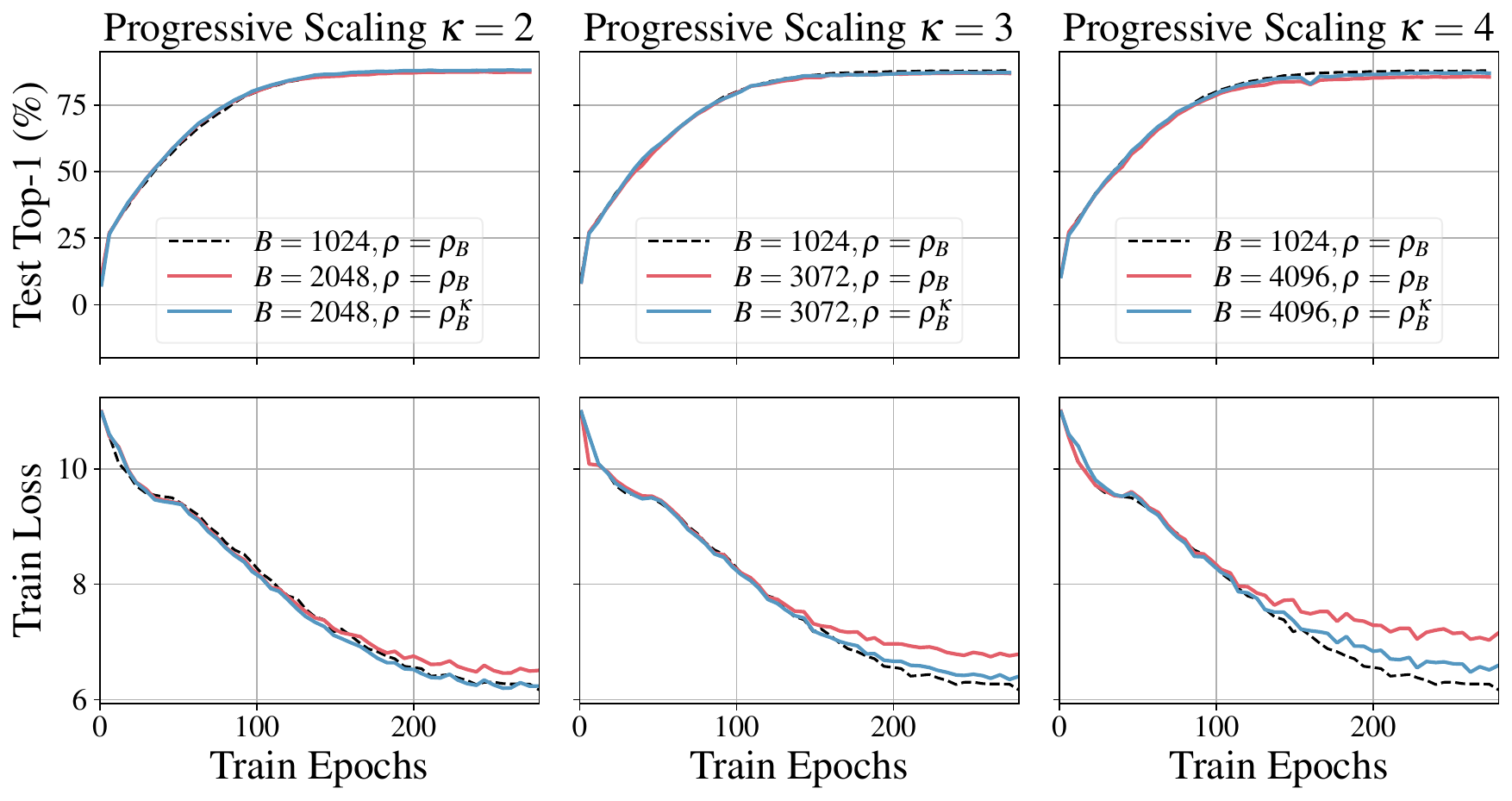}
    \caption{
    \emph{DINO ViT-B/16 on CIFAR-10} for different scalings $\kappa$ and base teacher momentum $\rho_B=0.992$.
    The baseline model ($\kappa=1$, black dashed) uses batch size 1024 and center momentum $\rho_c=0.9$, and is scaled up from batch size 2048 (left) to 4096 (right) with (blue, $\rho=\rho_B^\kappa$) and without (red, $\rho=\rho_B$) the \gls{ema} Scaling Rule. Progressive Scaling is employed with a sharp transition at epoch 100, leading to a collapse phenomenon.
     }
    \label{fig:dino-cifar10-0.992-step}
\end{figure}

\begin{figure}[htb!]
    \centering
    \includegraphics[width=0.99\textwidth]{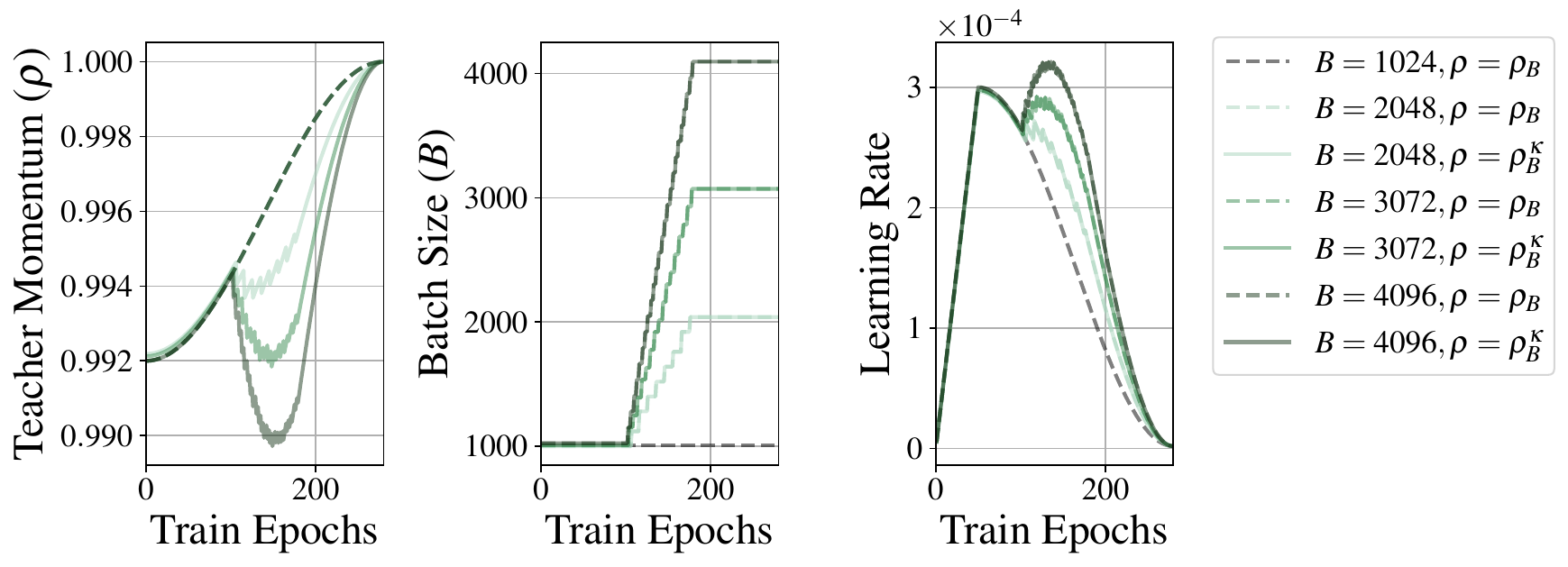}
    \caption{
    \emph{DINO ViT-B/16 on CIFAR-10} with $\rho_B=0.992$ and a sharp transition in batch size at epoch 100.
    We show how the hyperparameters $\rho$, $B$ and learning rate change with sudden scaling. These hyperparameters correspond to the training runs in \Cref{fig:dino-cifar10-0.992-step}.
    }
    \label{fig:dino-cifar10-0.992-hp-step}
    \vspace{-0.2cm}
\end{figure}

\FloatBarrier
We also attempted to use a sharp batch size transition (\Cref{fig:dino-cifar10-0.992-step,fig:dino-cifar10-0.992-hp-step}), which leads to the collapse pheonomena observed in prior work. 
This collapse happens with and without the \gls{ema} Scaling Rule.
We suspect this is due to dynamics specific to \gls{dino}'s early phase that are even more challenging to replicate under discretization than those of \gls{byol}.

\ifthenelse{\equal{\anonymous}{0}}{
\FloatBarrier

\paragraph{Compute} The compute usage for the DINO investigations is detailed in 
\Cref{tab:dino-compute}.

\begin{table}[ht]
  \caption{
  Compute usage for DINO investigations. 
  Values \emph{include} node allocation times (typically a small \% of corresponding total runtime), giving a practical estimate of reproduction cost. 
  All experiments conducted are using 80Gb A100s.}
  \label{tab:dino-compute}
  \centering
  \small
\begin{tabular}{cccccc}
\toprule
                             Batch Size &  GPUs &  Time (h) &  Compute/Run (GPUh) &  Runs &  Compute (GPUh) \\
\midrule
                              $1,024$ &    24 &       6.8 &               163.5 &     2 &           327.0 \\
             $1,024\rightarrow 2,048$ &    40 &       4.6 &               182.4 &     1 &           182.4 \\
             $1,024\rightarrow 3,072$ &    48 &       4.0 &               189.9 &     1 &           189.9 \\
             $1,024\rightarrow 4,096$ &    64 &       3.3 &               212.3 &     1 &           212.3 \\
$1,024\rightarrow 2,048$ (100 Epochs) &    40 &       4.8 &               190.6 &     4 &           762.3 \\
$1,024\rightarrow 3,072$ (100 Epochs) &    48 &       4.0 &               192.5 &     4 &           769.9 \\
$1,024\rightarrow 4,096$ (100 Epochs) &    64 &       3.6 &               232.1 &     4 &           928.2 \\ \midrule 
\multicolumn{5}{l}{All other compute, e.g. code development, runs with errors, and debugging} & 38239.2 \\ \midrule 
      \textbf{Total} &&&&& \textbf{41,611.1} \\
\bottomrule
  \end{tabular}
\end{table}

}{\compute}

Our results in this section show it is possible to scale \gls{dino} to large batch sizes \emph{without} sacrificing performance by using \emph{both} the \gls{ema} Scaling Rule and Progressive Scaling, providing the batch size schedule of Progressive Scaling is not sudden.

\section{Additional details on numerical stability}

A general analysis of overflow and underflow of the \gls{ema} Update (\Cref{def:emaUpdateDefinition}) or \gls{ema} Scaling Rule
(\Cref{def:emaScalingRuleExponentialVersion})
for different momenta $\rho$, particularly for IEE-754 floating point values, is beyond the scope of this work due to non-linearity from mechanisms like gradual underflow \citep{iee}.

In our setting, do not suffer from practical overflow or underflow issues through exponentiation when applying the \gls{ema} Scaling Rule, as \texttt{FP32} precision allows a maximum $\rho=1-\varepsilon$, or minimum $\rho=\varepsilon$ with $\varepsilon\approx 1.2\times 10^{-7}$.
Take self-supervised image representation learning (\Cref{subsec:self-supervised}) as a baseline, with $\kappa=1$ corresponding to batch size $B=4096$ with momentum $\rho_B=0.996$.
The maximum value of $\rho$ corresponds to scaling
$\kappa=\log(\rho_B)/\log(\varepsilon)\approx 1/(32K)$,
give a batch size less than one, while 
the minimum value of $\rho$ corresponds to scaling
$\kappa=\log(\rho_B)/\log(1-\varepsilon)\approx 4K$, giving a batch size $B\approx 8M$ which is beyond current hardware feasibility, and beyond the breakdown of known optimizer scaling rules \citep{DBLP:conf/nips/LiMA21}.

\begin{figure}
    \centering
    \includegraphics[width=0.8\textwidth]{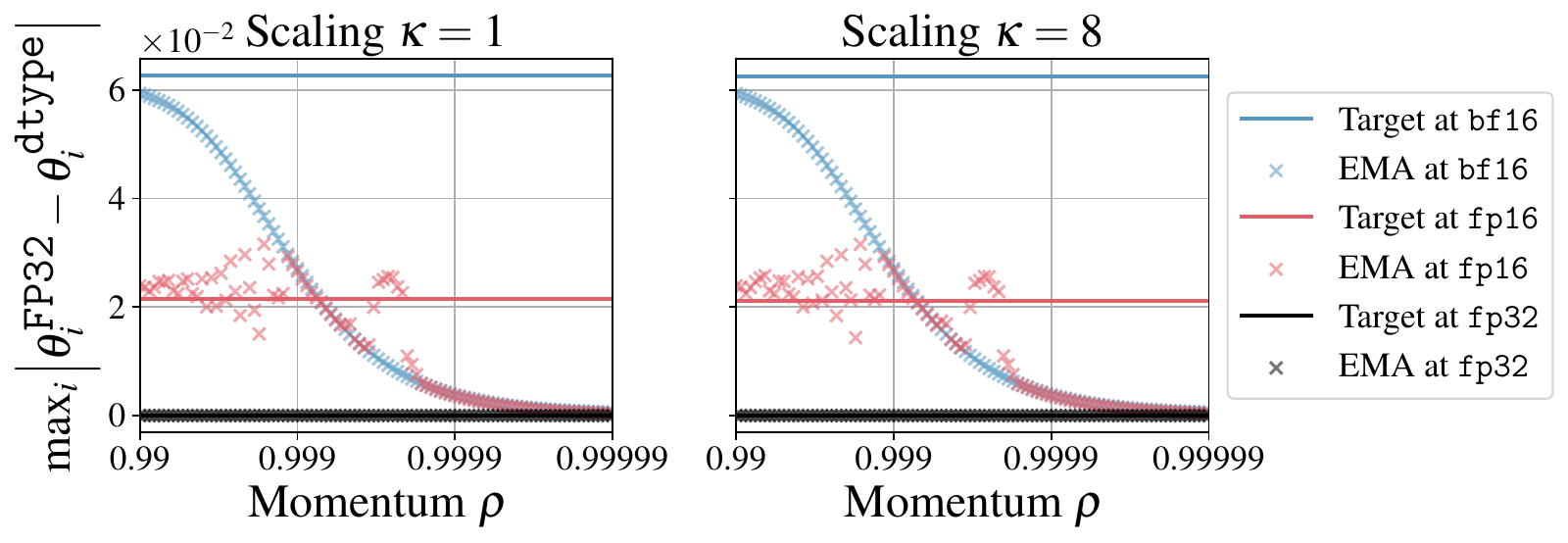}
    \caption{Numerical precision of target and EMA networks compared to an \texttt{FP32} reference on a regression task for a range of momenta.}
    \label{fig:numerics}
\end{figure}

To examine how momentum may induce numerical errors in practice during training, we train a linear regression model with a Polyak-Ruppert average \Cref{def:polyak-ruppert-average}, and
and track the difference between \texttt{FP32} model weights and weights in i) \texttt{BF16}; ii) \texttt{FP16}; and iii) a second \texttt{FP32} run, which act as a proxy for overflow and underflow.
In \Cref{fig:numerics} we plot these differences using the maximum absolute difference between model parameters, where the maximum is taken over individual weights
\begin{equation}
    \text{MaxAbsDiff}(\texttt{dtype})=\max_{i=1}^P\left|\theta_i^{\texttt{FP32}}-\theta_i^{\texttt{dtype}}\right|,
\end{equation}
where $P$ is the number of parameters in the model.
We observe that when model weights and \gls{ema} weights are \texttt{FP16} (never done in practice), an increasing variance happens for \texttt{FP16} as the value of the momentum $\rho$ approaches 0.99999, whereas \texttt{BF16} and \texttt{FP32} are stable. 
We stress that all experiments presented in the paper store weights for target model \emph{and} \gls{ema} in \texttt{FP32} and use automatic-mixed precision to cast them to \texttt{BF16} during training, and so do not encounter momentum-induced overflow or underflow.

\ifthenelse{\equal{\anonymous}{0}}{\FloatBarrier
\section{Contributions}
\label{sec:attribution}

All authors contributed to writing this paper, designing the experiments, discussing results at each stage of the project.

\paragraph{Preliminary work} 
Derivation of the \gls{ema} Scaling Rule with learning rate $\eta=0$, initial synthetic and self-supervised ImageNet1k experiments done by Dan Busbridge.

\paragraph{EMA scaling rules for constant gradients} Original proof of 
\Cref{eq:scalingRuleSummaryEquation}
and the form of $\delta(\eta,\rho,\kappa)$ in \Cref{eq:scaling-error} done by Eeshan Gunesh Dhekane.
Final proof presented in 
\Cref{app:matrix-calculations}
done by Dan Busbridge, verified by Eeshan Gunesh Dhekane and Pierre Ablin.

\paragraph{EMA approximation theorems with SDEs}
Proofs of validity of EMA Scaling Rule in the SDE limit presented in 
\Cref{subsec:ema-sdes} and 
\Cref{app:ema-approximation-theorem}
done by Pierre Ablin.

\paragraph{Polyak-Ruppert averaging in a simple setting}
Design of noisy parabola setting of \Cref{subsec:toy-experiment} and initial experiments done by Russ Webb.
Design of $\rho^*$-optimality search (\Cref{eq:optimal-momentum}), final experiments and analysis of \Cref{subsec:toy-experiment} and \Cref{app:noisy-parabola} done by Dan Busbridge.

\paragraph{Polyak-Ruppert averaging on image classification}
ResNetv2-50 reproduction (\Cref{tab:sup-r50-recipe}) and baseline momentum identification done by Jason Ramapuram.
Final ImageNet1k experiments and analysis of \Cref{subsec:supervised-polyakking} and \Cref{app:subsec:polyak-image-classification,subsec:polyak-bn} done by Dan Busbridge.

\paragraph{Automatic speech recognition}
Experiments and analysis of automatic speech recognition using Polyak-Ruppert averaging (\Cref{subsec:supervised-polyakking}) and continuous pseudo-labeling (\Cref{subsec:semi-supervised} and \Cref{app:speech}), as well as design choice of a seed model to start pseudo-labeling (aligning quality of the seed models for different batch size settings before pseudo-labeling process) done by Tatiana Likhomanenko.

\paragraph{Self-supervised image representation learning}
BYOL ResNet-18 recipe (\Cref{tab:byol-r18-recipe}) and experiments on CIFAR10 using \gls{sgd} (\Cref{fig:r18-byol}), and BYOL ResNet-50 experiments using LARS (\Cref{subsec:byol-additional}) done by Dan Busbridge. BYOL ResNet 50 baseline implementation and BYOL \gls{vit} recipe (\Cref{tab:byol-recipe}) done by Jason Ramapuram.
BYOL \gls{vit} exploratory ablations done by Eeshan Gunesh Dhekane and Jason Ramapuram.
All final BYOL \gls{vit} experiments and analysis 
(\Cref{fig:vitb-byol} and 
\Cref{app:byol-progressive-scaling-regimes,app:byol-waterfall,app:byol-vit-ln-vs-bn}) done by Jason Ramapuram. Baseline DINO reproduction done by Dan Busbridge. 
DINO experiments and analysis (\Cref{subsec:dino}) done by Xavier Suau Cuadros.

\paragraph{Progressive Scaling}
Progressive Scaling (\Cref{def:progressive-scaling} and \Cref{alg:progressive-scaling}) 
is proposed by Dan Busbridge based on discussions with Xavier Suau Cuadros, Tatiana Likhomanenko, Jason Ramapuram, Russ Webb, and the authors of \citet{DBLP:conf/nips/MalladiLPA22}. 
Adaptation of progressive scaling to semi-supervised learning in  automatic speech recognition (\Cref{app:speech-progressive}) done by Tatiana Likhomanenko, and to self-supervised learning in vision done by Dan Busbridge and Jason Ramapuram for BYOL (\Cref{fig:r18-byol,fig:vitb-byol} and \Cref{app:byol-waterfall,app:byol-vit-ln-vs-bn,subsec:byol-additional}) and Xavier Suau Cuadros for DINO (\Cref{subsec:dino}).

\paragraph{Limiting behavior of Polyak-Ruppert averaging} Original proof of limiting behavior of Polyak-Ruppert averaging done by Eeshan Gunesh Dhekane.
Final proof presented in 
\Cref{app:asymptoticAnalysis}
done by Dan Busbridge, verified by Eeshan Gunesh Dhekane.

\paragraph{Numerical stability analysis} Polyak-Ruppert experiment (\Cref{fig:numerics}) using linear regression for various floating point precisions done by Jason Ramapuram.

\paragraph{Implementation details} 
Investigations carried out in two distributed, scalable frameworks: Jax for automatic speech recognition experiments, done by Tatiana Likhomanenko; and PyTorch for all remaining investigations, done by 
Dan Busbridge, Xavier Suau Cuadros, Eeshan Gunesh Dhekane, Jason Ramapuram and Russ Webb. 
Initial implementation of progressive scaling experiments for incremental-style strategies (e.g. \Cref{app:byol-waterfall}) showing feasibility done by Jason Ramapuram, and subsequent progressive scaling implementations for smooth strategies (e.g. \Cref{subsec:byol-additional,subsec:dino}) done by Dan Busbridge and Xavier Suau Cuadros.
}{}

\end{document}